\documentclass{article}

\usepackage{arxiv}

\usepackage[utf8]{inputenc} 
\usepackage[T1]{fontenc}    
\usepackage{hyperref}       
\usepackage{url}            
\usepackage{booktabs}       
\usepackage{amsfonts}       
\usepackage{microtype}      
\usepackage{graphicx}
\usepackage{doi}

\usepackage{amssymb}
\usepackage{amsthm}
\usepackage{listings}
\usepackage{amsmath}
\usepackage{subfig}
\usepackage{pdflscape}
\usepackage{array}
\usepackage{multirow}
\usepackage{lscape}

\newtheorem{defka}{Definition}
\newtheorem{thm}{Theorem}
\newtheorem{assumption}{Assumption}
\newtheorem{corollary}{Corollary}
\newtheorem{lem}[thm]{Lemma}
\newtheorem{pf}{Proof}

\title{Mathematical modelling and virtual decomposition control of heavy-duty parallel–serial hydraulic manipulators}

\author{ \href{https://orcid.org/0000-0002-5767-815X}{\includegraphics[scale=0.06]{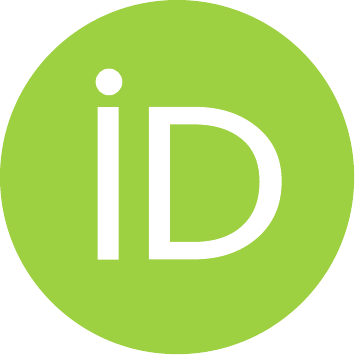}\hspace{1mm}Goran R.~Petrovi\'c}\thanks{Address all correspondence to this author. \newline
		\includegraphics[width= .025\textwidth]{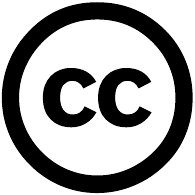} \includegraphics[width= .025\textwidth]{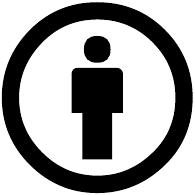} \includegraphics[width= .025\textwidth]{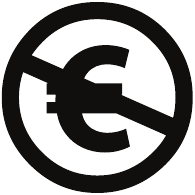} \includegraphics[width= .025\textwidth]{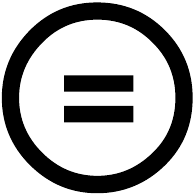} \newline
©2021. This manuscript version is made available under the CC-BY-NC-ND 4.0 license \newline \url{https://creativecommons.org/licenses/by-nc-nd/4.0/}} \\
	Faculty of Engineering and Natural Sciences, \\ Unit of Automation Technology \\ and Mechanical Engineering, \\ Tampere University, Tampere, Finland \\
	\texttt{goran.petrovic@tuni.fi} \\
	\And
	\href{https://orcid.org/0000-0003-1799-4323}{\includegraphics[scale=0.06]{orcid.pdf}\hspace{1mm}Jouni Mattila} \\
Faculty of Engineering and Natural Sciences, \\ Unit of Automation Technology \\ and Mechanical Engineering, \\ Tampere University, Tampere, Finland \\
	\texttt{jouni.mattila@tuni.fi} \\
}

\renewcommand{\headeright}{}
\renewcommand{\undertitle}{Accepted manuscript, \url{dx.doi.org/10.1016/j.mechmachtheory.2021.104680}}
\renewcommand{\shorttitle}{Mathematical modelling and virtual decomposition control of heavy-duty parallel–serial hydraulic manipulators}

\hypersetup{
pdftitle={VDC modelling paper preprint},
pdfsubject={q-bio.NC, q-bio.QM},
pdfauthor={Goran R.~Petrovic},
pdfkeywords={VDC modelling, arXiv},
}

\begin{document}
\maketitle

\begin{abstract}
	This paper proposes a novel modelling approach for a heavy-duty manipulator with parallel–serial structures connected in series. Each considered parallel–serial structure contains a revolute segment with rigid links connected by a passive revolute joint and actuated by a linear hydraulic actuator, thus forming a closed kinematic loop. In addition, prismatic segments, consisting of prismatic joints driven by hydraulic linear actuators, also are considered. 
	Expressions for actuator forces are derived using the Newton–Euler (N–E) dynamics formulation. The derivation process does not assume massless actuators decoupled from manipulator links,  which is common in the Lagrange dynamics formulation. Actuator pressure dynamics are included in the analysis, leading in total to a third-order system of ordinary differential equations (ODEs). With fewer parameters than its predecessors, the proposed model in the N–E framework inspires revision of the virtual decomposition control (VDC) systematic process to formulate a control law based on the new model. 
	The virtual stability of each generic manipulator revolute and prismatic segment is obtained, leading to the Lyapunov stability of the entire robot.
\end{abstract}


\section{Introduction}

Mathematical modelling and model-based control have drawn much attention in the field of electrically driven robots, as in \cite{slotine1987adaptive}, \cite{khosla1988experimental}, \cite{lewis2003robot} and \cite{zhu2013precision}.

Hydraulic actuators have higher power-to-weight ratio and robustness and less cost compared to their electric counterparts for given payloads. In addition, they can generate high force/torque without overheating, and load holding can be carried out without any energy use, as reported in detail in \cite{semini2010hyq}. 

Hydraulic robotic manipulators also are receiving attention, primarily because different original equipment manufacturers invest vast resources in developing automated solutions for their products, as recounted in more detail in \cite{son2020expert}. Aspirations to increase productivity and reduce human error,  operating costs and energy consumption are bound to change heavy-duty working machines into field-robotic systems. In addition to all this, a high number of units sold, together with the substantial growth in market size that is  projected, further motivates hydraulic robotics research \cite{mattila2017survey}.

Nonlinear model-based (NMB) control techniques have stood out as being able to provide the most advanced control performance for hydraulic robotic manipulators, exploiting the highly nonlinear mathematical model of a manipulator and the desired motion dynamics  \cite{mattila2017survey}, \cite{bech2013experimental}.  These model-based methods in {motion control} rely predominantly on explicit equations of motion (EOMs). Implementing these controllers requires solving the inverse dynamics problem, i.e. calculating the forces/moments required to produce linear/angular accelerations in a rigid-body system. When using these calculations in real-time motion control, the choice of a modelling framework and a computationally efficient inverse dynamics algorithm is highly significant. No less important is the proper formulation of feedback terms, which are bound to exist in the control law to ensure closed-loop stability. Thus, it is also desirable for a modelling framework to be in line with the systematic addressing of closed-loop stability.

Among the various approaches for dynamics modelling, the Lagrange formulation, based on the kinetic and potential energy, and the N–E formulation, based on the balance of forces acting on a rigid manipulator link, are the two most common \cite{siciliano2016springer}, {and the N–E approach is  considered more fundamental} \cite{jazar2010theory}. Apart from these two most frequently employed formalisms, varieties of Kane's equations \cite{kane1983use} also have been used to model different manipulators.

The Lagrange formulation has been used extensively to model revolute segments of parallel–serial manipulators with passive revolute joints actuated by linear actuators. In the mainstream of this kind of approach, a hydraulic robot with closed kinematic loops is modelled mainly as a series of nearly rigid bodies in a kinematic chain, with linear actuators considered decoupled from manipulator links and usually approximated as massless. This removes the need to calculate the Lagrange multipliers, \cite{marques2021examination}. However, the question of modelling accuracy is raised since discrepancies in dynamics calculations may exist at high velocities, where actuator inertial forces can significantly affect the system dynamics. At zero velocity, actuator masses also induce forces/moments of a certain magnitude. Notable examples of this kind of simplified modelling, using the Lagrange formalism, can be found in  \cite{habibi1991computed}, \cite{bu2000observer} and \cite{mattila2000energy}. The equivalent piston and cylinder mass notion has been a step closer to a more accurate model in \cite{bech2013experimental}, but closed-loop kinematics were not considered. 
Even with these a priori approximations, after including actuator dynamics, control of a hydraulic manipulator, described with a third-order model in the form of ODEs, using specific NMB techniques such as feedback linearization or backstepping, introduces significant  complexities.

A dynamics model of a hydraulic excavator with closed kinematic loops containing passive revolute joints and linear actuators also has been given in the form of Kane's equations in \cite{vsalinic2014dynamic}. However, this neglected the masses of the hydraulic actuators, and it did not deal with closed-loop kinematics. 
A complete and general dynamics model without any approximations, together with the determination of actuator forces for rigid links connected with passive joints, actuated by the electrically driven linear actuators, is given in \cite{cibicik2019dynamic}. Kane's equations, obtained by combining screw theory and the principle of virtual work, were used. Actuator dynamics and a control algorithm were not included.

The reformulation of the N–E approach, using 6D (spatial) vectors, given in \cite{featherstone1987robot}, has had the most notable use in the virtual decomposition control (VDC) community. It has become a central modelling tool, which has been proven in real-time control applications and has shown superior performance, \cite{mattila2017survey}. 

In the VDC approach, any manipulator is divided into several modular subsystems. Kinematics and dynamics of each subsystem obtained after the decomposition are being separately analysed. Relying on the notions of required velocity and required force, as will be  demonstrated, and taking full advantage of the N–E dynamics, the stability of the entire robot with actuator dynamics accounted can be rigorously guaranteed by systematically choosing control laws at the subsystem level, without imposing additional approximations. 

Control action in VDC predominantly relies on feed-forward terms that, in essence, generate actuation forces based on inverse dynamics. Feedback exists to overcome uncertainties, maintain stability and address transition issues. With careful choice of control values, virtual stability (see Definition \ref{def: virt stab}) on the level of each particular manipulator segment can be achieved. The virtual stability of every manipulator segment (revolute or prismatic) per se will guarantee the equivalent of Lyapunov stability, $L_2$ and $L_{\infty}$, stability of the entire robot  \cite{zhu2010virtual}. 

The modularity in the VDC approach arises because changing the control (or dynamics) equations of one subsystem does not affect the control equations of the rest of the system. This property makes the VDC a candidate for a leading control technique in future industrial innovations since the modularity property is crucial for handling complexity, as discussed in \cite{mastellone2021impact}.

Many SoA real-world robotics control performances have been reported for both hydraulic and electric actuators in the VDC framework, such as \cite{zhu1998adaptive}, \cite{zhu2002experimental}, \cite{zhu2005adaptive}, \cite{koivumaki2015stability}, \cite{koivumaki2016stability} and \cite{koivumaki2019energy}. A  performance index $\rho$ that presents the ratio of a maximum absolute position error and maximum absolute velocity, see \cite{mattila2017survey}, has often been used for benchmarking. A small value of this performance index (which indicates high performance) is obtained in all works assumed to be SoA and which rely on accurate system modelling.

Depending on how the virtual decomposition is performed, governing sets of EOMs vary since reaction forces are different at different virtual cutting points (VCPs). Thus, the most convenient way to decompose the manipulator is to determine the least number of subsystems, accompanied by the least number of parameters having a clear physical interpretation. At any rate, the final values of actuator forces/moments must remain the same, irrespective of the decomposition. Furthermore, it is necessary to emphasise that control computations in the VDC are proportional to the number of subsystems (see \cite{zhu2010virtual}), and every additional subsystem means that more on-line computational burden is added. A simple mass object, observed as a subsystem with $N$ DOFs, introduces a need to form at least $4  N$ additional equations since substantial quantities as total forces, forces, required total forces, and required forces must be calculated as will be elaborated in more detail. Different transformation matrices, on-line parameter value updates in adaptive control, and others may also be required.

\textit{This paper contributes with the novel mathematical model in the N–E framework, using 6D vectors, to address the dynamics of the considered class of hydraulic manipulators.} The proposed model has the following new essential properties:
\textbf{1)} the required number of virtual subsystems in the analysis of the manipulator dynamics is decreased, \textbf{2)} the straightforward expression for the actuator forces is derived, free of many surplus factors used earlier, \textbf{3)} the model can be easily incorporated into the VDC framework, providing the simplified control forming process and guaranteeing stability.

\begin{figure}[h!]
\centering
\includegraphics[width=.72\textwidth]{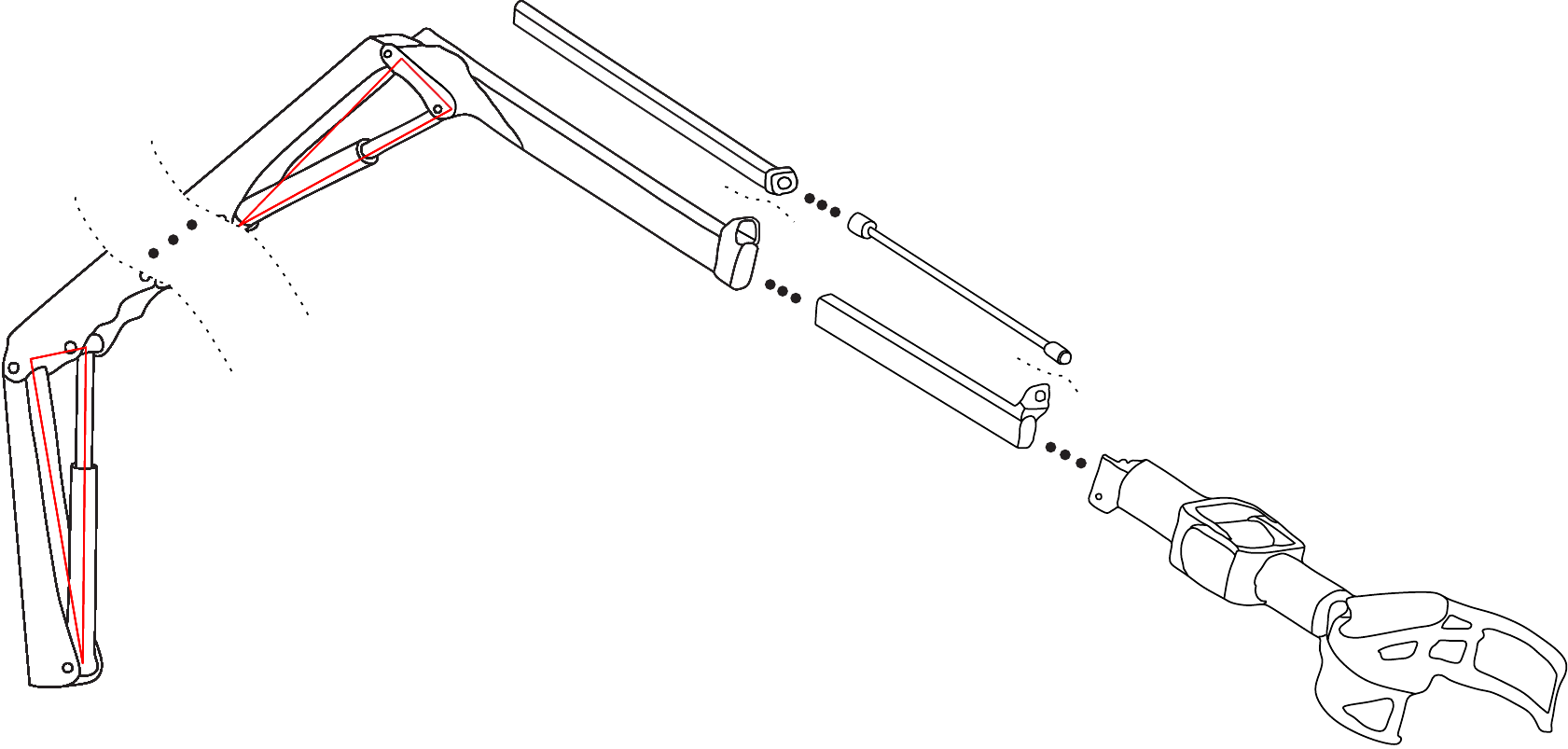}
\caption{Virtual decomposition of a manipulator using the approach presented here.} \label{fig: example new}
\end{figure}

The more detailed descriptions of the above contributions are as follows: 

\textbf{1)} A general manipulator from Fig. \ref{fig: example new}, containing revolute segments with a closed kinematic loop and prismatic segments connected in series, can be decomposed into fewer subsystems than in the VDC mainstream approach. 

\begin{figure}[h!]
\centering
\includegraphics[width=.72\textwidth]{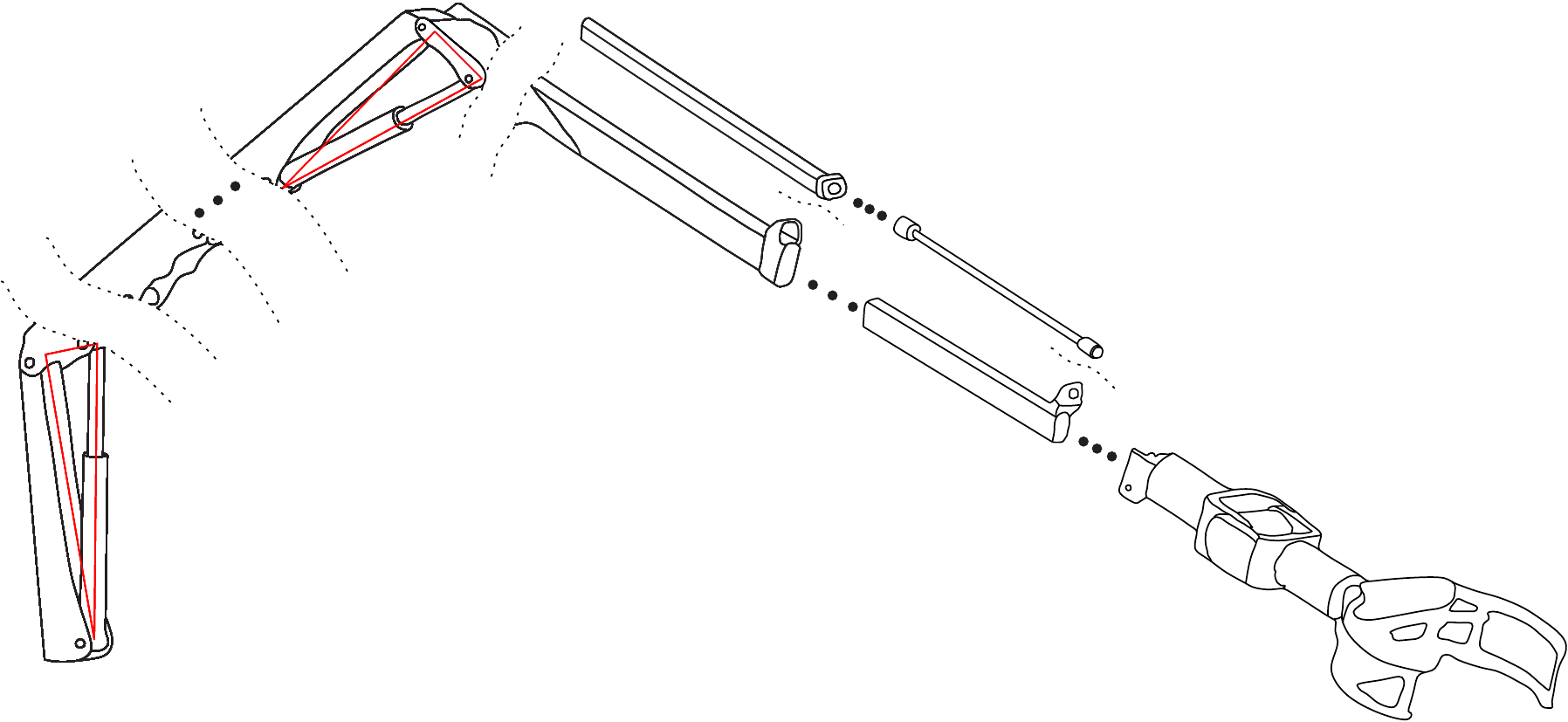}
\caption{Virtual decomposition of a manipulator in the VDC mainstream.} \label{fig: example old}
\end{figure}

\newpage

\begin{figure}[h!]
\centering
\includegraphics[width=.7\textwidth]{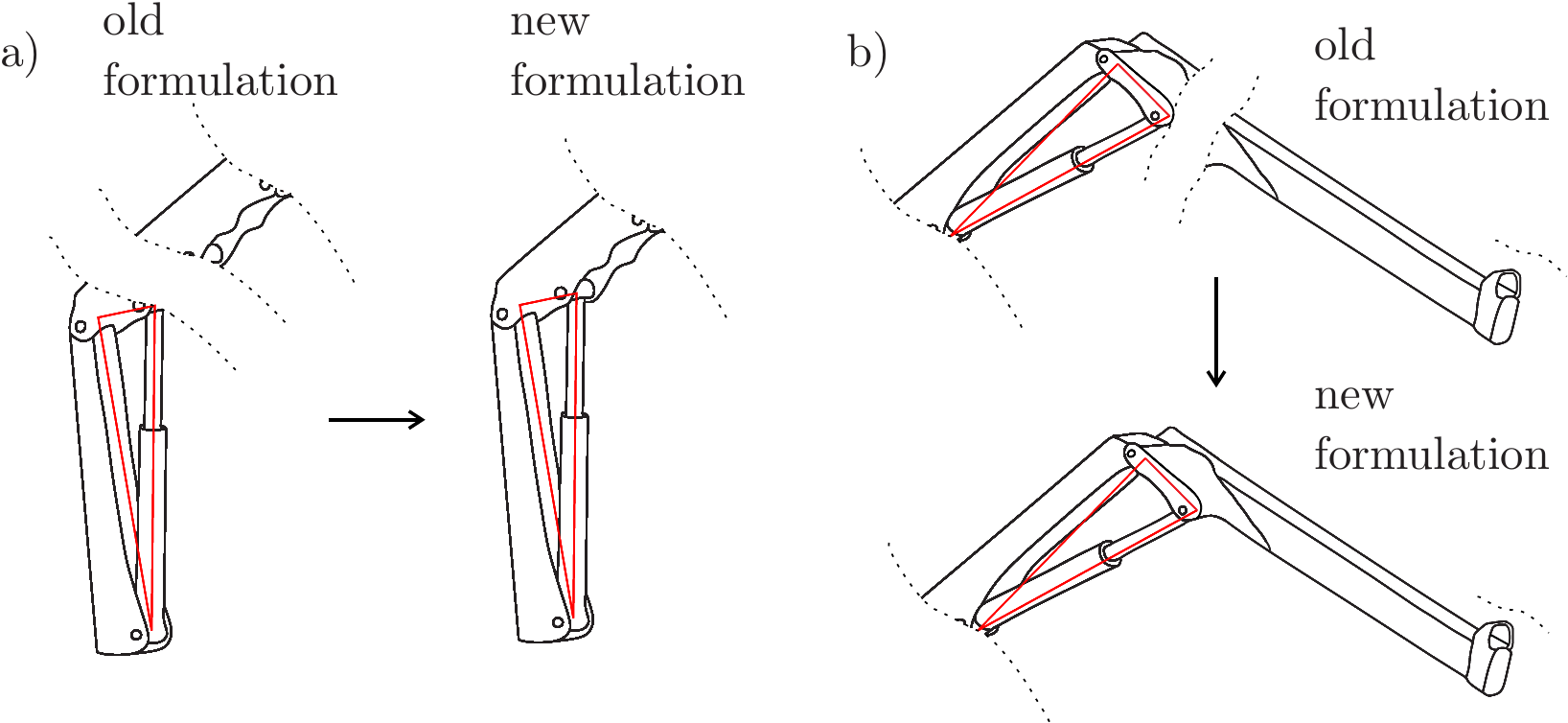}
\caption{Comparing virtual decompositions of relevant manipulator segments.} \label{fig: example oldnew}
\end{figure}

Figure  \ref{fig: example old} shows the same general manipulator decomposition resulting from the prevailing VDC approach. {Figure \ref{fig: example oldnew} compares  in more detail virtual decompositions using the currently widely accepted and here presented approach. As Fig. \ref{fig: example oldnew}a) shows, a revolute segment followed by another revolute segment now has one less subsystem in the analysis, and Fig. \ref{fig: example oldnew}b) shows that the same applies to a revolute segment followed by a prismatic segment.} 

Considering that prismatic segments are a minority compared to revolute segments, this result is more striking, since a robot consisting of $n$ revolute segments in series will now have $n$ subsystems instead of $2 \, n$.  This means that at least $4 \,  n \,N$ fewer equations will be formed in the process.

Apart from the decrease of computational load, the preparation process is shortened since fewer inertia tensors, masses, and position vectors have to be known and thus determined using CAD software or identified.

\textbf{2)} Dynamic interconnections between subsystems in the stated structures with closed kinematic chains have been strictly addressed using the so-called load distribution factors and internal force vectors in the existing VDC literature and papers. Analytic expressions for these did not exist until they were derived and validated in \cite{koivumaki2018addressing}, and since then are widely used. 
The broader use of derived expressions is hindered by them being overly burdensome, which is a pity considering their potential real-world impact. 

This paper shows that both load distribution factors and internal force vectors are surplus terms when calculating actuator forces in these manipulators. Actuator forces are now calculated using a computationally more efficient procedure. Apart from the decrease of the computational load,  clarity for the broader audience of readers is increased. The new model also makes the identification procedures more intuitive and makes all subsequent analyses less complex.

\textbf{3)} Finally, a systematic  process for formulating the control law using the new mathematical model is presented. Respective qualities of the model are passed to this new VDC implementation scheme, making it consequently more intuitive and requiring less on-line computations than earlier schemes. All the related stability proofs can be easily derived using the new model.

The rest of the paper is organised as follows. Section \ref{sec: mathematical foundations} presents essential mathematical preliminaries used for modelling, control formulation and stability analysis. Section \ref{sec: kinematics and dynamics analysis} presents kinematics and dynamics analysis with central theoretical modelling results presented here. Sections \ref{sec: control action forming} and \ref{sec: stability analysis} present how to form control values and provide stability of the whole manipulator. Section \ref{sec: contol algorithm} presents an algorithm for the systematic formulation of control law. Section \ref{sec: simulation results} validates the theoretical results by comparing them to results obtained using a commercial software package. Section \ref{sec: discussion} provides a discussion of the results. Section \ref{sec: conclusions} presents the conclusions. Proofs of theorems are given in appendices.

\section{Mathematical foundations and some VDC preliminaries}

\label{sec: mathematical foundations}

All the preliminaries in this section are reproduced from \cite{zhu2010virtual}, where relevant proofs for the theorems stated here can be found.

\subsection{Dynamics of a rigid body}

Every rigid body in the analysis will have at least one three-dimensional coordinate system $\left\lbrace \bf A \right\rbrace$ (called frame $\left\lbrace \bf A \right\rbrace$ in the following text) attached to it. 

Let the linear and angular velocities as sensed in frame $\left\lbrace \bf A \right\rbrace$ be denoted throughout the paper as ${^{\bf A}\boldsymbol{v}} = \begin{pmatrix}
{^{\bf A} v_{\rm x}} & {^{\bf A} v_{\rm y}} & {^{\bf A} v_{\rm z}}
\end{pmatrix}^{T}$ and  ${^{\bf A} \boldsymbol{\omega}} = \begin{pmatrix}
{^{\bf A} \omega_{\rm x}} & {^{\bf A} \omega_{\rm y}} & {^{\bf A} \omega_{\rm z}}
\end{pmatrix}^{T}$, respectively. Further, adopting the notation from \cite{zhu2010virtual}, the 6D linear/angular velocity vector of frame $\left\lbrace \bf A \right\rbrace$ can be written as:
\begin{equation}
{^{\bf A} \boldsymbol{V}} = \begin{pmatrix}
	{^{\bf A}\boldsymbol{v}^{T}} & {^{\bf A}\boldsymbol{\omega}^{T}}
\end{pmatrix}^{T} \in \mathbb{R}^6.
\label{eqn: AV}
\end{equation}

Let the force and moment vectors applied to the origin of frame $\left\lbrace \bf A \right\rbrace$  be similarly denoted as velocities using notation ${^{\bf A}\boldsymbol{f}} = \begin{pmatrix}
{^{\bf A} f_{\rm x}} & {^{\bf A} f_{\rm y}} & {^{\bf A} f_{\rm z}}
\end{pmatrix}^{T}$ for forces and notation ${^{\bf A} \boldsymbol{m}} = \begin{pmatrix}
{^{\bf A} m_{\rm x}} & {^{\bf A} m_{\rm y}} & {^{\bf A} m_{\rm z}}
\end{pmatrix}^{T}$ for moments. 

Same as the 6D linear/angular velocity vector in Eq. (\ref{eqn: AV}), the 6D force/moment vector, as sensed and expressed in frame $\left\lbrace \bf A \right\rbrace$, is introduced as:
\begin{equation}
{^{\bf A} \boldsymbol{F}} = \begin{pmatrix}
	{^{\bf A}\boldsymbol{f}^{T}} & {^{\bf A}\boldsymbol{m}^{T}}
\end{pmatrix}^{T} \in \mathbb{R}^6.
\label{eqn: AF}
\end{equation}

Further, let frame $\left\lbrace \bf B \right\rbrace$ also be attached to the same rigid body as frame $\left\lbrace \bf A \right\rbrace$. Also, moving the force from the frame $\left\lbrace \bf A \right\rbrace$ origin to the frame $\left\lbrace \bf B \right\rbrace$ origin introduces the moment of that force about the frame $\left\lbrace \bf B \right\rbrace$ origin. Consequently, quantities from Eqs. (\ref{eqn: AV}) and  (\ref{eqn: AF}) transform as:
\begin{equation}
{^{\bf B} \boldsymbol{V}} = {^{\bf A} \mathbf{U}_\mathbf{B}^T} \, 	{^{\bf A} \boldsymbol{V}},
\label{eqn: BV trans}
\end{equation}
and
\begin{equation}
{^{\bf A} \boldsymbol{F}} = {^{\bf A}\mathbf{U_B}} \, 	{^{\bf B} \boldsymbol{F}},
\label{eqn: BF trans}
\end{equation}
where ${^{\bf A}\mathbf{U_B}} \in \mathbb{R}^{6 \times 6}$ in Eqs. (\ref{eqn: BV trans}) and (\ref{eqn: BF trans}) is a force/moment transformation matrix, transforming the force/moment vector measured and expressed in frame $\left\lbrace \bf B \right\rbrace$ to the same force/moment vector measured and expressed in frame $\left\lbrace \bf A \right\rbrace$. The transformation matrix can be further written as:
\begin{equation}
{^{\bf A}\mathbf{U_B}} = \begin{pmatrix}
	{^{\bf A}\mathbf{R_B}} & \mathbf{O}_{3 \times 3} \\ \left( {^{\bf A}\boldsymbol{r}_\mathbf{{AB}}} \times  \right)
	{^{\bf A}\mathbf{R_B}} & {^{\bf A}\mathbf{R_B}} \\ 
\end{pmatrix},
\end{equation}
where  ${^{\bf A}\mathbf{R_B}} \in \mathbb{R}^{3 \times 3}$ is a rotation (direction cosine) matrix from  frame $\left\lbrace \bf A \right\rbrace$ to frame $\left\lbrace \bf B \right\rbrace$, and $\left(  {^{\bf A}\boldsymbol{r}_\mathbf{{AB}}}  \times  \right)$ is a skew-symmetric matrix operator defined as:
\begin{equation}
\left(  {^{\bf A}\boldsymbol{r}_\mathbf{{AB}}}  \times  \right) = \begin{pmatrix}
	0 & -r_{\rm z} & r_{\rm y} \\ r_{\rm z} & 0 & -r_{\rm x} \\ -r_{\rm y} & r_{\rm x} & 0
\end{pmatrix},
\end{equation}
with $r_{\rm x}$, $r_{\rm y}$ and $r_{\rm z}$ denoting distances from the origin of frame $\left\lbrace \bf A \right\rbrace$ to the origin of frame $\left\lbrace \bf B \right\rbrace$ along the frame $\left\lbrace \bf A \right\rbrace$ $x$-, $y$- and $z$-axis, respectively.

The net force/moment vector ${^{\bf A} \boldsymbol{F}^*} \in \mathbb{R}^{6}$ of the rigid body, in frame $\left\lbrace \bf A \right\rbrace$ is:
\begin{equation}
\mathbf{M}_{\bf A} \dfrac{\rm d}{\mathrm{d}t} \left( {^{\bf A}\boldsymbol{V}} \right) + \mathbf{C}_{\bf A} \left( {^{\bf A}\boldsymbol{\omega}}  \right) {^{\bf A}\boldsymbol{V}} + \mathbf{G}_{\bf A} = {^{\bf A} \boldsymbol{F}^*},
\label{eqn: tot force}
\end{equation}
where $\mathbf{M}_{\bf A} \in \mathbb{R}^{6 \times 6}$ is the mass matrix, $ \mathbf{C}_{\bf A} \left( {^{\bf A}\boldsymbol{\omega}}  \right) \in \mathbb{R}^{6 \times 6}$ is the matrix of Coriolis and centrifugal terms and  $\mathbf{G}_{\bf A} \in \mathbb{R}^{6}$ includes the gravity terms. Detailed expressions describing these matrices are given in \cite{zhu2010virtual}.

\subsection{Required velocities and required forces/moments}

\label{subsec: req vels}

The required velocity is a significant notion in the VDC framework, and it differs from the desired velocity. While the desired velocity serves as the reference trajectory of velocity with respect to time, the required velocity includes both the desired velocity and one or more terms that are related to control errors, such as position and force errors. When the control objective is to make a robot track its desired trajectory, it is required that the desired position and desired velocity from the path generator match as better as possible the measured velocity and position. If the control is performed in the joint space with the goal to track the given trajectory (we note that the required trajectory tracking can also be designed directly in the Cartesian space as shown in Section 3.3.6 in \cite{zhu2010virtual}), then the required joint velocities can be formed as:
\begin{equation*}
\dot{q}_{jr} = \dot{q}_{jd} + \lambda_{qj} \, \left(q_{jd} - q_j \right),
\end{equation*}
if the joint is revolute, or as:
\begin{equation*}
\dot{x}_{tjr} = \dot{x}_{tjd} + \lambda_{xj} \, \left(x_{tjd} - x_{tj} \right),
\end{equation*}
if the joint is prismatic, where $\dot{q}_{jd}$ is the desired revolute joint angular velocity, $q_{jd}$ is the desired revolute joint angle, $q_{j}$ is the measured revolute joint angle, $\dot{x}_{tjd}$ is the desired prismatic joint linear velocity, $x_{tjd}$ is the desired prismatic joint linear extension, $x_{tj}$ is the measured prismatic joint extension with $\lambda_{xj}, \lambda_{qj} > 0$, being positive adjustable parameters.

\newpage

Once the required joint velocities are obtained, the required linear/angular velocity vectors in any frame $\left\lbrace\bf A \right\rbrace$, labelled as  ${^{\bf A} \boldsymbol{V}_r}$ are formed just by following relations in the kinematic chain, going from the manipulator base to the tip and replacing  $\dot{q}_{j}$ with $\dot{q}_{jr}$ and $\dot{x}_{tj}$ with $\dot{x}_{tjr}$. The only additional effort in the case of closed kinematic loops considered here would be to relate required velocities in  kinematic closed-loops using the loop-closure functions, as will be shown in more detail. 
Required force/moment vector presents a standard notion in the VDC and is defined as:
\begin{equation}
\mathbf{M}_{\bf A} \dfrac{\rm d}{\mathrm{d}t} \left( {^{\bf A}\boldsymbol{V}_r} \right) + \mathbf{C}_{\bf A} \left( {^{\bf A}\boldsymbol{\omega}}  \right) {^{\bf A}\boldsymbol{V}_r} + \mathbf{G}_{\bf A} + \mathbf{K}_{\bf A} \, \left({^{\bf A} \boldsymbol{V}_r} - {^{\bf A} \boldsymbol{V}} \right) = {^{\bf A} \boldsymbol{F}_r^*},
\label{eqn: tot force 2 r}
\end{equation}
where $ \mathbf{K}_{\bf A} \in \mathbb{R}^{6 \times 6}$ is a positive-definite matrix. Its forming is based on Eq. (\ref{eqn: tot force}), and usage will be addressed in more detail later in the paper.


\begin{defka}
\cite{zhu2010virtual}. The Lebesgue space, denoted as $L_p$ with $p$ being a positive integer, contains all Lebesgue measurable and integrable functions $f(t)$:
\begin{eqnarray}
	\left\lvert \left\lvert f(t)  \right\rvert \right\rvert_p = \displaystyle\lim_{T \rightarrow \infty} \left[ \displaystyle\int_{0}^T  \left\lvert f(t)  \right\rvert^p d \tau  \right]^{1/p} < +\infty.
\end{eqnarray}
\end{defka}

\begin{itemize}
\item[a)] A Lebesgue measurable function $f(t) \in L_2$ if and only if \\ $\displaystyle\lim_{T \rightarrow \infty}  \displaystyle\int_{0}^T  \left\lvert f(t)  \right\rvert^2 d \tau   < +\infty$.
\item[b)] A Lebesgue measurable function $f(t)  \in L_\infty$ if and only if \\ $  \displaystyle\max_{t \in [0, +\infty)}  \left\lvert f(t)  \right\rvert d \tau   < +\infty$.
\end{itemize}

\begin{lem}
\cite{tao1997simple}. If $e(t) \in L_2$ and $\dot{e}(t) \in L_{\infty}$ then $\displaystyle\lim_{t\rightarrow \infty} e(t) \rightarrow 0$.
\label{lemma: Tao}
\end{lem}

\subsection{Virtual stability}

Simple oriented graphs are used to represent the topological structure and control relations of a complex robot.

\begin{defka}
\cite{zhu2010virtual}. A graph consists of nodes and edges. A directed graph is a graph in which all the edges have directions. An oriented graph is a directed graph in which each edge has a unique direction. A simple oriented graph is an oriented graph in which no loop is formed.
\label{thm: simple oriented graph}
\end{defka}

Virtual power flow (VPF) is an essential feature of the VDC approach. VPF defines dynamic interactions between subsystems per Definition \ref{def: VPF}, and it plays a vital role in virtual stability, which is given  with Definition \ref{def: virt stab}.
\begin{defka}
\cite{zhu2010virtual}. With respect to frame $\left\lbrace \bf A \right\rbrace$, virtual power flow (VPF) is defined as the inner product of the linear/angular velocity vector error and the {force/moment} vector error, that is:
\begin{equation}
	p_{\bf A} = \left( {^{\bf A} \boldsymbol{V}_r} -  {^{\bf A} \boldsymbol{V}} \right)^T \, \left( {^{\bf A} \boldsymbol{F}_r} -  {^{\bf A} \boldsymbol{F}} \right),
\end{equation}
where ${^{\bf A} \boldsymbol{V}_r} - {^{\bf A} \boldsymbol{V}} \in \mathbb{R}^6$, ${^{\bf A} \boldsymbol{F}_r} - {^{\bf A} \boldsymbol{F}} \in \mathbb{R}^6$.
\label{def: VPF}
\end{defka}

\begin{defka}
\cite{zhu2010virtual}. A subsystem that is virtually decomposed from a complex robot is said to be virtually stable, with its affiliated vector $\mathbf{x}(t)$ being a virtual function in $L_{\infty}$ and its affiliated vector $\mathbf{y}(t)$ being a function in $L_2$, if and only if there exists a non-negative accompanying function:
\begin{equation}
	\nu \geqslant \dfrac{1}{2} \, \mathbf{x}(t)^T \, \mathbf{P} \, \mathbf{x}(t),
\end{equation}
such that
\begin{equation}
	\dot{\nu} \leqslant - \mathbf{y}(t)^T \, \mathbf{Q} \, \mathbf{y}(t)    - s(t) + p_{\bf A} - \ p_{\bf C}
\end{equation}
holds, subject to:
\begin{equation}
	\displaystyle\int_0^{\infty} s(t) d\tau \geqslant -\gamma_s,
\end{equation} 
where $0 \leqslant \gamma_s \leqslant \infty$, and $\mathbf{P}$ and $\mathbf{Q}$ are two block-diagonal positive-definite matrices, and $p_{\bf A}$ and $p_{\bf C}$ denote the sum of VPFs in the sense of Definition \ref{def: VPF} at frames $\left\lbrace \mathbf{A} \right\rbrace$ (placed at driven VCPs) and $\left\lbrace \mathbf{C} \right\rbrace$ (placed at driving VCPs).
\label{def: virt stab}
\end{defka}

In Definition \ref{def: virt stab}, $s(t) = 0$ is a special case that satisfies the definition.

In Definition \ref{def: virt stab} a virtual function in $L_p$ is a candidate function in $L_p$ for $p = 2, \infty$. A virtual function in $L_p$ becomes a function in $L_p$ when every subsystem of a complex robot is virtually stable.

The unique characteristic of virtual stability is that VPF appears in the time derivative of the non-negative accompanying function assigned to each subsystem. These VPFs represent the dynamic interactions among subsystems. All the VPFs in the system cancel out since one subsystem's driven point is the driving point of the adjoined subsystem and virtual power flows are the same, except that they have opposite signs.

\begin{thm}
\cite{zhu2010virtual}. Consider a complex robot that is virtually decomposed into subsystems and is represented by a simple oriented graph in Definition \ref{thm: simple oriented graph}. If every subsystem is virtually stable in the sense of Definition  \ref{def: virt stab}, then all virtual functions in $L_2$ are functions in $L_2$ and all virtual functions in $L_{\infty}$ are functions in  $L_{\infty}$.
\label{thm: virt stab}
\end{thm}

Theorem \ref{thm: virt stab} makes it possible to focus on the virtual stability of every subsystem in lieu of ensuring the stability of the entire robot.

\section{Kinematics and dynamics of a generic manipulator structure}

\label{sec: kinematics and dynamics analysis}

\begin{figure}[h!]
\centering
\includegraphics[width=.8\textwidth]{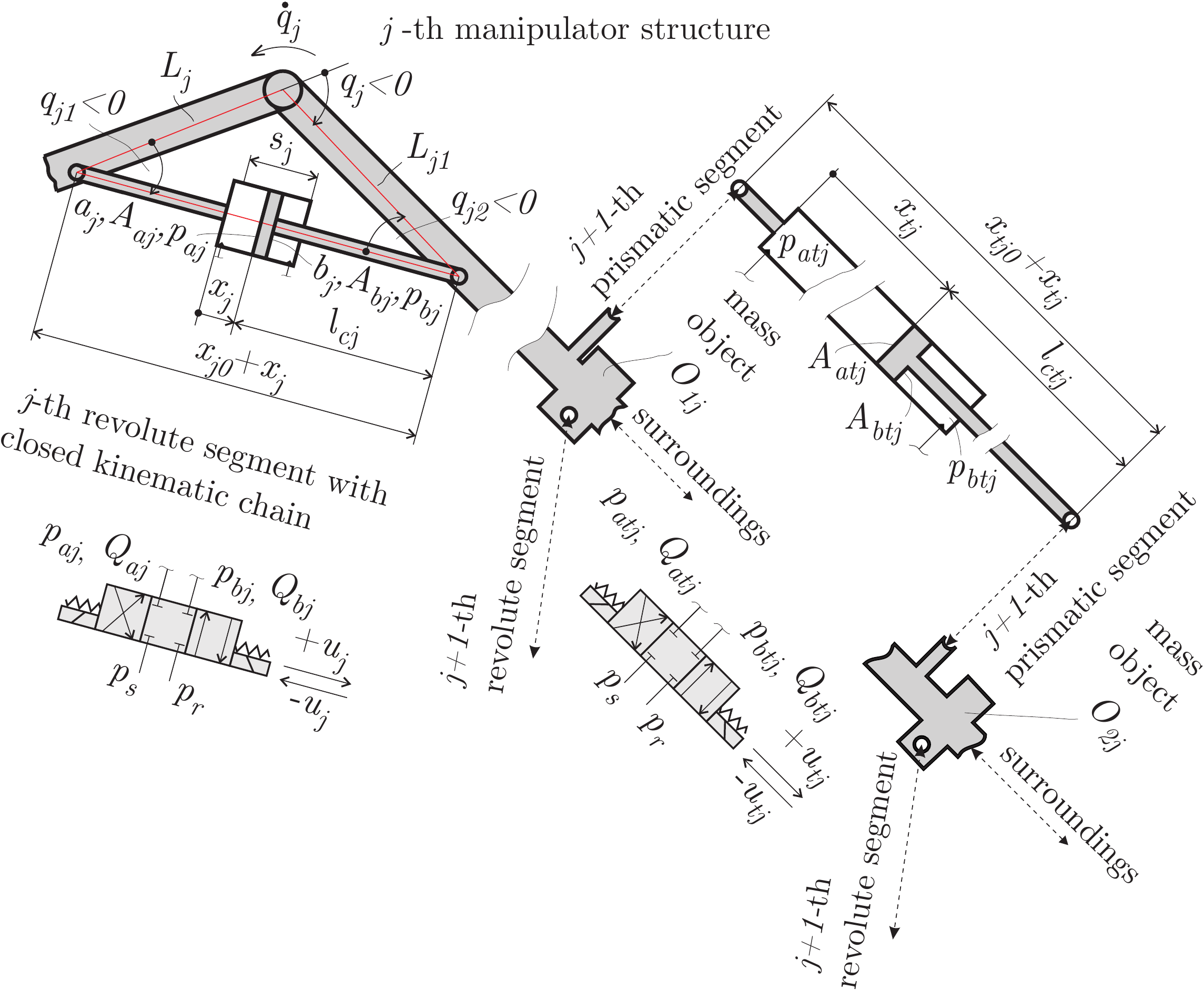}
\caption{Generic $j$-th manipulator structure.} \label{fig: closed chain}
\end{figure}

The following analysis considers the $j$-th generic manipulator structure, as shown in Fig. \ref{fig: closed chain}. It consists of a revolute and a prismatic segment. Without loss of generality, it can be assumed that the prismatic segment has only one linear hydraulic actuator for clarity.

The hydraulic manipulator consists of $n$ structures like this in series, whereas prismatic segments might not exist in every structure, depending on the analysed configuration.

\subsection{Virtual decomposition}

In the analysis of the closed kinematic loop, virtual cutting can be performed, and open kinematic tree structures are obtained as the first step \cite{murray1989dynamic}, \cite{luh1985computation} and \cite{lin1990dynamics}. Virtual cutting points are also significant in the VDC approach as separation interfaces that conceptually cut through a rigid body.  

Parts that come from the VCP maintain equal positions and orientations. Furthermore, at a VCP, force/moment vectors can be exerted from one part to another. Every VCP is considered a driving VCP for a subsystem from which the force/moment vector is exerted, and it is considered a driven VCP for another subsystem to which the force/moment vector is exerted \cite{zhu2010virtual}.

\begin{figure}[h!]
\centering
\includegraphics[width= .8\textwidth]{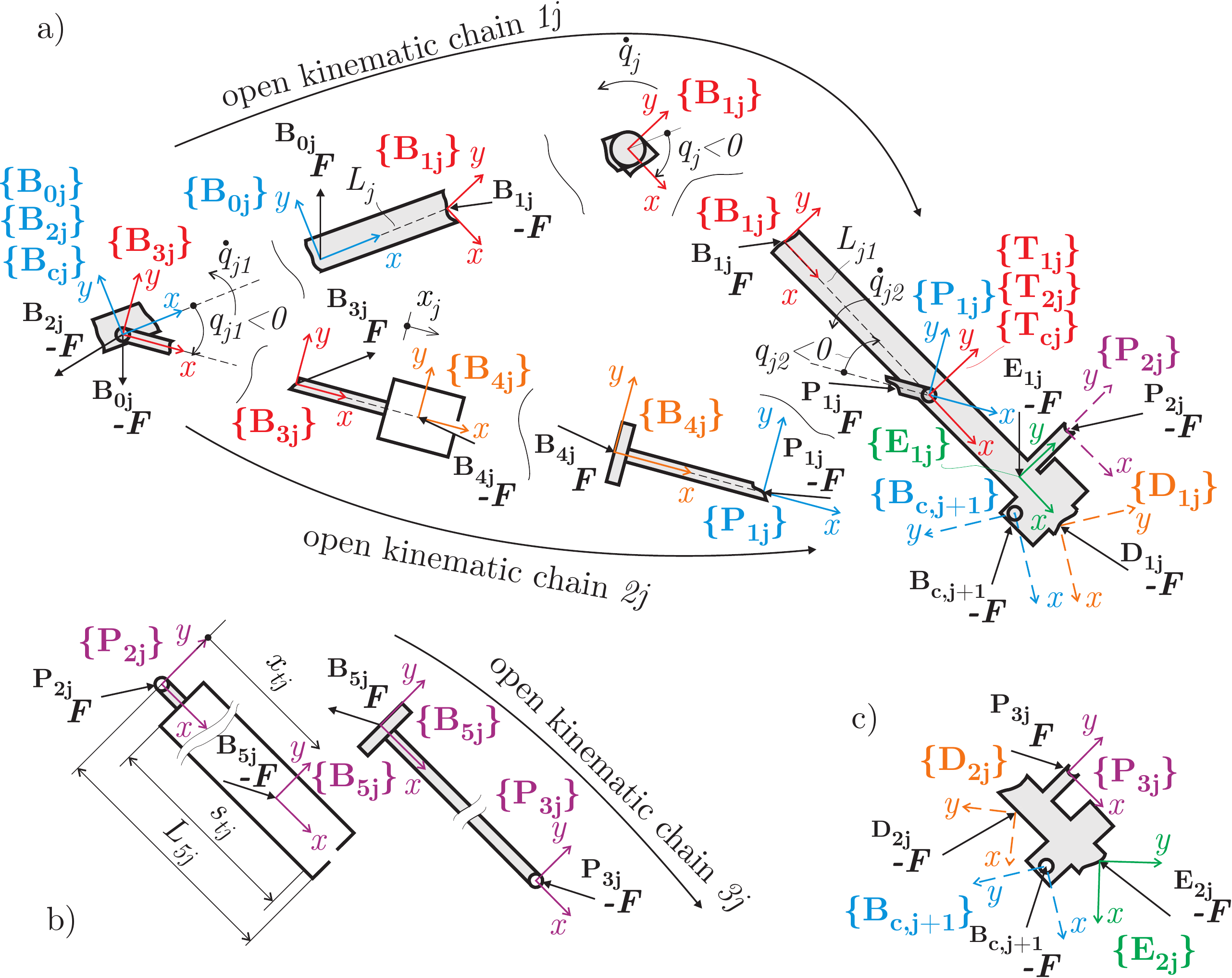}
\caption{Virtually decomposed manipulator structure.}
\label{fig: decomposed chain}
\end{figure}

The mass object $O_{1j}$ and the link $L_{j1}$ from Fig. \ref{fig: closed chain} are treated as joined in the revolute segment decomposition, contrasting all the prevailing approaches. The proposed decomposition is shown in more detail in Figure \ref{fig: decomposed chain}a.   

After virtually cutting the closed kinematic chain, the open kinematic chain $1j$ has a passive revolute joint, and the open kinematic chain $2j$ contains a linear hydraulic actuator. The structure can now be presented as a set of modular subsystems. The optional subsystems that originate from the prismatic segment are shown in Fig. \ref{fig: decomposed chain}b and  Fig. \ref{fig: decomposed chain}c. They are treated as in the VDC mainstream. 

The optional existence of the prismatic segment also introduces the open kinematic chain $3j$, which contains a linear hydraulic actuator, into the analysis. 

The division of a manipulator segment into subsystems by virtual decomposition creates a demand for various well-defined body-fixed frames so that all linear/angular velocity and force/moment relations can be established. 

Different frames are used here to describe the motion of every subsystem obtained by virtual decomposition. To analyse the $j$-th revolute segment kinematics and dynamics, 14 body-fixed frames $\left\lbrace \bf B_{cj} \right\rbrace$, $\left\lbrace \bf B_{0j} \right\rbrace$, $\left\lbrace \bf B_{2j} \right\rbrace$, $\left\lbrace \bf B_{1j} \right\rbrace$, $\left\lbrace \bf B_{3j} \right\rbrace$, $\left\lbrace \bf B_{4j} \right\rbrace$,   $\left\lbrace \bf T_{cj} \right\rbrace$, $\left\lbrace \bf T_{1j} \right\rbrace$, $\left\lbrace \bf T_{2j} \right\rbrace$,  $\left\lbrace \bf P_{1j} \right\rbrace$, $\left\lbrace \bf P_{2j} \right\rbrace$, $\left\lbrace \bf B_{c,j+1} \right\rbrace$, $\left\lbrace \bf D_{1j} \right\rbrace$ and $\left\lbrace \bf E_{1j} \right\rbrace$  are introduced, and they are all attached, as shown in Fig. \ref{fig: decomposed chain}a. If a prismatic segment exists in the $j$-th segment, additional frames $\left\lbrace \bf B_{5j} \right\rbrace$, $\left\lbrace \bf P_{3j} \right\rbrace$, $\left\lbrace \bf D_{2j} \right\rbrace$ and $\left\lbrace \bf E_{2j} \right\rbrace$ will be needed to account for kinematics and dynamics of that linear hydraulic actuator and mass object $O_{2j}$.  They are shown in Figs. \ref{fig: decomposed chain}b and \ref{fig: decomposed chain}c. It is important to note here that some frames are just configuration-dependent auxiliary frames, introducing generality into the analysis.

\begin{figure}[h!]
\centering
\includegraphics[width=.8\textwidth]{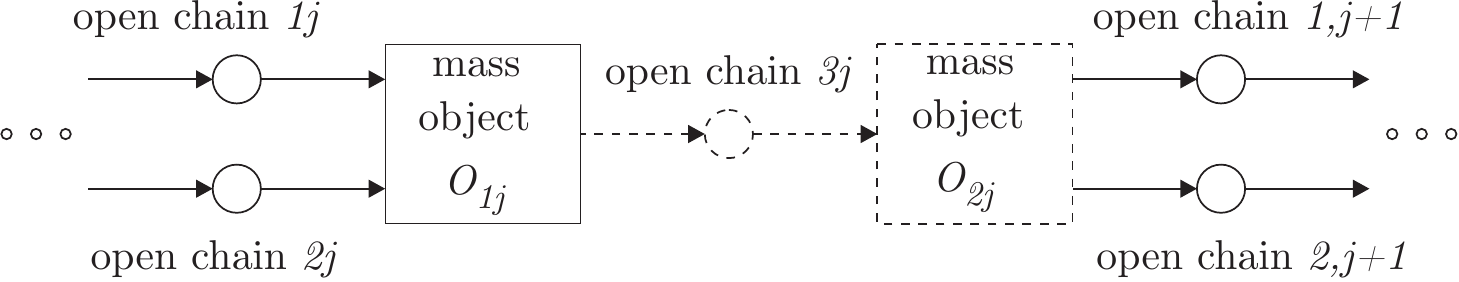}
\caption{Simple oriented graph of the manipulator structure.}
\label{fig: SOG}
\end{figure}

For a start, frame $\left\lbrace \bf E_{1j} \right\rbrace$ in  Fig. \ref{fig: decomposed chain}a coincides with one of three frames shown in dashed lines: $\left\lbrace \bf B_{c,j+1} \right\rbrace$, $\left\lbrace \bf P_{2j} \right\rbrace$ or $\left\lbrace \bf D_{1j} \right\rbrace$. The other two frames shown with dashed lines will  not exist, depending on manipulator configuration, as will be discussed. The same applies to frame $\left\lbrace \bf E_{2j} \right\rbrace$ in relation to frames $\left\lbrace \bf B_{c,j+1} \right\rbrace$ and $\left\lbrace \bf D_{2j} \right\rbrace$ in Fig. \ref{fig: decomposed chain}c. Some of the frames will mainly serve in the result derivation process, so extreme care about them may not be needed once all necessary relations are derived, and a new algorithmic approach to VDC deployment is established.  A simple oriented graph is shown in Fig. \ref{fig: SOG}, where an optional prismatic segment is shown with dashed lines.

It is worth emphasising that frames at the driven point of the revolute segment are coincident (see Fig. \ref{fig: decomposed chain}a):
\begin{equation}
\left\lbrace \bf B_{cj} \right\rbrace = \left\lbrace \bf B_{0j} \right\rbrace = \left\lbrace \bf B_{2j} \right\rbrace,
\label{eqn: Bcj is B0j is B2j}
\end{equation}
as at the revolute segment driving point, frames $\left\lbrace \bf T_{cj} \right\rbrace$, $\left\lbrace \bf T_{1j} \right\rbrace$ and $\left\lbrace \bf T_{2j} \right\rbrace$ coincide:
\begin{equation}
\left\lbrace \bf T_{cj} \right\rbrace = \left\lbrace \bf T_{1j} \right\rbrace = \left\lbrace \bf T_{2j} \right\rbrace.
\label{eqn: Tcj is T1j is T2j}
\end{equation}

The $x$-$y$ plane of every frame is located at the plane defined by three revolute joints. Their angles reference passive revolute joints as $q_j$-, $q_{j1}$- and $q_{j2}$-joint.

The origins of frames $\left\lbrace \bf B_{cj} \right\rbrace = \left\lbrace \bf B_{0j} \right\rbrace = \left\lbrace \bf B_{2j} \right\rbrace$ are located at the $q_{j1}$-joint, and their $x$-axes point to the $q_j$-joint. Then, the frame $\left\lbrace \bf B_{1j} \right\rbrace$ origin is located at the  $q_j$-joint, and its $x$-axis points to the $q_{j2}$-joint. Origins of frames $\left\lbrace \bf T_{cj} \right\rbrace = \left\lbrace \bf T_{1j} \right\rbrace = \left\lbrace \bf T_{2j} \right\rbrace$ are located at the  $q_{j2}$-joint, and they have the same orientation as frame $\left\lbrace \bf B_{1j} \right\rbrace$. The origin of frame $\left\lbrace \bf B_{3j} \right\rbrace$ is located at the $q_{j1}$-joint, and its $x$-axis points to the $q_{j2}$-joint. The origin of the $\left\lbrace \bf B_{4j} \right\rbrace$ frame is at the end of the linear hydraulic actuator piston, and its $x$-axis points to the $q_{j2}$-joint. The last frame of the open kinematic chain $2j$ is $\left\lbrace \bf P_{1j} \right\rbrace$, located at the $q_{j2}$-joint with the same orientation as the $\left\lbrace \bf B_{4j} \right\rbrace$ frame. Frame $\left\lbrace \bf P_{2j} \right\rbrace$ has its origin where the driven point of a prismatic segment and the manipulator link connect.  The action of the surroundings on the manipulator is expressed in the frame $\left\lbrace \bf D_{1j} \right\rbrace$.

Similarly, as shown in Fig. \ref{fig: decomposed chain}c, frame $\left\lbrace \bf P_{3j} \right\rbrace$ has its origin where the prismatic segment driving point and the manipulator link that follows connect. 
Frame $\left\lbrace \bf D_{2j} \right\rbrace$ expresses the effect of surroundings on the manipulator imposed on the link that comes after the linear actuator. Since the manipulator usually interacts with its surroundings only using the end-effector, frames $\left\lbrace \bf D_{1j} \right\rbrace$ and $\left\lbrace \bf D_{2j} \right\rbrace$, $j = 1,...,n-1$ can be removed from consideration. Depending on the manipulator configuration, i.e. whether the end-effector is connected to the link that precedes or follows a prismatic segment, only frame $\left\lbrace \bf D_{1n} \right\rbrace$ or frame $\left\lbrace \bf D_{2n} \right\rbrace$ exists, and they are mutually exclusive.

\begin{assumption}
A manipulator interacts with its surroundings using its end-effector.
\label{ass: end effector}
\end{assumption}

\begin{corollary}
Only frame $\left\lbrace \bf D_{1n} \right\rbrace$ or frame $\left\lbrace \bf D_{2n} \right\rbrace$ exists on the manipulator.
\end{corollary}

Recall that the notion of the end-effector implies that there are no manipulator links after it in the manipulator serial chain. Therefore, the existence of frame $\left\lbrace \bf D_{1n} \right\rbrace$ excludes the existence of frames $\left\lbrace \bf B_{c,n+1} \right\rbrace$ and $\left\lbrace \bf P_{2n} \right\rbrace$. The same applies to frames $\left\lbrace \bf D_{2n} \right\rbrace$ and $\left\lbrace \bf B_{c,n+1} \right\rbrace$. Frame $\left\lbrace \bf B_{c,j+1} \right\rbrace$ coincides with the driven point of the $(j+1)$-th revolute segment. In practice, it is not likely to encounter the prismatic joint driven point on the same link where there exists a revolute segment driven connection. This leads to the following assumption, considering manipulator configurations:

\begin{assumption}
The driven point of a revolute segment and the driven point of a prismatic segment are never connected to the same link.
\label{ass: tele closed}
\end{assumption}

\begin{corollary}
The existence of the frame $\left\lbrace \bf B_{c,j+1} \right\rbrace$ excludes the existence of the frame $\left\lbrace \bf P_{2j} \right\rbrace$ and vice versa.
\end{corollary}

As noted, frames $\left\lbrace \bf E_{1j} \right\rbrace$ and $\left\lbrace \bf E_{2j} \right\rbrace$ are introduced for the sake of  generality. Based on Fig. \ref{fig: decomposed chain}, Assumption \ref{ass: end effector}, and Assumption \ref{ass: tele closed}, frame 	${\left\lbrace{\bf E_{1j}}\right\rbrace}$ can be:
\begin{equation}
{\left\lbrace {\bf E_{1j}} \right\rbrace} = \left\lbrace \begin{array}{cc}
	{\left\lbrace{\bf B_{c,j+1}}\right\rbrace}, & \text{if a revolute segment follows, going from base to tip,}\\
	{\left\lbrace{\bf P_{2j}}\right\rbrace}, & \text{if a prismatic segment follows, going from base to tip,} \\
	{\left\lbrace{\bf D_{1n}}\right\rbrace}, & \text{if an end-effector is at the end of a link.}
\end{array} \right.
\label{eqn: frame E1j}
\end{equation}
Moreover, frame ${\left\lbrace {\bf E_{2j}} \right\rbrace}$ can be one of the following:
\begin{equation}
{\left\lbrace {\bf E_{2j}} \right\rbrace} = \left\lbrace \begin{array}{cc}
	{\left\lbrace{\bf B_{c,j+1}}\right\rbrace}, & \text{if a revolute segment follows, going from base to tip,}\\
	{\left\lbrace{\bf D_{2n}}\right\rbrace}, & \text{if an end-effector is at the end of a link.}
\end{array} \right.
\label{eqn: frame E2j}
\end{equation}

Relations given by Eqs. (\ref{eqn: frame E1j}) and (\ref{eqn: frame E2j}) play a significant role in the general algorithm.

\subsection{Kinematic relations}

In VDC, kinematic relations are used primarily to formulate the total forces needed to act on rigid bodies, based on the required velocities. Before carrying out the kinematic analysis of the $j$-th manipulator structure, important geometric relations for the $j$-th closed chain, representing constraints, must be introduced.

\subsubsection{Loop-closure functions}

Constraints must be introduced to accompany the virtual cutting approach in a closed kinematic loop analysis. 
Here, analytic loop-closure functions can be incorporated into the analysis.

The joint angle of every revolute segment must be related to the corresponding piston displacement and the other two closed-chain angles $q_{j1}$ and $q_{j2}$, using known lengths.

Referring to Fig. \ref{fig: closed chain} and considering the introduced sign conventions, these relations are:
\begin{equation}
x_j = \sqrt{L_j^2 + L_{j1}^2 + 2 \, L_j \, L_{j1} \, \cos q_j } - x_{j0},
\label{eqn: xj}
\end{equation}
\begin{equation}
q_{j1} = -\arccos \left( \dfrac{L_{j1}^2 - (x_j + x_{j0})^2 - L_{j}^2}{-2 \, (x_j + x_{j0}) \, L_j} \right),
\label{eqn: qj1}
\end{equation}
\begin{equation}
q_{j2} = -\arccos \left( \dfrac{L_{j}^2 - (x_j + x_{j0})^2 - L_{j1}^2}{-2 \, (x_j + x_{j0}) \, L_{j1}} \right),
\label{eqn: qj2}
\end{equation}
where $L_{j}$ and $L_{j1}$ are lengths between the frames, and $x_{j0}$ is the effective length of the hydraulic linear actuator at zero piston stroke. These values are assumed to be known with a high degree of accuracy. Differentiating Eqs. (\ref{eqn: xj})–(\ref{eqn: qj2}) and appropriately transforming them, the following expressions for velocities hold:
\begin{equation}
\dot{x}_j = - \dfrac{L_j \, L_{j1} \, \sin q_{j}}{x_j + x_{j0}} \,  \dot{q}_j,
\label{eqn: dxj}
\end{equation}
\begin{equation}
\dot{q}_{j1} = - \dfrac{(x_j + x_{j0}) - L_j \, \cos q_{j1}}{(x_j + x_{j0}) \, L_j \, \sin q_{j1}} \, \dot{x}_j,
\label{eqn: dqj1}
\end{equation}
\begin{equation}
\dot{q}_{j2} = - \dfrac{(x_j + x_{j0}) - L_{j1} \, \cos q_{j2}}{(x_j + x_{j0}) \, L_{j1} \, \sin q_{j2}} \, \dot{x}_j.
\label{eqn: dqj2}
\end{equation}

\subsubsection{Subsystem velocities}

Let the linear/angular velocity vector ${^{\bf B_{cj}} \boldsymbol{V}}$ be known from recursive calculations carried out through preceding subsystems. From Eq. (\ref{eqn: Bcj is B0j is B2j}), and Fig. \ref{fig: decomposed chain}a) it follows that:
\begin{equation}
{^{\bf B_{cj}}\boldsymbol{V}} = {^{\bf B_{0j}}\boldsymbol{V}} = {^{\bf B_{2j}}\boldsymbol{V}}.
\label{eqn: BV}
\end{equation}

Linear/angular velocities in the open kinematic chain $1j$ can be written as:
\begin{equation}
{^{\bf B_{1j}}\mathbf{V}} = \mathbf{z}_{\tau} \, \dot{q}_j + {^{\bf B_{0j}}\mathbf{U}_{\mathbf{B_{1j}}}^T} \, {^{\bf B_{0j}}\boldsymbol{V}},
\label{eqn: B1jV}
\end{equation}
and
\begin{equation}
{^{\bf T_{1j}}\boldsymbol{V}} = {^{\bf B_{1j}}\mathbf{U}_{\mathbf{T_{1j}}}^T} \, {^{\bf B_{1j}}\boldsymbol{V}},
\end{equation}
where $\mathbf{z}_{\tau} = \begin{pmatrix}
0 & 0 & 0 & 0 & 0 & 1
\end{pmatrix}^T$.

Similarly, linear/angular velocities in the open kinematic chain $2j$ are:
\begin{equation}
{^{\bf B_{3j}}\boldsymbol{V}} = \mathbf{z}_\tau \, \dot{q}_{j1} + {^{\bf B_{2j}}\mathbf{U}_{\mathbf{B_{3j}}}^T} \, {^{\bf B_{2j}}\boldsymbol{V}},
\label{eqn: B3jV}
\end{equation}
\begin{equation}
{^{\bf B_{4j}}\boldsymbol{V}} = \mathbf{x}_f \, \dot{x}_{j} + {^{\bf B_{3j}}\mathbf{U}_{\mathbf{B_{4j}}}^T} \, {^{\bf B_{3j}}\boldsymbol{V}},  
\label{eqn: B4jV}
\end{equation}
and
\begin{equation}
{^{\bf T_{2j}}\boldsymbol{V}} = \mathbf{z}_\tau \, \dot{q}_{j2} +  {^{\bf B_{4j}}\mathbf{U}_{\mathbf{T_{2j}}}^T} \, {^{\bf B_{4j}}\boldsymbol{V}},
\end{equation}
where  $\mathbf{x}_f = \begin{pmatrix}
1 & 0 & 0 & 0 & 0 & 0
\end{pmatrix}^T$.

Per Eq. (\ref{eqn: Tcj is T1j is T2j}), linear/angular velocities at the driving VCP of the closed kinematic chain are:
\begin{equation}
{^{\bf T_{cj}}\boldsymbol{V}}  = {^{\bf T_{1j}}\boldsymbol{V}}  = {^{\bf T_{2j}}\boldsymbol{V}}.
\end{equation}

Finally, linear/angular velocities measured and expressed in frame $\left\lbrace \bf E_{1j} \right\rbrace$ can be calculated as:
\begin{equation}
{^{\bf E_{1j}}\boldsymbol{V}} = {^{\bf T_{cj}}\mathbf{U}_{\mathbf{E_{1j}}}^T} \, {^{\bf T_{cj}}\boldsymbol{V}}.
\label{eqn: E1jV}
\end{equation}

When a prismatic joint exists in the considered $j$-th manipulator structure, additional kinematic relations have to be established. In the case of a prismatic joint existence, Eq. (\ref{eqn: frame E1j}) holds that $\bf \left\lbrace E_{1j} \right\rbrace =  \left\lbrace P_{2j} \right\rbrace$, and Eq. (\ref{eqn: E1jV}) reads:
\begin{equation}
{^{\bf P_{2j}}\boldsymbol{V}} = {^{\bf T_{cj}}\mathbf{U}_{\mathbf{P_{2j}}}^T} \, {^{\bf T_{cj}}\boldsymbol{V}}.
\label{eqn: P2jV}
\end{equation}

Further kinematic relations for a linear hydraulic actuator per Fig. \ref{fig: decomposed chain}b and Fig. \ref{fig: decomposed chain}c can  be written as:
\begin{equation}
{^{\bf B_{5j}}\boldsymbol{V}} = \mathbf{x}_f \, \dot{x}_{tj}  + {^{\bf P_{2j}}\mathbf{U}_{\mathbf{B_{5j}}}^T} \, {^{\bf P_{2j}}\boldsymbol{V}},
\label{eqn: B5jV}
\end{equation}
and
\begin{equation}
{^{\bf P_{3j}}\boldsymbol{V}} =  {^{\bf B_{5j}}\mathbf{U}_{\mathbf{P_{3j}}}^T} \, {^{\bf B_{5j}}\boldsymbol{V}}.
\label{eqn: P3jV}
\end{equation}

\subsubsection{Forming required velocities}

It is shown in \cite{zhu2010virtual} that the position control can be performed through a velocity controller by incorporating a position error term into the required velocity. Once the desired velocities are calculated solving the inverse kinematics, the required velocities are formed, as already described in \ref{subsec: req vels}. For revolute segments, in the case of integrated piston position feedback, required velocities can be formed as:
\begin{eqnarray}
\dot{x}_{jr} = \dot{x}_{jd} + \lambda_{xj} \, (x_{jd} - x_j),
\label{eqn: xjr}
\end{eqnarray}
where $\lambda_{xj} \in \mathbb{R}$ is a positive constant.

By contrast, if the required piston velocities are known, the rotating joint velocities needed can be calculated using Eq. (\ref{eqn: dxj}) again. If the integrated rotating joint position feedback is implemented, required velocities in closed chains can be alternatively formed as:
\begin{eqnarray}
\dot{q}_{jr} = \dot{q}_{jd} + \lambda_{qj} \, (q_{jd} - q_j),
\label{eqn: qjr}
\end{eqnarray}
where $\lambda_{qj} \in \mathbb{R}$ is again a positive constant, and the required piston velocities can be inferred from Eq.  (\ref{eqn: dxj}). For prismatic joints with integrated piston position feedback, required velocities are formed as:
\begin{eqnarray}
\dot{x}_{tjr} = \dot{x}_{tjd} + \lambda_{tj} \, (x_{tjd} - x_{tj}),
\label{eqn: xtjr}
\end{eqnarray}
where $\lambda_{tj} \in \mathbb{R}$ is a positive constant. Knowing required velocities for pistons and joints enables forming the required velocities, and this is essential for control, since required forces are formed using these, as will be shown.

Reusing Eqs. (\ref{eqn: BV})–(\ref{eqn: E1jV}) with required joint velocities $\dot{q}_{jr}$ and piston velocities $\dot{x}_{tjr}$ gives an extensive set of equations to determine the required velocities:
\begin{equation}
{^{\bf B_{cj}}\boldsymbol{V}_r} = {^{\bf B_{0j}}\boldsymbol{V}_r} = {^{\bf B_{2j}}\boldsymbol{V}_r},
\label{eqn: BVr}
\end{equation}
\begin{equation}
{^{\bf B_{1j}}\mathbf{V}_r} = \mathbf{z}_{\tau} \, \dot{q}_{jr}+ {^{\bf B_{0j}}\mathbf{U}_{\mathbf{B_{1j}}}^T} \, {^{\bf B_{0j}}\boldsymbol{V}_r},
\label{eqn: B1jVr}
\end{equation}
\begin{equation}
{^{\bf T_{1j}}\boldsymbol{V}_r} = {^{\bf B_{1j}}\mathbf{U}_{\mathbf{T_{1j}}}^T} \, {^{\bf B_{1j}}\boldsymbol{V}_r},
\end{equation}
\begin{equation}
{^{\bf B_{3j}}\boldsymbol{V}_r} = \mathbf{z}_\tau \, \dot{q}_{j1r} + {^{\bf B_{2j}}\mathbf{U}_{\mathbf{B_{3j}}}^T} \, {^{\bf B_{2j}}\boldsymbol{V}_r},
\label{eqn: B3jVr}
\end{equation}
\begin{equation}
{^{\bf B_{4j}}\boldsymbol{V}_r} = \mathbf{x}_f \, \dot{x}_{jr} + {^{\bf B_{3j}}\mathbf{U}_{\mathbf{B_{4j}}}^T} \, {^{\bf B_{3j}}\boldsymbol{V}_r},
\label{eqn: B4jVr}
\end{equation}
\begin{equation}
{^{\bf T_{2j}}\boldsymbol{V}_r} = \mathbf{z}_\tau \, \dot{q}_{j2r} +  {^{\bf B_{4j}}\mathbf{U}_{\mathbf{T_{2j}}}^T} \, {^{\bf B_{4j}}\boldsymbol{V}_r},
\end{equation}
\begin{equation}
{^{\bf T_{cj}}\boldsymbol{V}_r}  = {^{\bf T_{1j}}\boldsymbol{V}_r}  = {^{\bf T_{2j}}\boldsymbol{V}_r}, 
\end{equation}
and
\begin{equation}
{^{\bf E_{1j}}\boldsymbol{V}_r} = {^{\bf T_{cj}}\mathbf{U}_{\mathbf{E_{1j}}}^T} \, {^{\bf T_{cj}}\boldsymbol{V}_r}.
\label{eqn: E1jVr}
\end{equation}

In the case where there exists a prismatic joint, $\bf \left\lbrace E_{1j} \right\rbrace =  \left\lbrace P_{2j} \right\rbrace$, and these expressions follow:
\begin{equation}
{^{\bf P_{2j}}\boldsymbol{V}_r} = {^{\bf T_{cj}}\mathbf{U}_{\mathbf{P_{2j}}}^T} \, {^{\bf T_{cj}}\boldsymbol{V}_r},
\label{eqn: P2jVr}
\end{equation}
\begin{equation}
{^{\bf B_{5j}}\boldsymbol{V}_r} = \mathbf{x}_f \, \dot{x}_{tjr}  + {^{\bf P_{2j}}\mathbf{U}_{\mathbf{B_{5j}}}^T} \, {^{\bf P_{2j}}\boldsymbol{V}_r}, 
\label{eqn: B5jVr}
\end{equation}
and
\begin{equation}
{^{\bf P_{3j}}\boldsymbol{V}_r} =  {^{\bf B_{5j}}\mathbf{U}_{\mathbf{P_{3j}}}^T} \, {^{\bf B_{5j}}\boldsymbol{V}_r}.
\label{eqn: P3jVr}
\end{equation}

\subsection{Dynamics relations}

The proposed reduced set of governing EOMs for the decomposed $j$-th manipulator structure, using the suggested virtual decomposition, differs from the one obtained following the VDC mainstream. Nevertheless, the calculations proposed here yield the same values of actuator forces as earlier approaches had.  Equation (\ref{eqn: tot force}) describes the motion of a rigid body with constant mass. It is used here repeatedly to describe the motion of several subsystems: both revolute segment links from Fig. \ref{fig: decomposed chain}a, both cylinder cases and pistons with a rod from Figs. \ref{fig: decomposed chain}a and \ref{fig: decomposed chain}b and the mass object from Fig. \ref{fig: decomposed chain}c. 
The total force acting on each subsystem from Fig. \ref{fig: decomposed chain} can be modelled as:
\begin{equation}
\mathbf{M}_{\bf A} \dfrac{\rm d}{\mathrm{d}t} \left( {^{\bf A}\boldsymbol{V}} \right) + \mathbf{C}_{\bf A} \left( {^{\bf A}\boldsymbol{\omega}}  \right) {^{\bf A}\boldsymbol{V}} + \mathbf{G}_{\bf A} = {^{\bf A} \boldsymbol{F}^*},
\label{eqn: tot force 2}
\end{equation}
where instead of a general frame ${\bf A}$, one of the distinct frames ${\bf B_{0j}}$, ${\bf B_{1j}}$, ${\bf B_{3j}}$, ${\bf B_{4j}}$, ${\bf P_{2j}}$, ${\bf B_{5j}}$ or ${\bf P_{3j}}$ is used for each separate subsystem. Velocities ${^{\bf B_{0j}}\boldsymbol{V}}$, ${^{\bf B_{1j}}\boldsymbol{V}}$, ${^{\bf B_{3j}}\boldsymbol{V}}$, ${^{\bf B_{4j}}\boldsymbol{V}}$, ${^{\bf P_{2j}}\boldsymbol{V}}$, ${^{\bf B_{5j}}\boldsymbol{V}}$ and ${^{\bf P_{3j}}\boldsymbol{V}}$ also are necessary. They are given with Eqs. (\ref{eqn: BV})–(\ref{eqn: P3jV}).

\subsubsection{Prismatic segment dynamics}

Let the force/moment vector ${^{\bf E_{2j}}\boldsymbol F}$ be known from previous recursive calculations through other subsystems. From both Eq. (\ref{eqn: frame E2j}) and Fig. \ref{fig: decomposed chain}c, it  follows that:
\begin{equation}
{^{\bf E_{2j}}\boldsymbol F} = \left\lbrace \begin{array}{cc}
	{^{\bf B_{c,j+1}}\boldsymbol F}, & \text{if a revolute segment preceded, going from tip to base,}\\
	{^{\bf D_{2n}}\boldsymbol F}, & \text{if an end-effector is at the end of a link.}
\end{array} \right.
\label{eqn: frame E2j F}
\end{equation}

The total force/moment acting on the mass object $O_{2j}$ can be expressed as:
\begin{equation}
{^{\bf P_{3j}}\boldsymbol F^*} = {^{\bf P_{3j}}\boldsymbol F} - {^{\bf P_{3j}}\mathbf{U_{E_{2j}}}} {^{\bf E_{2j}}\boldsymbol F}.
\label{eqn: O2jF}
\end{equation}

This enables the force/moment vector ${^{\bf P_{3j}}\boldsymbol F}$ to be calculated as:
\begin{equation}
{^{\bf P_{3j}}\boldsymbol F} = {^{\bf P_{3j}}\boldsymbol F^*} + {^{\bf P_{3j}}\mathbf{U_{E_{2j}}}} {^{\bf E_{2j}}\boldsymbol F}.
\label{eqn: P3jF}
\end{equation}

Next, the total force/moment acting on the  piston from Fig. \ref{fig: decomposed chain}b is:
\begin{equation}
{^{\bf B_{5j}}\boldsymbol F^*} = {^{\bf B_{5j}}\boldsymbol F} - {^{\bf B_{5j}}\mathbf{U_{P_{3j}}}} {^{\bf P_{3j}}\boldsymbol F}.
\end{equation}

Thus, force ${^{\bf B_{5j}}\boldsymbol F}$  can be expressed as:
\begin{equation}
{^{\bf B_{5j}}\boldsymbol F} = {^{\bf B_{5j}}\boldsymbol F^*} + {^{\bf B_{5j}}\mathbf{U_{P_{3j}}}} {^{\bf P_{3j}}\boldsymbol F}.
\end{equation}

Linear actuator force can now be readily calculated as:
\begin{equation}
f_{ctj} = \mathbf{x}_f^T \, {^{\bf B_{5j}}\boldsymbol F},
\end{equation}
and force ${^{\bf P_{2j}}\boldsymbol F}$, which propagates to the revolute segment, is:
\begin{equation}
{^{\bf P_{2j}}\boldsymbol F} = {^{\bf P_{2j}}\boldsymbol F^*} + {^{\bf P_{2j}}\mathbf{U_{B_{5j}}}} {^{\bf B_{5j}}\boldsymbol F}.
\label{eqn: P2jF}
\end{equation}

\subsubsection{Revolute segment dynamics}

As in the VDC mainstream, it is assumed that the main friction in this type of revolute segment occurs between the piston and cylinder case. Thus, other frictions will be neglected, and this yields the following assumption:
\begin{assumption}
Friction moments in all the rotating joints are equal to zero.
\label{ass: no friction in joints}
\end{assumption}

\begin{corollary}
The following expressions can be written and assumed to be valid:

\begin{equation}
	\mathbf{z}_{\tau}^T \, {^{\bf B_{1j}} \boldsymbol{F}} = 0,
	\label{eqn: fric qj}
\end{equation}	
\begin{equation}
	\mathbf{z}_{\tau}^T \, {^{\bf P_{1j}} \boldsymbol{F}} = 0,
	\label{eqn: fric qj2}
\end{equation}
\begin{equation}
	\mathbf{z}_{\tau}^T \, {^{\bf B_{3j}} \boldsymbol{F}} = 0.
	\label{eqn: fric qj1}
\end{equation}
\end{corollary}

The introduction of Assumption \ref{ass: no friction in joints} does not make this analysis less general. Rather, it simplifies it using practical claims.  If friction moments in particular joints cannot be neglected, they can be added to the analysis.

Let the force/moment vector ${^{\bf E_{1j}}\boldsymbol F}$ be known from previous recursive calculations through other subsystems. Again, from both Eq. (\ref{eqn: frame E1j}) and Fig. \ref{fig: decomposed chain}a, it  follows that:
\begin{equation}
{^{\bf E_{1j}}\boldsymbol F} = \left\lbrace \begin{array}{cc}
	{^{\bf B_{c,j+1}}\boldsymbol F}, & \text{if a revolute segment preceded,  going from tip to base,}\\
	{^{\bf P_{2j}}\boldsymbol F}, & \text{if a prismatic segment preceded,  going from tip to base,} \\
	{^{\bf D_{1n}}\boldsymbol F}, & \text{if an end-effector is at the end of a link.}
\end{array} \right.
\label{eqn: frame E1j F}
\end{equation}

The dynamics of  all subsystems using recursive calculations can be addressed starting from EOM for link $L_{j1}$:
\begin{equation}
{^{\bf B_{1j}}\boldsymbol F} = {^{\bf B_{1j}}\boldsymbol F^*} + {^{\bf B_{1j}}\mathbf{U_{E_{1j}}}}	\, {^{\bf {E_{1j}} }\boldsymbol F}	- {^{\bf B_{1j}}\mathbf{U_{P_{1j}}}}	\, {^{\bf P_{1j}}\boldsymbol F}.
\label{eqn: B1jF}
\end{equation}

The other dynamics equation for the open kinematic chain $1j$ is the EOM for link $L_j$:
\begin{equation}
{^{\bf B_{0j}}\boldsymbol F} = {^{\bf B_{0j}}\boldsymbol F^*} +   {^{\bf B_{0j}}\mathbf{U_{B_{1j}}}}	\, {^{\bf B_{1j}}\boldsymbol F}.
\label{eqn: B0jF}
\end{equation}

Similarly, EOMs for the open kinematic chain $2j$ are:
\begin{equation}
{^{\bf B_{4j}}\boldsymbol F} = {^{\bf B_{4j}}\boldsymbol F^*} + {^{\bf B_{4j}}\mathbf{U_{P_{1j}}}}	\, {^{\bf P_{1j}}\boldsymbol F},
\label{eqn: B4jF}
\end{equation}
\begin{equation}
{^{\bf B_{3j}}\boldsymbol F} = {^{\bf B_{3j}}\boldsymbol F^*} + {^{\bf B_{3j}}\mathbf{U_{B_{4j}}}}	\, {^{\bf B_{4j}}\boldsymbol F},
\label{eqn: B3jF}
\end{equation}
\begin{equation}
{^{\bf B_{2j}}\boldsymbol F} =  {^{\bf B_{2j}}\mathbf{U_{B_{3j}}}}	\, {^{\bf B_{3j}}\boldsymbol F},
\label{eqn: B2jF}
\end{equation}
and finally, the force/moment vector  at the driven VCP is:
\begin{equation}
{^{\bf B_{cj}}\boldsymbol F} = {^{\bf B_{0j}}\boldsymbol F} +  {^{\bf B_{2j}}\boldsymbol F}.
\label{eqn: BcjF}
\end{equation}

The actuator force is calculated as:
\begin{equation}
f_{cj} = \mathbf{x}_f^T \, {^{\bf B_{4j}}\boldsymbol F}.
\label{eqn: piston force in general}
\end{equation}
which is an essential expression, since it is directly used in the control law and stability analysis; thus, a straightforward solution is sought.

\begin{thm}
Let the force ${^{\bf E_{1j}}\boldsymbol F}$ acting on the revolute segment from Fig. \ref{fig: decomposed chain}a be known from previous calculations through other subsystems. Then, the linear actuator force can be determined as:
\begin{equation}
	\begin{array}{ll}
		f_{cj} =  &  \mathbf{x}_f^T \, {^{\mathbf{B_{4j}}}\boldsymbol{F}^*}     - \dfrac{\mathbf{z}_{\tau}^T \left( {^{\bf B_{1j}}\boldsymbol{F}^*} + {^{\bf B_{1j}}\mathbf{U_{E_{1j}}}}	\, {^{\bf E_{1j}}\boldsymbol{F}} \right)}{L_{j1} \, \sin q_{j2}}   - \\ \\
		& \dfrac{\mathbf{z}_{\tau}^T \, ({^{\mathbf{B_{3j}}}\boldsymbol{F}^*}) + \mathbf{z}_{\tau}^T \, ({^{\mathbf{B_{4j}}}\boldsymbol{F}^*}) + \mathbf{y}_{f}^T \left( {^{\mathbf{B_{4j}}}\boldsymbol{F}^*} \right) \, (x_j + x_{j0}  - l_{cj})}{(x_j + x_{j0} ) \, \tan q_{j2}}.
	\end{array}
	\label{eqn: piston force}
\end{equation}

\label{thm: piston force}
\end{thm}

\newpage

\begin{thm}
Let the force ${^{\bf E_{1j}}\boldsymbol F}$ acting on the revolute segment from Fig. \ref{fig: decomposed chain}a be known from previous calculations through other subsystems. Then, the total force acting on the driven point of the revolute segment is:
\begin{equation}
	\begin{array}{ll}
		{^{\bf B_{cj}}\boldsymbol F} = & {^{\bf B_{0j}}\boldsymbol F^*} + {^{\bf B_{0j}}\mathbf{U_{B_{1j}}}}  {^{\bf B_{1j}}\boldsymbol F^*}  + {^{\bf B_{2j}}\mathbf{U_{B_{3j}}}}  {^{\bf B_{3j}}\boldsymbol F^*} + \\ & {^{\bf B_{2j}}\mathbf{U_{B_{3j}}}}   {^{\bf B_{3j}}\mathbf{U_{B_{4j}}}} {^{\bf B_{4j}}\boldsymbol F^*} + {^{\bf B_{0j}}\mathbf{U_{B_{1j}}}} {^{\bf B_{1j}}\mathbf{U_{E_{1j}}}}	\, {^{\bf E_{1j}}\boldsymbol F}.
	\end{array}
	\label{eqn: driven cc force}
\end{equation}
\label{thm: driven cc force}
\end{thm}
The proofs of Theorem \ref{thm: piston force} and Theorem \ref{thm: driven cc force} are given in Appendix A and  Appendix B, respectively.

\subsubsection{Forming required force/moment vectors}

The required force/moment vector for every subsystem of the $j$-th manipulator structure can be easily obtained by replacing  ${\bf A}$ in Eq. (\ref{eqn: tot force}) with ${\bf B_{0j}}$, ${\bf B_{1j}}$, ${\bf B_{3j}}$, ${\bf B_{4j}}$, ${\bf P_{2j}}$, ${\bf B_{5j}}$ and ${\bf P_{3j}}$ for different subsystems. 

Velocities ${^{\bf B_{0j}}\boldsymbol{V}_r}$, ${^{\bf B_{1j}}\boldsymbol{V}_r}$, ${^{\bf B_{3j}}\boldsymbol{V}_r}$, ${^{\bf B_{4j}}\boldsymbol{V}_r}$, ${^{\bf P_{2j}}\boldsymbol{V}_r}$, ${^{\bf B_{5j}}\boldsymbol{V}_r}$ and ${^{\bf P_{3j}}\boldsymbol{V}_r}$ also are necessary, and they are obtained with series of Eqs. (\ref{eqn: BVr})–(\ref{eqn: P3jVr}).  If frame $\left\lbrace \bf E_{2j} \right\rbrace$ exists, then the required force/moment vector is deduced from Eq. (\ref{eqn: frame E2j F}) as:
\begin{equation}
{^{\bf E_{2j}}\boldsymbol{F}_r} = \left\lbrace \begin{array}{cc}
	{^{\bf B_{c,j+1}}\boldsymbol{F}_r}, & \text{ if a revolute segment preceded  going from tip to base,}\\
	{^{\bf D_{2n}}\boldsymbol{F}_r}, & \text{ if an end-effector is at the end of a link.}
\end{array} \right.
\label{eqn: frame E2j Fr}
\end{equation}

Then the following required force/moment vectors can be calculated:
\begin{equation}
{^{\bf P_{3j}}\boldsymbol F_r} =  {^{\bf P_{3j}}\boldsymbol F^*_r} + {^{\bf P_{3j}}\mathbf{U_{E_{2j}}}} {^{\bf E_{2j}}\boldsymbol F_r},
\label{eqn: P3jFr}
\end{equation}
\begin{equation}
{^{\bf B_{5j}}\boldsymbol F_r} = {^{\bf B_{5j}}\boldsymbol F^*_r} + {^{\bf B_{5j}}\mathbf{U_{P_{3j}}}} {^{\bf P_{3j}}\boldsymbol F_r},
\end{equation}
\begin{equation}
f_{ctjr} = \mathbf{x}_f^T \, {^{\bf B_{5j}}\boldsymbol F_r},
\end{equation}
\begin{equation}
{^{\bf P_{2j}}\boldsymbol F_r} = {^{\bf P_{2j}}\boldsymbol F^*_r} + {^{\bf P_{2j}}\mathbf{U_{B_{5j}}}} {^{\bf B_{5j}}\boldsymbol F_r}.
\label{eqn: frame P2jFr}
\end{equation}

Required force/moment vector ${^{\bf E_{1j}}\boldsymbol F}$ is easily deduced from Eq. (\ref{eqn: frame E1j Fr}) as:
\begin{equation}
{^{\bf E_{1j}}\boldsymbol F_r} = \left\lbrace \begin{array}{cc}
	{^{\bf B_{c,j+1}}\boldsymbol F_r}, & \text{if a revolute segment preceded, going from tip to base,}\\
	{^{\bf P_{2j}}\boldsymbol F_r}, & \text{if a prismatic segment preceded, going from tip to base,} \\
	{^{\bf D_{1n}}\boldsymbol F_r}, & \text{if an end-effector is at the end of a link.}
\end{array} \right.
\label{eqn: frame E1j Fr}
\end{equation}

One of the last essential results is a straightforward expression for the required linear actuator force calculation, that follows from Eq. (\ref{eqn: piston force}) as:
\begin{equation}
\begin{array}{ll}
	f_{cjr} =    & \mathbf{x}_f^T \, {^{\mathbf{B_{4j}}}\boldsymbol{F}^*_r}   - \dfrac{\mathbf{z}_{\tau}^T \left( {^{\bf B_{1j}}\boldsymbol{F}^*_r} + {^{\bf B_{1j}}\mathbf{U_{E_{1j}}}}	\, {^{\bf E_{1j}} \boldsymbol{F}_r} \right)}{L_{j1} \, \sin q_{j2}}   - \\ \\ &  \dfrac{\mathbf{z}_{\tau}^T \, \left({^{\mathbf{B_{3j}}}\boldsymbol{F}^*_r}\right) + \mathbf{z}_{\tau}^T \, \left({^{\mathbf{B_{4j}}}\boldsymbol{F}^*_r} \right) + \mathbf{y}_{f}^T \left( {^{\mathbf{B_{4j}}}\boldsymbol{F}^*_r} \right) \, (x_j + x_{j0}  - l_{cj})}{(x_j + x_{j0} ) \, \tan q_{j2}}.
\end{array}
\label{eqn: required piston force}
\end{equation}

Finally, the required force/moment vector acting on the driven point of a revolute segment also is obtained independent of any internal forces and moments as:
\begin{equation}
\begin{array}{ll}
	{^{\bf B_{cj}}\boldsymbol F_r} = & {^{\bf B_{0j}}\boldsymbol F^*_r} + {^{\bf B_{0j}}\mathbf{U_{B_{1j}}}}  {^{\bf B_{1j}}\boldsymbol F^*_r}  + {^{\bf B_{2j}}\mathbf{U_{B_{3j}}}}  {^{\bf B_{3j}}\boldsymbol F^*_r} +\\ & {^{\bf B_{2j}}\mathbf{U_{B_{3j}}}}   {^{\bf B_{3j}}\mathbf{U_{B_{4j}}}} {^{\bf B_{4j}}\boldsymbol F^*_r} + {^{\bf B_{0j}}\mathbf{U_{B_{1j}}}} {^{\bf B_{1j}}\mathbf{U_{E_{1j}}}}	\, {^{\bf E_{1j}}\boldsymbol F_r}.
\end{array}
\label{eqn: required cc force}
\end{equation}
This vector is used for calculations related to the revolute segment. This concludes the kinematics and dynamics analysis.

\subsection{Pressure and spool valve dynamics}

Both pressure and spool valve dynamics play an essential role in forming control action and securing the stability of the whole system. The same analysis shown for the linear hydraulic actuator in the $j$-th revolute segment can be applied to the linear hydraulic actuator driving the $j$-th prismatic joint, and it will not be repeated. It is only necessary to note the differences in subscripts for these two in the later analysis.

Piston force in the $j$-th revolute segment calculated using Eq. (\ref{eqn: piston force}) does not include friction contribution. The friction model, if assumed to be increasing, continuous and antisymmetric, accords with many types of friction encountered in practice, including Coulomb, viscous and LuGre friction (see \cite{zhu2013precision}). The piston force from Eq. (\ref{eqn: piston force}) with added friction term $f_{fj}$ can be written as:
\begin{equation}
f_{pj} = f_{cj} + f_{fj}.
\label{eqn: fp1}
\end{equation}

Piston force, Eq. (\ref{eqn: fp1}), also can be determined from  chamber pressures:
\begin{equation}
f_{pj} = A_{aj} \, p_{aj} - A_{bj} \, p_{bj},
\label{eqn: fp2}
\end{equation}
where $A_{aj}$ and $A_{bj}$ denote cross-sectional areas, while $p_{aj}$ and $p_{bj}$ are pressures in linear hydraulic actuator cylinder chambers, all shown in Fig. \ref{fig: closed chain}.

Continuity equations written for linear hydraulic actuators describe  pressure changes in both cylinder chambers as:
\begin{equation}
\dot{p}_{aj} = \dfrac{\beta}{A_{aj}} \left( \dfrac{Q_{aj}}{x_j} - A_{aj} \, \dfrac{\dot{x}_j}{x_j} \right),
\label{eqn: dpaj}
\end{equation}
and
\begin{equation}
\dot{p}_{bj} = \dfrac{\beta}{A_{bj}} \left( \dfrac{Q_{bj}}{s_j - x_j} + A_{bj} \, \dfrac{\dot{x}_j}{s_j - x_j} \right),
\label{eqn: dpbj}
\end{equation}
where $\beta$ denotes the oil bulk modulus. 

\begin{assumption}
Piston positions $x_{j}$ and $x_{tj}$ in linear hydraulic actuators never reach their  limiting positions at cylinder ends.
\label{ass: end positions}
\end{assumption}

\begin{corollary}
The following inequalities hold: $ 0 < x_{j} < s_j$, $ 0 < x_{tj} < s_{tj}$.
\end{corollary}

Assumption \ref{ass: end positions} on piston positions ensures that singularities are avoided in Eqs. (\ref{eqn: dpaj}) and (\ref{eqn: dpbj}). Their validity is ensured by careful trajectory planning. These assumptions can also be removed by investing additional modelling effort to model dead volumes and leakage flows, as shown in \cite{lampinen2019improved}.
Volumetric flows through orifices of the $j$-th spool valve in Fig. \ref{fig: closed chain}, which controls the flow to linear hydraulic actuator chambers in the revolute segment, are:
\begin{equation}
Q_{aj} = c_{p1j} \, \upsilon(p_s - p_{aj}) \, u_j \, S(u_j) + c_{n1j} \, \upsilon(p_{aj} - p_{r}) \, u_j \, S(-u_j), 
\label{eqn: Qaj}
\end{equation}
and
\begin{equation}
Q_{bj} =  - c_{n2j} \, \upsilon(p_{bj} - p_{r}) \, u_j \, S(u_j)-c_{p2j} \, \upsilon(p_s - p_{bj}) \, u_j \, S(-u_j),
\label{eqn: Qbj}
\end{equation}
where $c_{p1j}$, $c_{p2j}$, $c_{n1j}$  and $c_{n2j}$ are flow coefficients, $p_s$ is the supply pressure, $p_r$ is the return-line pressure, and $S(u_j)$ is the selection function defined as:
\begin{eqnarray}
S(u_j) = \left\lbrace \begin{array}{ll}
	1, & \text{if} \, \quad u_j > 0, \\
	0, & \text{if} \, \quad u_j \leqslant 0, \\
\end{array} \right.
\end{eqnarray}
\noindent and $\upsilon(\Delta p_j)$ is the function related to drops in pressure:
\begin{eqnarray}
\upsilon(\Delta p_j) = \text{sign}(\Delta p_j)  \, \sqrt{\left\lvert \Delta p_j \right\rvert}.
\end{eqnarray}

A voltage-related term that is significant for further analysis is defined as:
\begin{eqnarray}
u_{fj} = \dfrac{Q_{aj}}{x_j} - \dfrac{Q_{bj}}{s_j - x_j},
\label{eqn: ufj def}
\end{eqnarray}
and it lets Eq. (\ref{eqn: fp2}) be rewritten, using Eq. (\ref{eqn: dpaj}) and (\ref{eqn: dpbj}), as:
\begin{eqnarray}
\dot{f}_{pj} = \beta \left( u_{fj} - \left( \dfrac{A_{aj}}{x_j} + \dfrac{A_{bj}}{s_j - x_j} \right) \, \dot{x}_j \right).
\label{eqn: fpj with ufj}
\end{eqnarray}

The voltage-related term $u_{fj}$, Eq. (\ref{eqn: ufj def}), can be expressed from Eq. (\ref{eqn: fpj with ufj}) as:
\begin{equation}
u_{fj} = \dfrac{\dot{f}_{pj}}{\beta} + \left( \dfrac{A_{aj}}{x_j} + \dfrac{A_{bj}}{s_j - x_j} \right) \, \dot{x}_j.
\label{eqn: ufj}
\end{equation}

\begin{assumption}
Pressures in linear hydraulic actuator chambers are always smaller than the supply pressure, and they are always higher than the return line pressure, which is never zero.
\label{ass: pressures}
\end{assumption}

\begin{corollary}
The following inequalities for pressures hold: $p_s > p_{aj} > p_r \geqslant 0$,  $p_s > p_{bj} > p_r \geqslant 0$, $p_s > p_{atj} > p_r \geqslant 0$,  $p_s > p_{btj} > p_r \geqslant 0$.
\end{corollary}

Assumption \ref{ass: pressures} provides univalence between $u_j$ and $u_{fj}$. Combining Assumption \ref{ass: pressures} with Eqs. (\ref{eqn: Qaj}) and (\ref{eqn: Qbj}), the spool valve voltage can be expressed as:
\begin{equation}
\begin{array}{cc}
	u_{j}  = & \dfrac{u_{fj} \, S(u_{fj})}{\left( \dfrac{c_{p1j} \, \upsilon(p_s - p_{aj})}{x_j} + \dfrac{c_{n2j} \, \upsilon(p_{bj} - p_r)}{s_j - x_j} \right)}  + \\ & \dfrac{u_{fj} \, S(-u_{fj})}{\left( \dfrac{c_{p2j} \, \upsilon(p_s - p_{bj})}{s_j - x_j} + \dfrac{c_{n1j} \, \upsilon(p_{aj} - p_r)}{x_j} \right)}.
	\label{eqn: uj}
\end{array}	
\end{equation}

\section{Forming the control action}

\label{sec: control action forming}

A control action is formed so that system stability is ensured and all values converge to their respective required levels. The required value for the voltage-related term from Eq. (\ref{eqn: ufj}) for a revolute segment is formed as:
\begin{equation}
\begin{array}{cc}
	u_{fjr} = & \dfrac{\dot{f}_{pjr}}{\beta} + \left( \dfrac{A_{aj}}{x_j} + \dfrac{A_{bj}}{s_j - x_j} \right) \, \dot{x}_j + \\ & k_{xj} \, (\dot{x}_{jr} - \dot{x}_j) + k_{fj} \, (f_{pjr} - f_{pj}),
\end{array}
\label{eqn: ufjr}
\end{equation}
and the control voltage is calculated as:
\begin{equation}
\begin{array}{cc}
	u_{j} = & \dfrac{u_{fjr} \, S(u_{fjr})}{\left( \dfrac{c_{p1j} \, \upsilon(p_s - p_{aj})}{x_j} + \dfrac{c_{n2j} \, \upsilon(p_{bj} - p_r)}{s_j - x_j} \right)}  + \\ & \dfrac{u_{fjr} \, S(-u_{fjr})}{\left( \dfrac{c_{p2j} \, \upsilon(p_s - p_{bj})}{s_j - x_j} + \dfrac{c_{n1j} \, \upsilon(p_{aj} - p_r)}{x_j} \right)}.
\end{array}
\label{eqn: uj2}
\end{equation}

For a prismatic segment, the required voltage-related term Eq. (\ref{eqn: ufj}) is:
\begin{equation}
\begin{array}{cc}
	u_{ftjr}  = & \dfrac{\dot{f}_{ptjr}}{\beta} + \left( \dfrac{A_{atj}}{x_{tj}} + \dfrac{A_{btj}}{s_{tj} - x_{tj}} \right) \, \dot{x}_{tj} + \\ & k_{xtj} \, (\dot{x}_{tjr} - \dot{x}_{tj}) + k_{ftj} \, (f_{ptjr} - f_{ptj}),
\end{array}	
\label{eqn: uftrjj}
\end{equation}
and the control voltage is:
\begin{equation}
\begin{array}{cc}
	u_{tj} = & \dfrac{u_{ftjr} \, S(u_{ftjr})}{\left( \dfrac{c_{p1tj} \upsilon(p_s - p_{atj})}{x_{tj}} + \dfrac{c_{n2tj} \, \upsilon(p_{btj} - p_r)}{s_{tj} - x_{tj}} \right)}  + \\ & \dfrac{u_{ftjr}  \, S(-u_{ftjr})}{\left( \dfrac{c_{p2tj} \, \upsilon(p_s - p_{btj})}{s_{tj} - x_{tj}} + \dfrac{c_{n1tj} \upsilon(p_{atj} - p_r)}{x_{tj}} \right)}.
\end{array}
\label{eqn: utj}
\end{equation}

\section{Stability analysis}

\label{sec: stability analysis}

Per Theorem \ref{thm: virt stab}, this analysis can be carried out only for the general $j$-th manipulator segment. The stability of the entire manipulator is mathematically equivalent to the virtual stability of every manipulator module.

For purposes of analysis, a set of frames important for stability analysis is:
\begin{equation}
\quad \bf{S}_j = \left\lbrace \bf{B}_{0j}, \bf{B}_{1j}, \bf{B}_{3j}, \bf{B}_{4j}, \bf{B}_{5j}, \bf{P}_{2j}, \bf{P}_{3j} \right\rbrace.
\label{eqn: set of frames S}
\end{equation}

\newpage

\begin{thm}
Let the non-negative accompanying function for the $j$-th manipulator segment in the most general case considered here be chosen as the sum:
\begin{equation}
	{\nu}_{j} = \displaystyle\sum_{\bf A \in S_j} {\nu}_{\bf{A}}  + \nu_{pj} + \nu_{ptj}, 
	\label{eqn: nuj}
\end{equation}
where individual non-negative accompanying function ${\nu}_{\bf{A}}$ is chosen as:
\begin{equation}
	{\nu}_{\bf{A}} = \dfrac{1}{2} \, \left( {^{\bf A}{\boldsymbol V}_r} - {^{\bf A}{\boldsymbol V}}  \right)^T \, \boldsymbol{\rm M_{\rm A}} \, \left( {^{\bf A}{\boldsymbol V}_r} - {^{\bf A}{\boldsymbol V}}  \right),
	\label{eqn: accomp function}
\end{equation}
and let the non-negative accompanying functions for linear hydraulic actuators in the revolute segment and prismatic segment  be chosen as:
\begin{eqnarray}
	\nu_{pj} = \left( f_{pjr} - f_{pj} \right)^2/(2 \, \beta \, k_{xj}),
	\label{eqn: nupj}
\end{eqnarray}
and
\begin{eqnarray}
	\nu_{ptj} = \left( f_{ptjr} - f_{ptj} \right)^2/(2 \, \beta \, k_{xtj}),
	\label{eqn: nuptj}
\end{eqnarray}
respectively. By forming  control actions for linear hydraulic actuators in the revolute and prismatic segments using Eqs. (\ref{eqn: ufjr})–(\ref{eqn: utj}), a derivative of the non-negative accompanying function assigned to the $j$-th manipulator structure is:
\begin{equation}
	\begin{array}{ll}
		\dot{\nu}_{j} = 	& \dot{\nu}_{B_{5j}} + \dot{\nu}_{P_{2j}}  + \dot{\nu}_{P_{3j}}  + \dot{\nu}_{B_{1j}}  +  \dot{\nu}_{B_{0j}} + \dot{\nu}_{B_{4j}} + \dot{\nu}_{B_{3j}} + \dot{\nu}_{pj} + \dot{\nu}_{ptj} \leqslant \\
		& - \displaystyle\sum_{\bf A \in S_j} \left( {^{\bf A}{\boldsymbol V}_r} - {^{\bf A}{\boldsymbol V}}  \right)^T \, \, \mathbf{K}_{\bf A} \, \left( {^{\bf A}{\boldsymbol V}_r} - {^{\bf A}{\boldsymbol V}}  \right) +  p_{\bf B_{cj}} - p_{\bf E_{2j}} - \\
		& \dfrac{k_{fj}}{k_{xj}} \, \left( f_{pjr} - f_{pj} \right)^2 - \dfrac{k_{ftj}}{k_{xtj}} \, \left( f_{ptjr} - f_{ptj} \right)^2.
	\end{array}
\end{equation}
It implies the virtual stability of the $j$-th manipulator structure per Definition \ref{def: virt stab}, and consequently of the whole manipulator, per Theorem \ref{thm: virt stab}, implies:
\begin{equation}
	\begin{array}{ll}
		\dot{q}_{jd} - \dot{q}_j \rightarrow 0, &
		{q}_{jd} - {q}_j \rightarrow 0, \\
		\dot{x}_{jd} - \dot{x}_{j} \rightarrow 0, &
		{x}_{jd} - {x}_{j} \rightarrow 0, \\
		\dot{x}_{tjd} - \dot{x}_{tj} \rightarrow 0, &
		{x}_{tjd} - {x}_{tj} \rightarrow 0. \\
	\end{array}
\end{equation}
\end{thm}

The proof is given in Appendix C.

\section{Systematic formulation of control law}
\label{sec: contol algorithm}

\begin{figure}[h!]
\centering
\includegraphics[width=0.45\textwidth]{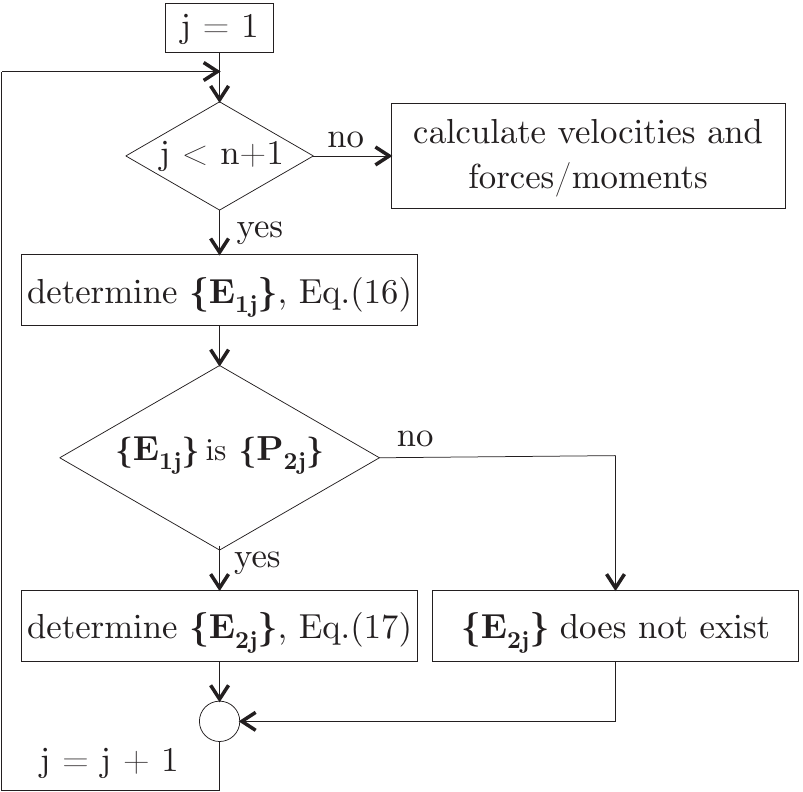}
\caption{Algorithm for determining $\left\lbrace \bf E_{1j} \right\rbrace$ and/or $\left\lbrace \bf E_{2j} \right\rbrace$ frames.}
\label{fig: algorithm 1}
\end{figure}

The discussed control law formulation can be shown algorithmically.

At the beginning of the systematic control law formulation, a step must be performed that will not be needed again. It is to determine which frames coincide with general frames $\bf \left\lbrace E_{1j} \right\rbrace$ and $\bf \left\lbrace E_{2j} \right\rbrace$ and whether a general frame  $\bf \left\lbrace E_{2j} \right\rbrace$ exists in a particular manipulator structure. 

By proceeding from the manipulator base to the manipulator end-effector, separate manipulator structures, as in Fig. \ref{fig: closed chain}, must be identified, and for each manipulator structure, frames $\bf \left\lbrace E_{1j} \right\rbrace$ and $\bf \left\lbrace E_{2j} \right\rbrace$ must be established, using the dense procedure given by Eqs. (\ref{eqn: frame E1j}) and (\ref{eqn: frame E2j}), Fig. \ref{fig: algorithm 1}. The latter does not exist if there is no prismatic segment in the considered manipulator structure.

Current values of linear/angular velocities are calculated first, going from the manipulator base to the end-effector. 
For each manipulator structure, as in Fig. \ref{fig: closed chain}, calculations are carried out from the driven to the driving point of a revolute segment, using Eqs. (\ref{eqn: BV})–(\ref{eqn: E1jV}). If a prismatic segment exists, calculations are continued using Eqs. (\ref{eqn: P2jV})–(\ref{eqn: P3jV}). The procedure is repeated for every manipulator structure in the series.

Current values of relevant forces are calculated from the end-effector, i.e. from frame $\left\lbrace \bf E_{2j} \right\rbrace$ of every $j$-th manipulator structure if the prismatic segment exists in the structure. If that is not the case, calculations start from frame $\left\lbrace \bf E_{1j} \right\rbrace$ in the $j$-th manipulator structure. Calculations are carried out to the manipulator structure's driven point, and this procedure is repeated for all manipulator structures up to the first, starting from $j = n$. If  frame $\left\lbrace \bf E_{2j} \right\rbrace$ exists, current values for forces in the prismatic segment  are calculated using Eqs. (\ref{eqn: O2jF})–(\ref{eqn: P2jF}). Frame $\left\lbrace \bf E_{1j} \right\rbrace$ will exist in  any case, regardless of whether frame $\left\lbrace \bf E_{2j} \right\rbrace$ exists. All the necessary values of forces in a revolute segment are calculated using Eqs. (\ref{eqn: piston force}) and  (\ref{eqn: driven cc force}). The procedure is repeated for all the subsystems. By performing these calculations, inertial and gravity terms are accounted for, but the reminder should be set again to bring the attention to the inclusion of the friction force whose impact can not be neglected.

\begin{figure}[h!]
\centering
\includegraphics[width=.775\textwidth]{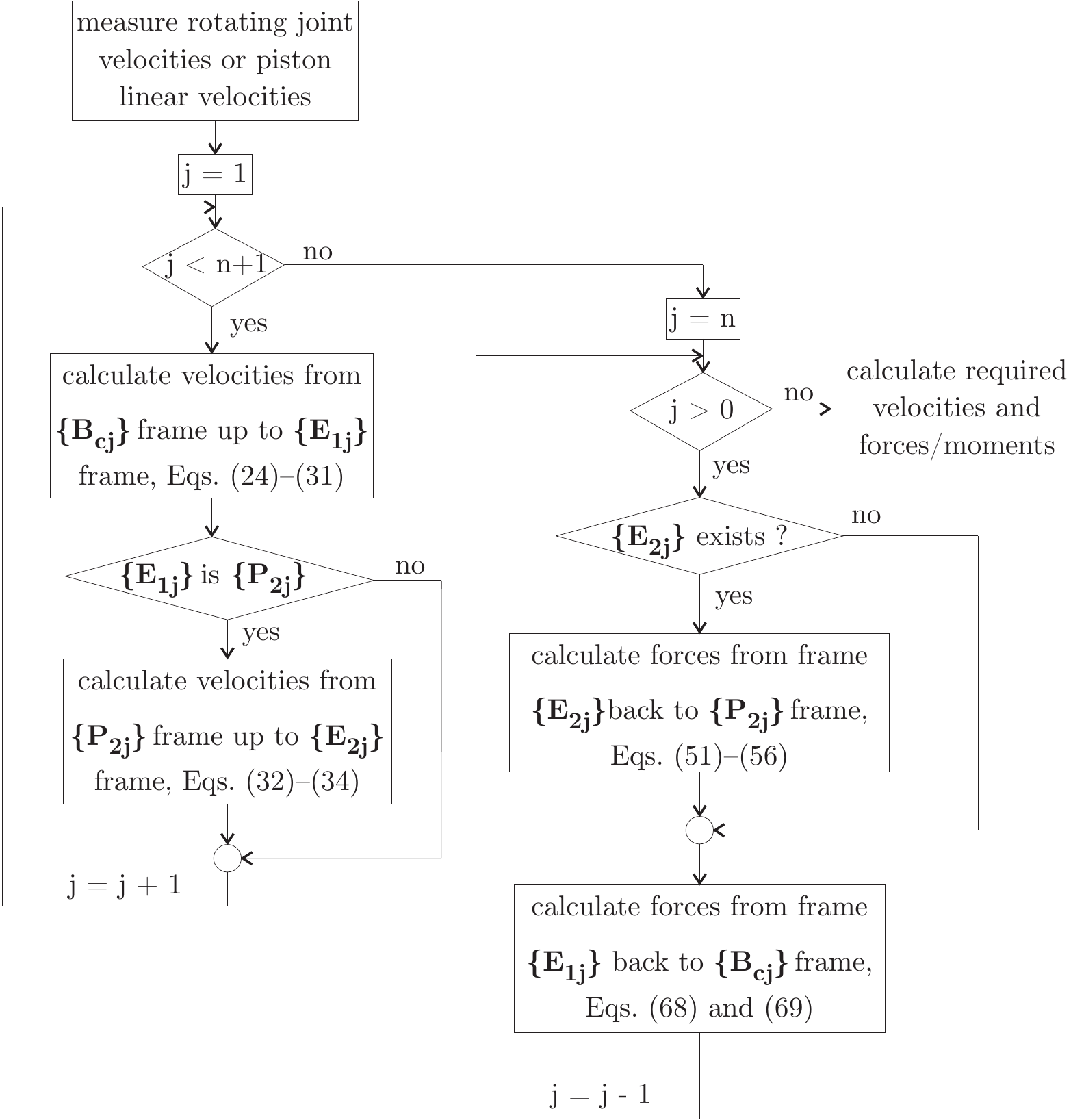}
\caption{Algorithm for calculating linear/angular velocities and  forces/moments.}
\label{fig: algorithm 2}
\end{figure}

\newpage

After solving the inverse kinematics problem, the required linear/angular velocities are calculated using relations Eqs. (\ref{eqn: BVr})–(\ref{eqn: E1jVr}) and eventually Eqs. (\ref{eqn: P2jVr})–(\ref{eqn: P3jVr}) in the case of prismatic segment existence. Equations (\ref{eqn: xjr})–(\ref{eqn: xtjr}) for required joint velocities are also needed in the process. The calculations of forces/moments starts with Eqs. (\ref{eqn: P3jFr})–(\ref{eqn: frame P2jFr}) if the prismatic segment exists, and it continues with expressions Eqs. (\ref{eqn: required piston force}) and (\ref{eqn: required cc force}). These are the only two expressions that require evaluation when a prismatic segment does not exist. Finally, control voltages are formed using Eqs. (\ref{eqn: ufjr}) and (\ref{eqn: uj2}) and/or Eqs. (\ref{eqn: uftrjj}) and (\ref{eqn: utj}).

\begin{figure}[h!]
\centering
\includegraphics[width=.775\textwidth]{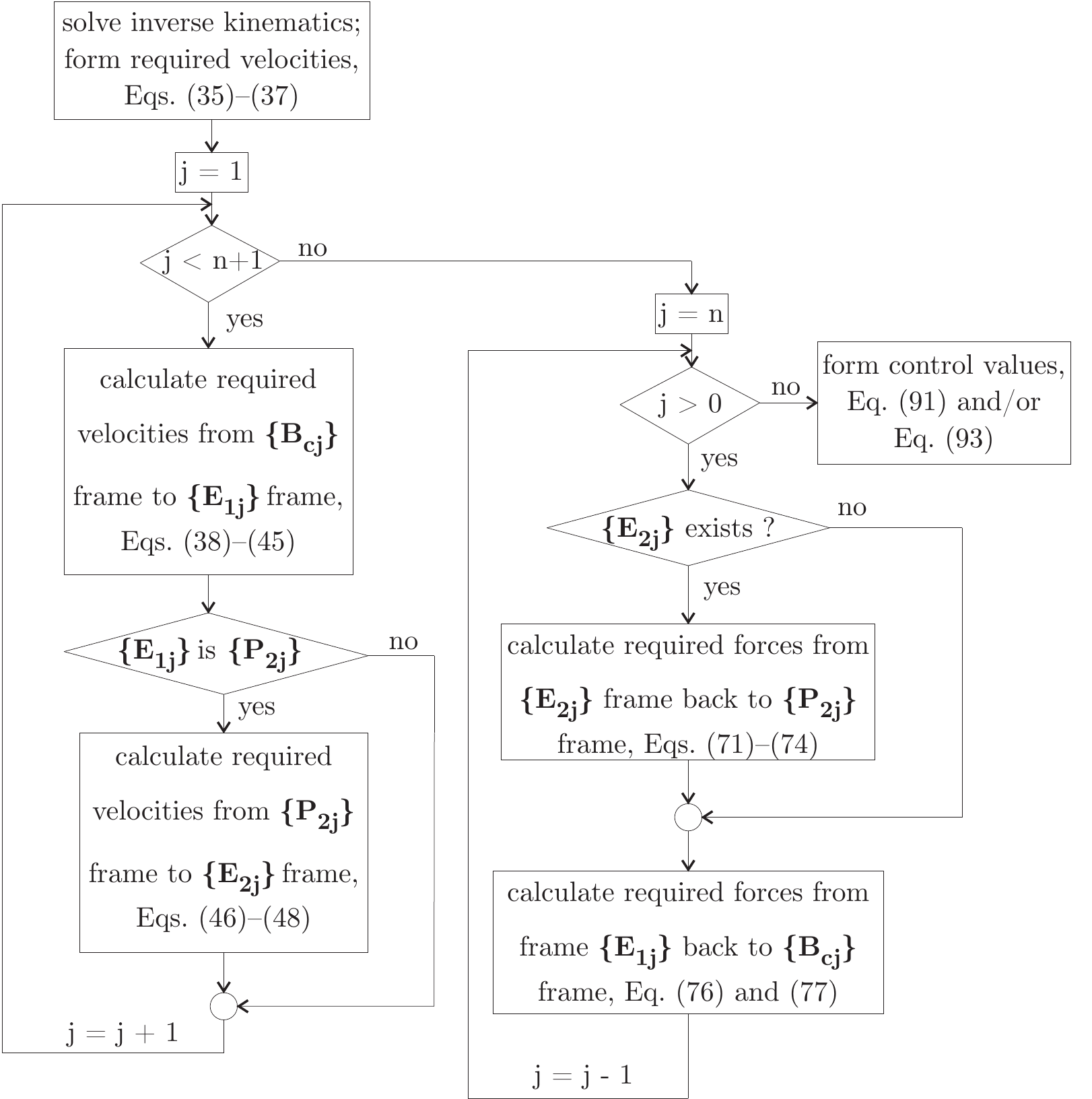}
\caption{Algorithm for calculating the required linear/angular velocities and forces/moments.}
\label{fig: algorithm 3}
\end{figure}

\section{Simulation results}

\label{sec: simulation results}

\begin{figure}[h!]
\centering
\includegraphics[width=.8\textwidth]{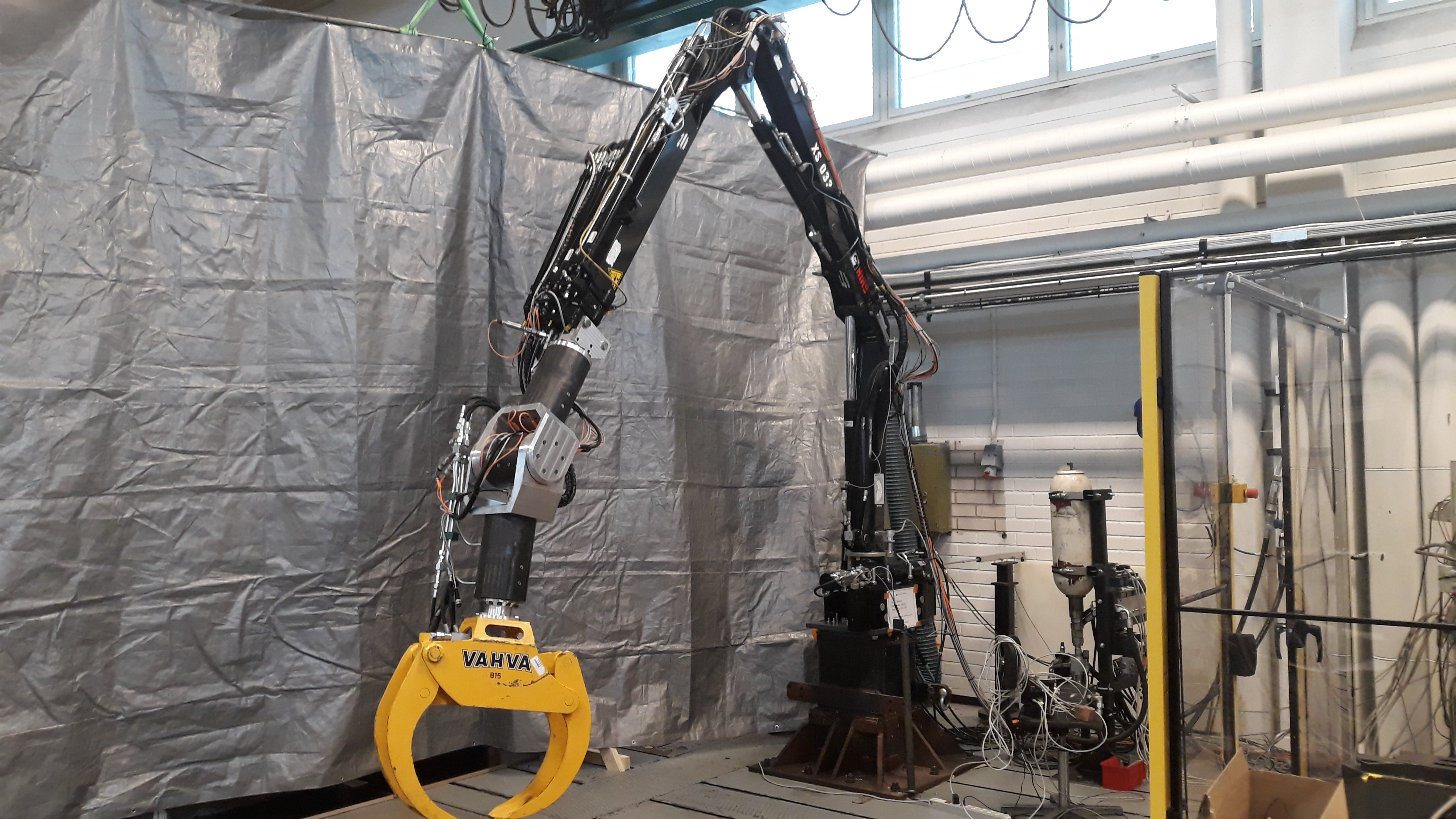}
\caption{Laboratory stand that inspired the simulation.} \label{fig: HIAB lab}
\end{figure}

The proposed forces modelling scheme has been verified in the Simscape {Multibody\texttrademark} simulation environment. A hydraulic manipulator from the subset of the manipulators proposed in this paper was the focus. The simulation was created using CAD models of the laboratory installation from Fig. \ref{fig: HIAB lab}.

A more detailed description of this particular experimental setup can be found in papers \cite{mustalahti2017nonlinear}, and \cite{mustalahti2019nonlinear}. For the simulation results that follow, it is relevant to emphasise that the servo valves are connected as:
\begin{itemize}
\item[1)] Bosch Rexroth NG6 size servo solenoid valve (40 l/min at $\Delta p = 35 ~\rm bar$ per notch) for the Lift cylinder,
\item[2)] Bosch Rexroth NG10 size servo solenoid valve (100 l/min at $\Delta p = 35 ~\rm bar$ per notch) for the Tilt cylinder.
\end{itemize}

\newpage

\begin{figure}[h!]
\centering
\includegraphics[width=.8\textwidth]{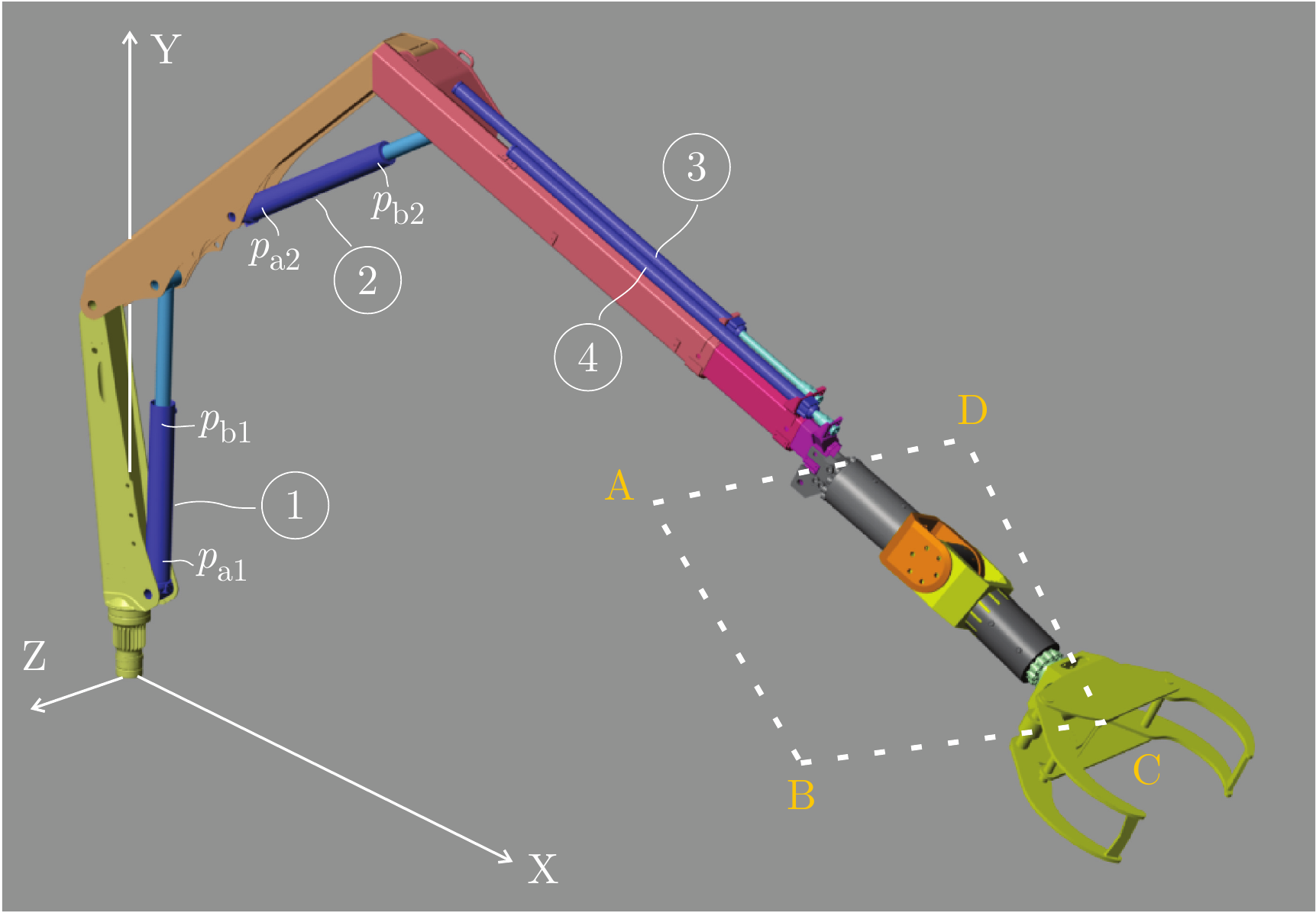}
\caption{Simulation explorer showing manipulator model used to validate the method.} \label{fig: HIABSnip}
\end{figure}

Fig. \ref{fig: HIABSnip} shows the simulation environment explorer at one point in time, in one of many poses during the simulation used to verify forces modelling method. The same figure also contains sketched the desired A-B-C-D path used later in the control system simulation and also contains labels for relevant pressures in the simulation, together with axis labels for the world frame. Pressures in the Lift cylinder are labelled as $p_{a1}$ and $p_{b1}$, while the pressures in the Tilt cylinder are labelled as $p_{a2}$ and $p_{b2}$.

\begin{figure}[h!]
\centering
\includegraphics[width=.8\textwidth]{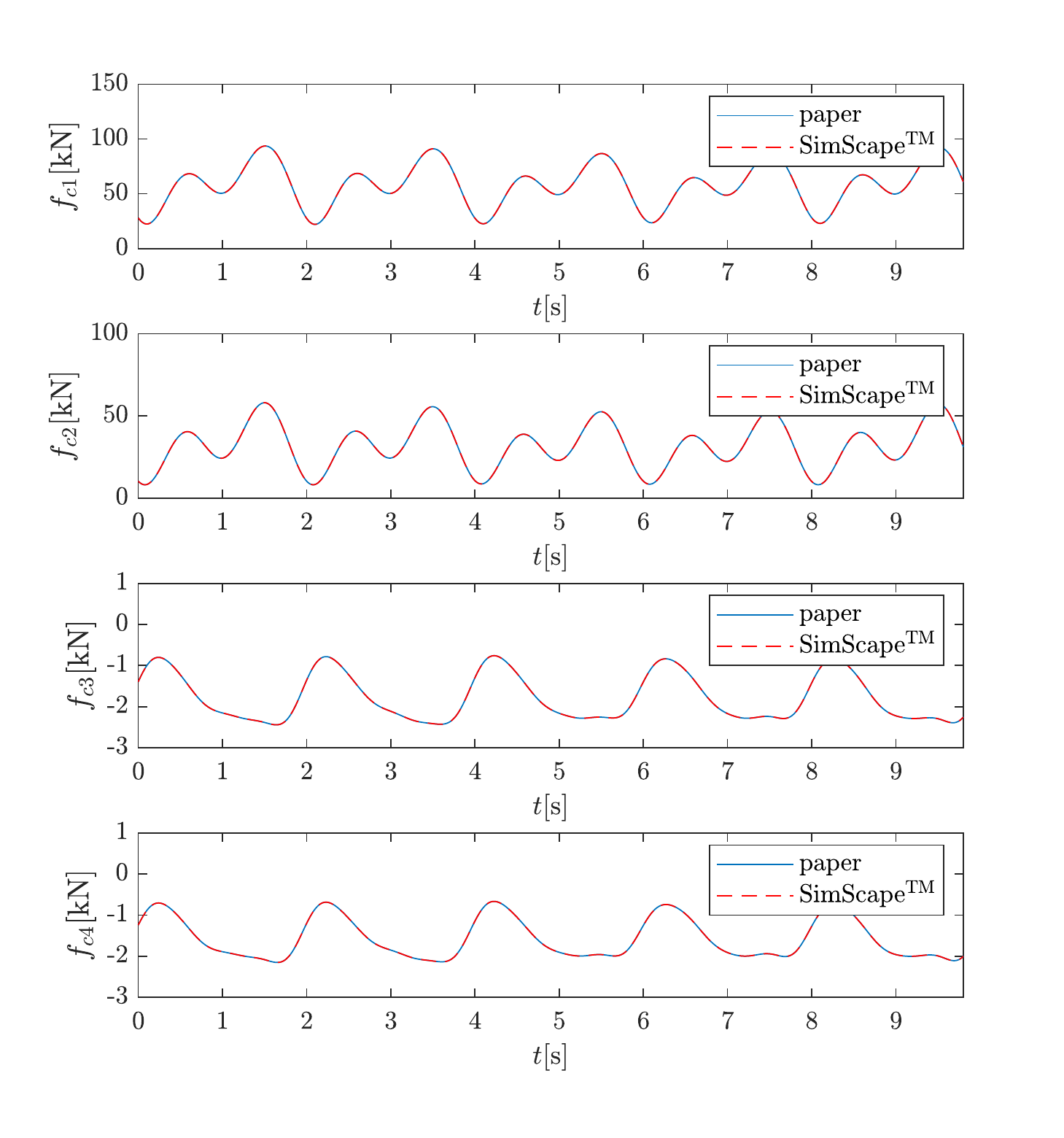}
\caption{Comparison of piston forces calculated using Eq. (\ref{eqn: piston force}) and SimScape\texttrademark.} \label{fig: sim forces}
\end{figure}

In order to verify the proposed inverse dynamics method, direct actuation in the simulation was performed using reasonably chosen but still generally random and uncorrelated changes for generalised manipulator coordinates. The type of actuation (hydraulic/electric) is not relevant at this point. The proposed theoretical approach was not related in any way to the particular test stand, and thus the results from Fig. \ref{fig: sim forces} validate the derived results numerically. 

Since there is no information about equations used in the simulation environment, and since this internal numerical approach is affected by many factors, such as the choice of the solver, a relative error of up to 0.2\% for reasonably short integration time, especially considering the magnitudes of the forces, qualifies the proposed inverse dynamics approach for further use.

To compare the execution times in Simulink, the run/sim time ratio from the Solver Profiler is used. Employing the variable-step ode45 solver with the relative tolerance setting set to $10^{-7}$, the run/sim time ratio in the case of the {SimScape\texttrademark} model is about 0.14$\rm s$. For the presented model, the same run/sim time ratio is 0.05$\rm s$ using the ode4 fixed-step solver with the step size of 0.001$\rm s$.

Physical quantities such as pressures, volumetric flows, and valve voltages were irrelevant when actuator forces were calculated since the method does not account for the type of actuators. When applying the proposed control law in the particular case of hydraulic actuators, it is required to calculate appropriate pressures and, consequently, valve control voltages as functions of the required forces.

For the considered manipulator, one typical planar task of reaching some starting point (labelled here as A), and after that performing two A-B-C-D loops with a fast transition (two seconds from one trajectory point to the other) along the quintic rest-to-rest path, as sketched in Fig. \ref{fig: HIABSnip} can be investigated in order to obtain values for pressures and control voltages and perform the preliminary assessment of the control algorithm.

\begin{figure}[h!]
\centering
\includegraphics[width=.85\textwidth]{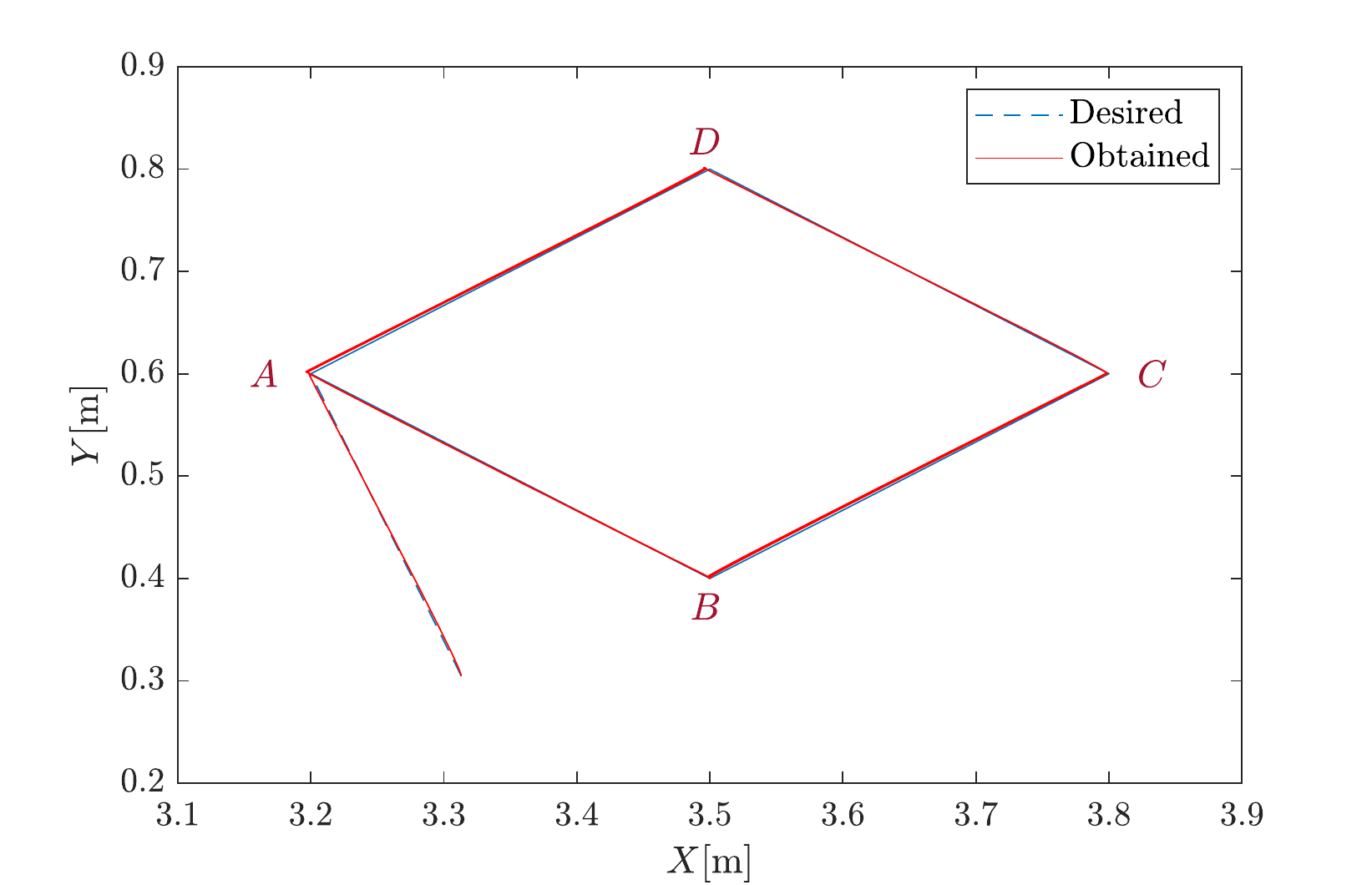}
\caption{Desired and obtained tool center point path.}
\label{fig: path following}
\end{figure} 

\begin{figure}[h!]
\centering
\includegraphics[width=.85\textwidth]{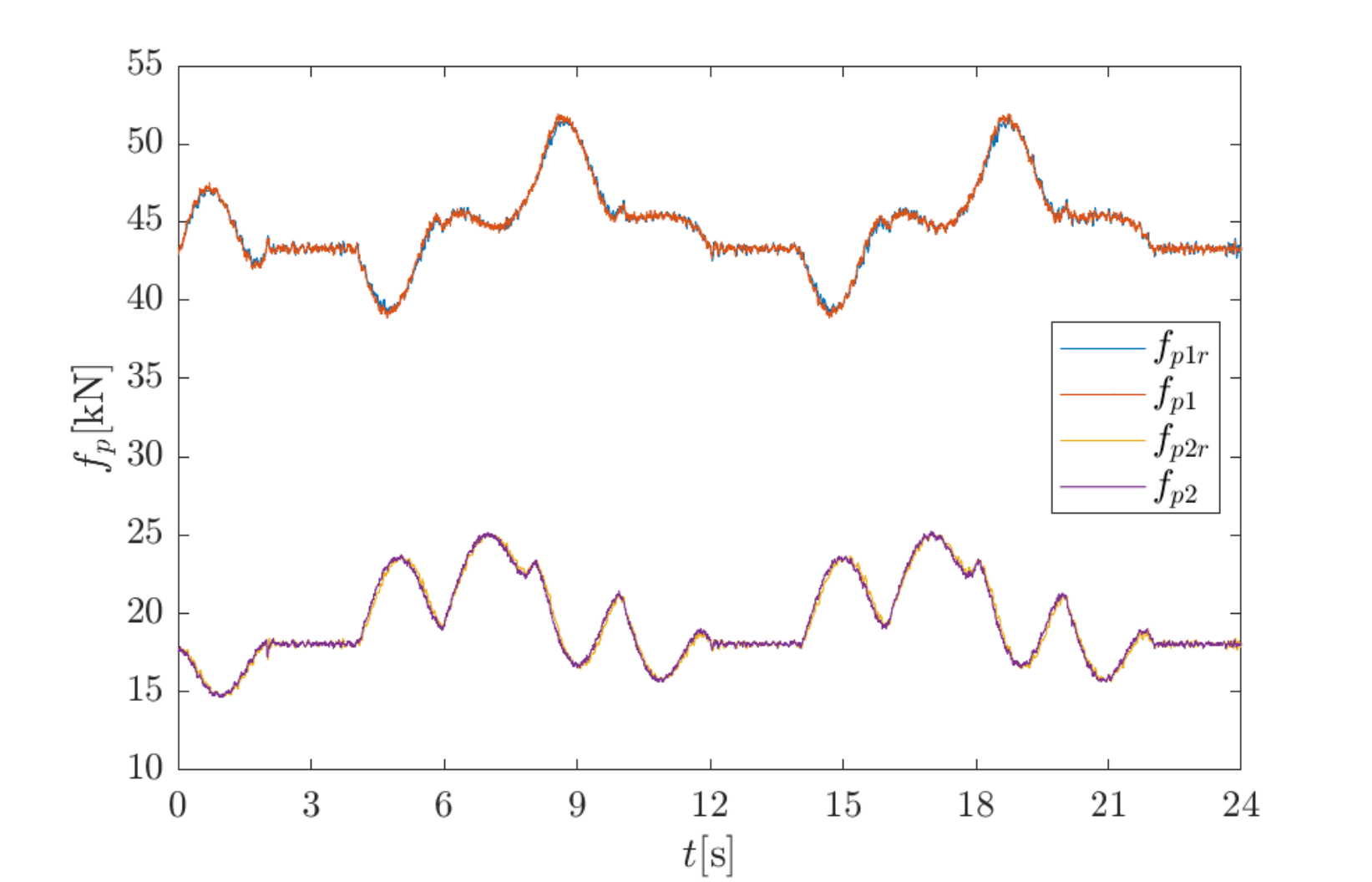}
\caption{Required and obtained piston forces.}
\label{fig: required forces}
\end{figure}

The desired and the obtained trajectory in the simulation are shown in  Fig. \ref{fig: path following}. Figure \ref{fig: required forces} shows required and achieved piston forces in the Lift ($f_{p1r}$, $f_{p1}$) and Tilt ($f_{p2r}$, $f_{p2}$) cylinders.

\newpage

\begin{figure}[h!]
\centering
\includegraphics[width=.85\textwidth]{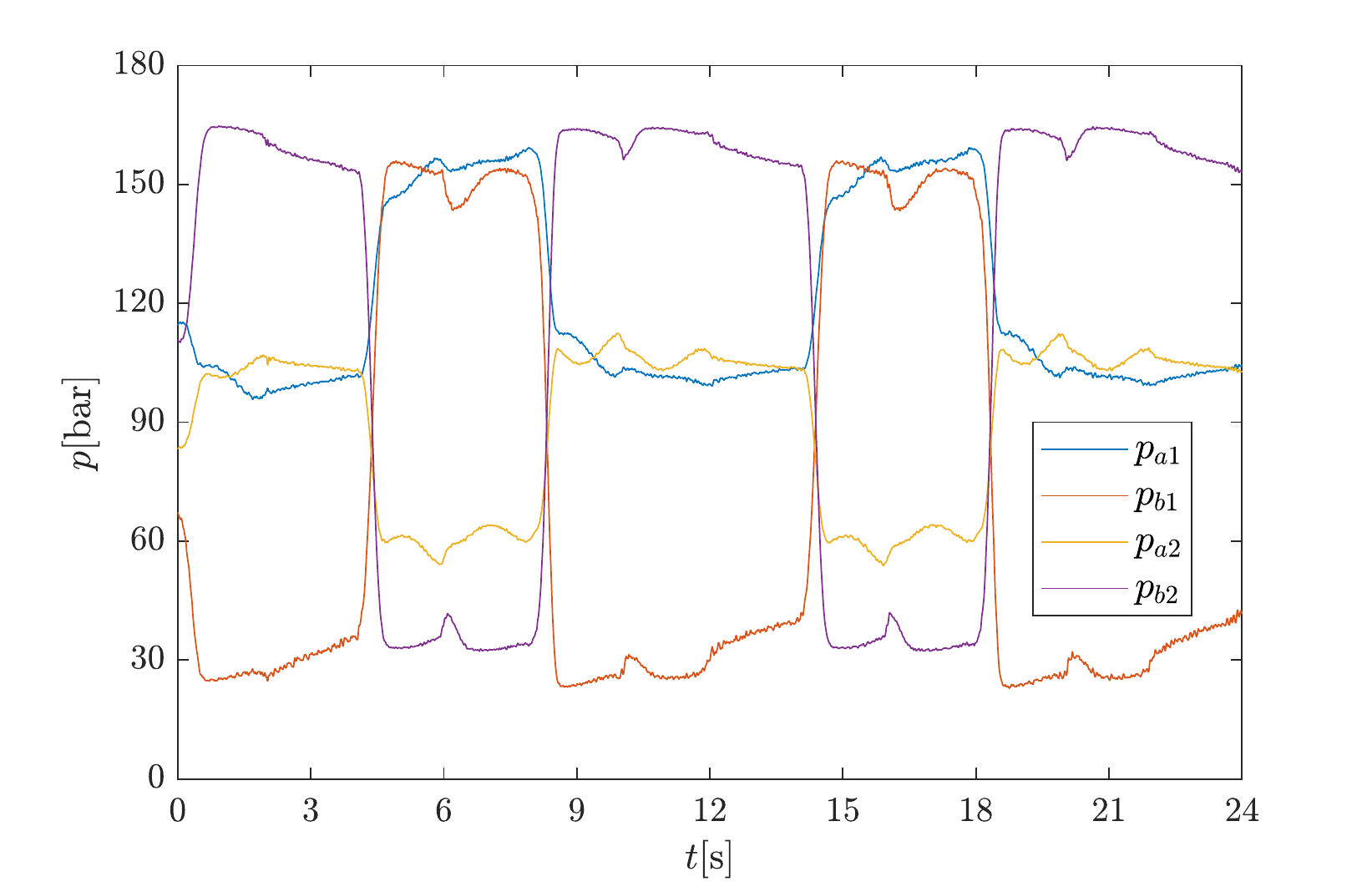}
\caption{Pressures in the Lift and Tilt cylinders.}
\label{fig: pressures exp}
\end{figure}

Figure \ref{fig: pressures exp} shows pressure changes in the Lift and Tilt cylinders during the simulated experiment. These are labeled in Fig. \ref{fig: HIABSnip}.

\begin{figure}[h!]
\centering
\includegraphics[width=.85\textwidth]{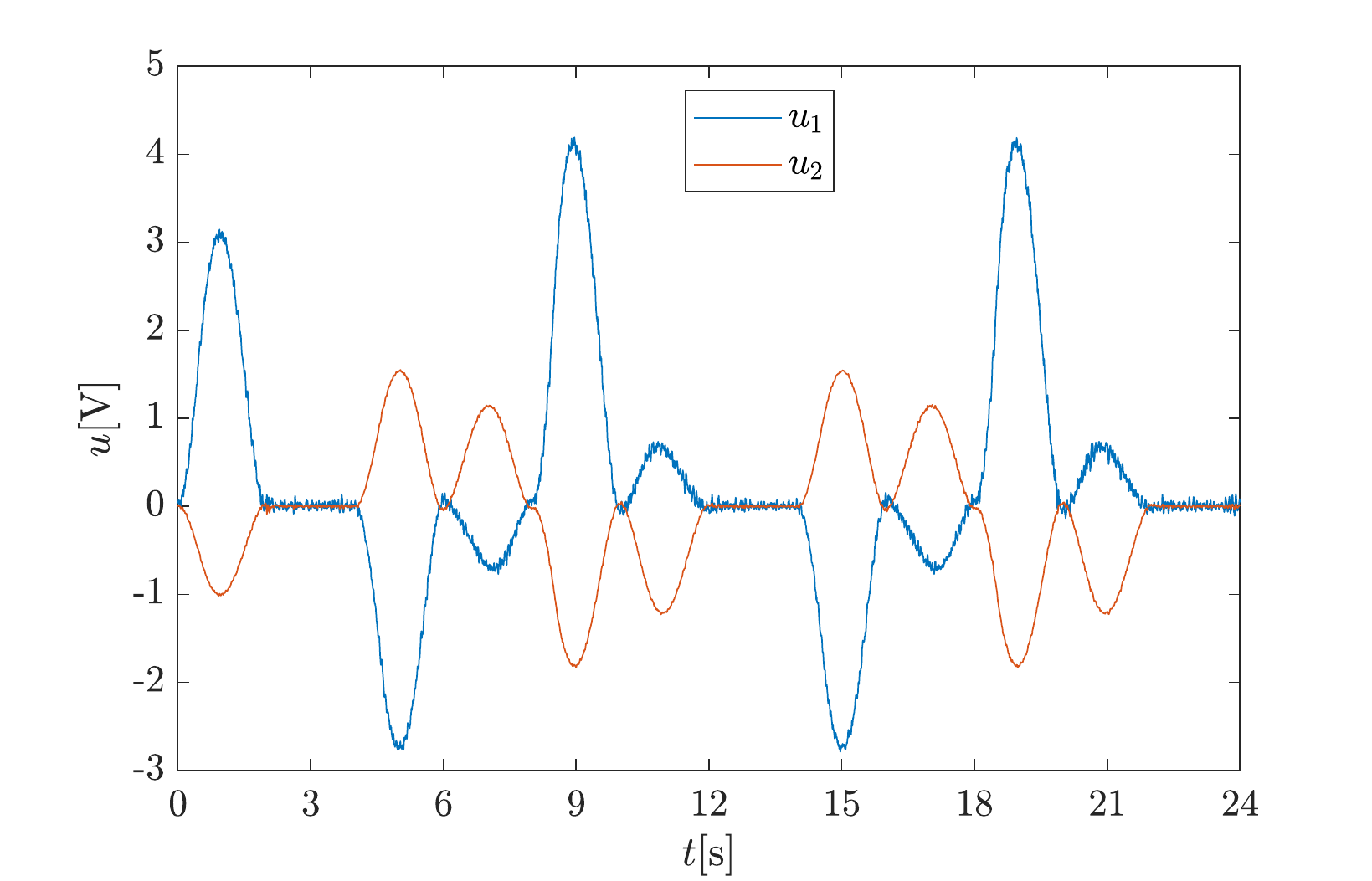}
\caption{Control voltages for directional valves.}
\label{fig: control voltages}
\end{figure}

Control voltages have a limited range of $\pm 10\, \rm V$, and the corresponding time changes for both control valves in the simulated experiment are shown in Figure \ref{fig: control voltages}.
The supply pressure is assumed to be constant at $185 \, \rm bar$ with the return pressure also being constant at $10 \, \rm bar$. 

All the results in Figs. \ref{fig: path following} $-$ \ref{fig: control voltages}  present the simulation results obtained using the control forming approach, which relies on the presented novel modelling scheme. The simulation has addressed some important practical aspects, such as the noise presence in all the measurements, which, in turn, required the introduction of appropriate signal filtering, which is all to be tested, verified and discussed in more detail in future experimental work. Also, the implemented friction model is assumed to be precisely known. The friction forces are not compensated using the parameter adaptation, which will probably be necessary during the experimental part to achieve better performance and will be addressed in more detail.

\section{Discussion}

\label{sec: discussion}

The final analytic expression presented here, used to calculate actuator forces in revolute segments and given by Eq. (\ref{eqn: piston force}) can also be obtained following the algorithm in \cite{koivumaki2018addressing} and performing a simplification through  careful symbolic manipulation. Anyhow, this has remained unnoticed so far.

Using directly Eq.  (\ref{eqn: piston force}), at least five different trigonometric operations, 14 multiplications, and one fewer division have to be performed to obtain the same linear actuator force value in one revolute segment. The number of necessary operations that are computationally expensive is significantly reduced, while the final SoA expressions remain intact. Since previously used, additional auxiliary quantities, are no longer needed, the whole analysis is now more intuitive and straightforward.

Apart from that, an independent comparison with results obtained using Simscape {Multibody\texttrademark} for one particular manipulator configuration was performed, providing confirmation of the correctness of the proposed expressions.

In addition, the proposed virtual decomposition approach decreases the number of subsystems needed for the analysis. Mass objects between the driving VCP of a revolute segment and the driven VCP of the following revolute/prismatic segment do not have to be considered separately anymore. This reduces the total number of equations for at least $4 \, n \, N$, and possibly more, depending on the additional algorithm complexities. It also lowers the number of inputs to the algorithm, i.e. fewer inertia tensors, lengths, masses, and similar must be known in the process.

Virtual stability of the generic manipulator structure from Fig. \ref{fig: closed chain} per Definition \ref{def: virt stab} is ensured by proper formulation of control values. This, in turn, guarantees the stability of the entire robot per Theorem \ref{thm: virt stab} and as a  consequence, physical quantities converge to their required values. 

This different decomposition leaves the virtual stability of the manipulator structure intact, and because of this, a subsystem can be controlled independently from the rest of the system, using one of the main ideas of VDC, modularity.

This reformulation of the linear hydraulic actuator forces model in types of manipulators relevant in practice can be used to rewrite all the existing VDC results and formulate new ones with less effort. The reformulation is expected to yield many results in this SoA NMB, stability-guaranteeing, modular control technique, proven to be  implementable in real time.

\section{Conclusions}

\label{sec: conclusions}

This paper provides reformulated general dynamics equations in the N–E framework, using the 6D vector formulation for a relevant parallel–serial hydraulic manipulator configuration often encountered in practice. 

As with previous N-E models in the VDC framework, the proposed model does not use usual approximations, which puts it in front of traditional models based on the Lagrange formulation.

The reformulation leads to a more straightforward analytic solution for calculation of a linear hydraulic actuator force when this actuator exists as a part of a 3-bar revolute segment with a passive joint in a hydraulic manipulator.

In contrast to the reported modelling schemes in the N–E framework, the method presented here is more intuitive and efficient since fewer equations are formed. Actuator forces can now be calculated without surplus factors required by previous approaches, using fewer calculation operations in the process.

The proposed scheme for actuator force calculation has been validated both in simulation and analytically, using numerical results from Simscape {Multibody\texttrademark} and the current SoA analytic expressions as a reference.

The analysis has considerable significance when VDC is used to control the hydraulic manipulators essential in practical applications.    

A systematic subsystem-based process to formulate control law has been presented in the VDC framework using the new dynamics model, which also encompasses pressure dynamics while rigorously guaranteeing Lyapunov stability of the whole manipulator.

It is also expected that this reformulated systematic approach for VDC implementation, based on the proposed model, becomes a de-facto standard in the VDC community since it leads to experimentally verified, state-of-the-art (SoA) analytic expressions in a much more straightforward way.

\section*{Declaration of Competing Interest}

The authors declare that they have no known competing financial interests or personal relationships that could have appeared to influence the work reported in this paper.

\section*{Acknowledgements}

This project STREAM has received funding from the Shift2Rail joint Undertaking (JU) under grant agreement No. 101015418. The JU receives support from the European Union's Horizon 2020 research and innovation programme and the Shift2Rail JU members other than the Union. The content of this paper does not reflect the official opinion of the Shift2Rail Joint Undertaking (S2R JU). Responsibility for the information and views expressed in the paper lies entirely with the authors.

\appendix

\section{Proof of Theorem \ref{thm: piston force}}

\begin{pf}

Let the motion of link $L_{j1}$ be described with Eq. (\ref{eqn: B1jF}). To derive the final result, some special properties of the transformation matrices are used. The force/moment transformation matrix, which relates frames $\left\lbrace \bf B_{1j} \right\rbrace$ and $\left\lbrace \bf P_{1j} \right\rbrace$ is:
\begin{equation}
	{^{\bf B_{1j}}\mathbf{U_{P_{1j}}}} = \begin{pmatrix}
		\cos q_{j2} & \sin q_{j2} & 0 & 0 & 0 & 0 \\ -\sin q_{j2} & \cos q_{j2} & 0 & 0 & 0 & 0 \\ 0 & 0 & 1 & 0 & 0 & 0 \\ 0 & 0 & 0 & \cos q_{j2} & \sin q_{j2} & 0 \\ 0 & 0 & -L_{j1} & -\sin q_{j2} & \cos q_{j2} & 0 \\- L_{j1} \, \sin q_{j2} & L_{j1} \, \cos q_{j2} & 0 & 0 & 0 & 1
	\end{pmatrix}.
	\label{eqn: B1jUP1j}
\end{equation}

The forces/moments with neglected friction in the $q_{j2}$-joint  per Eq. (\ref{eqn: fric qj2}) are:
\begin{equation}
	{^{\bf P_{1j}}\boldsymbol F} = \begin{pmatrix}
		{^{P_{1j}}f_{\rm x}} & {^{P_{1j}}f_{\rm y}} & 0 & 0 & 0 & 0
	\end{pmatrix}^T.
	\label{eqn: P1jF}
\end{equation}

The motion equation, Eq. (\ref{eqn: B1jF}), combined with Eq. (\ref{eqn: fric qj}) and  Eq. (\ref{eqn: P1jF}) gives:
\begin{equation}
	\mathbf{z}_{\tau}^T \left( {^{\bf B_{1j}}\boldsymbol F^*} + {^{\bf B_{1j}}\mathbf{U_{E_{1j}}}}	\, {^{\bf E_{1j}}\boldsymbol F} \right) = -	{^{P_{1j}}f_{\rm x}} \, L_{j1} \, \sin q_{j2} + {^{P_{1j}}f_{\rm y}} \, L_{j1} \, \cos q_{j2},
	\label{eqn: mcj}
\end{equation}
and Eq. (\ref{eqn: mcj}) presents the starting point from which the  solution is obtained.

The piston and rod subsystem moves only in the direction of the local frame $x$-axis, so its motion is constrained. It also has an angular velocity about the local frame $z$-axis different from zero. The force/moment vector at the $\left\lbrace \bf B_{4j} \right\rbrace$ frame is modelled as:
\begin{equation}
	{^{\bf{B}_{4j}}\boldsymbol F} = \begin{pmatrix}
		{^{B_{4j}}f_{\rm x}} & {^{B_{4j}}f_{\rm y}} & 0 & 0 & 0 & {^{B_{4j}}m_{\rm z}}
	\end{pmatrix}^T.
	\label{eqn: fcj}
\end{equation}

The force/moment transformation matrix from frame $\left\lbrace \bf P_{1j} \right\rbrace$ to frame $\left\lbrace \bf B_{4j} \right\rbrace$ has a very simple structure, because ${^{\bf B_{4j}}\mathbf{R_{P_{1j}}}} = \mathbf{I}_{3 \times 3}$:
\begin{equation}
	{^{\bf B_{4j}}\mathbf{U_{P_{1j}}}} = \begin{pmatrix}
		1 & 0 & 0 & 0 & 0 & 0 \\ 0 & 1 & 0 & 0 & 0 & 0 \\ 0 & 0 & 1 & 0 & 0 & 0 \\ 0 & 0 & 0 & 1 & 0 & 0 \\ 0 & 0 & -l_{cj} & 0 & 1 & 0 \\ 0 & l_{cj} & 0 & 0 & 0 & 1
	\end{pmatrix},
	\label{eqn: matrix B4jUP1j}
\end{equation}
where $l_{cj}$ is the length of the piston and rod from Fig. \ref{fig: closed chain}. Equation (\ref{eqn: B4jF}), combined with Eqs. (\ref{eqn: fcj}) and (\ref{eqn: matrix B4jUP1j}) yields a simpler form, which is very significant for further analysis. That is, separate EOMs from Eq. (\ref{eqn: B4jF}) become:
\begin{equation}
	{^{B_{4j}}f_{\rm x}} = \mathbf{x}_f^T \left( {\bf ^{B_{4j}}\boldsymbol{F}^*} \right) + {^{P_{1j}}f_{\rm x}},
	\label{eqn: b4j new 1}
\end{equation}
\begin{equation}
	{^{B_{4j}}f_{\rm y}} = \mathbf{y}_f^T \left( {\bf ^{B_{4j}}\boldsymbol{F}^*} \right) + {^{P_{1j}}f_{\rm y}},
	\label{eqn: b4j new 2}
\end{equation}
and
\begin{equation}
	{^{B_{4j}}m_{\rm z}} = \mathbf{z}_\tau^T \left( {\bf^{B_{4j}}\boldsymbol{F}^*} \right) + {^{P_{1j}}f_{\rm y}} \, l_{cj},
	\label{eqn: b4j new 3}
\end{equation}
where $\mathbf{y}_f = \begin{pmatrix}
	0 & 1 & 0 & 0 & 0 & 0
\end{pmatrix}^T$.

The force/moment transformation matrix from frame $\left\lbrace \bf B_{4j} \right\rbrace$ to frame $\left\lbrace \bf B_{3j} \right\rbrace$ also has a very simple structure. Again, this is because  ${^{\bf B_{3j}}\mathbf{R_{B_{4j}}}} = \mathbf{I}_{3 \times 3}$:
\begin{equation}
	{^{\bf B_{3j}}\mathbf{U_{B_{4j}}}} = \begin{pmatrix}
		1 & 0 & 0 & 0 & 0 & 0 \\ 0 & 1 & 0 & 0 & 0 & 0 \\ 0 & 0 & 1 & 0 & 0 & 0 \\ 0 & 0 & 0 & 1 & 0 & 0 \\ 0 & 0 & -x_j - x_{j0} + l_{cj} & 0 & 1 & 0 \\ 0 & x_j + x_{j0} - l_{cj} & 0 & 0 & 0 & 1
	\end{pmatrix}.
	\label{eqn: matrix B4jUB3j}
\end{equation}

The  forces/moments with neglected friction in the $q_{j1}$-joint per Eq. (\ref{eqn: fric qj1}) are:
\begin{equation}
	{^{\bf B_{3j}}\boldsymbol{F}} = \begin{pmatrix}
		{^{B_{3j}}f_{\rm x}} & {^{B_{3j}}f_{\rm y}} & 0 & 0 & 0 & 0
	\end{pmatrix}^T.
	\label{eqn: B3jF nf}
\end{equation}

Using Eqs. (\ref{eqn: matrix B4jUB3j}) and (\ref{eqn: B3jF nf}), separate EOMs from Eq. (\ref{eqn: B3jF}) are:
\begin{equation}
	{^{B_{3j}}f_{\rm x}} = \mathbf{x}_f^T \left( {\bf ^{B_{3j}}\boldsymbol{F}^*} \right) + {^{B_{4j}}f_{\rm x}},
\end{equation}
\begin{equation}
	{^{B_{3j}}f_{\rm y}} = \mathbf{y}_f^T  \left( {\bf ^{B_{3j}}\boldsymbol{F}^*} \right) + {^{B_{4j}}f_{\rm y}},
\end{equation}
and
\begin{equation}
	0 = \mathbf{z}_\tau^T \left( {\bf ^{B_{3j}}\boldsymbol{F}^*} \right) + {^{B_{4j}}m_{\rm z}} + {^{B_{4j}}f_{\rm y}} \, (x_j + x_{j0} - l_{cj}).
	\label{eqn: b3j new 3}
\end{equation}

Combining Eqs.  (\ref{eqn: b4j new 2}) and  (\ref{eqn: b4j new 3}) and Eq. (\ref{eqn: b3j new 3}), the solution for ${^{B_{4j}}f_{\rm y}}$ is:
\begin{equation}
	{^{B_{4j}}f_{\rm y}} = -\dfrac{\mathbf{z}_\tau^T  \left({^{\mathbf{B_{3j}}}\boldsymbol{F}^*} \right) + \mathbf{z}_\tau^T \left({^{\mathbf{B_{4j}}}\boldsymbol{F}^*} \right)}{x_j  + x_{j0}} + \dfrac{\mathbf{y}_f^T \left( {^{\mathbf{B_{4j}}}\boldsymbol{F}^*} \right) \, l_{cj}}{x_j  + x_{j0}}.
	\label{eqn: b4jy new}
\end{equation}

Further, combining equations Eq. (\ref{eqn: mcj}), Eqs. (\ref{eqn: b4j new 1}) and (\ref{eqn: b4j new 2}), and Eq. (\ref{eqn: b4jy new}), the final expression for the actuator force in the revolute segment can be obtained as:
\begin{equation}
	\begin{array}{ll}
		{^{B_{4j}}f_{\rm x}} = &  -  \dfrac{\mathbf{z}_{\tau}^T \left( {^{\bf B_{1j}}\boldsymbol{F}^*} + {^{\bf B_{1j}}\mathbf{U_{E_{1j}}}}	\, {^{\bf E_{1j}}\boldsymbol{F}} \right)}{L_{j1} \, \sin q_{j2}} + \mathbf{x}_f^T \left(  {^{\mathbf{B_{4j}}}\boldsymbol{F}^*} \right)  + \\ & \left({^{B_{4j}}f_{\rm y}} - \mathbf{y}_f^T \left(  {^{\bf B_{4j}}\boldsymbol F^*}  \right) \right)   \cot q_{j2}.
	\end{array}
\end{equation}
which is in the expanded form given with Eq. (\ref{eqn: piston force}), and this finishes the proof.
\end{pf}

\section{Proof of Theorem \ref{thm: driven cc force}}

\begin{pf}
Consider the revolute segment from Fig. \ref{fig: decomposed chain}a. Equations (\ref{eqn: B4jF}) and (\ref{eqn: B3jF})  can be combined into the following relation:
\begin{eqnarray}
	{^{\bf B_{3j}}\boldsymbol F} = {^{\bf B_{3j}}\boldsymbol F^*} + {^{\bf B_{3j}}\mathbf{U_{B_{4j}}}}	\, \left( {^{\bf B_{4j}}\boldsymbol F^*} + {^{\bf B_{4j}}\mathbf{U_{P_{1j}}}}	\, {^{\bf P_{1j}}\boldsymbol F} \right).
	\label{eqn: B3jF new 1}
\end{eqnarray}
Further, Eq. (\ref{eqn: B3jF}) and Eq. (\ref{eqn: B3jF new 1}) can be combined into:
\begin{equation}
	{{\bf^{B_{2j}}}\boldsymbol{F}} =  {^{\bf B_{2j}}\mathbf{U_{B_{3j}}}}  {^{\bf B_{3j}}\boldsymbol F^*} +{^{\bf B_{2j}}\mathbf{U_{B_{3j}}}}   {^{\bf B_{3j}}\mathbf{U_{B_{4j}}}} {^{\bf B_{4j}}\boldsymbol F^*} + {^{\bf B_{2j}}\mathbf{U_{B_{3j}}}}   {^{\bf B_{3j}}\mathbf{U_{B_{4j}}}} {^{\bf B_{4j}}\mathbf{U_{P_{1j}}}}	\, {^{\bf P_{1j}}\boldsymbol F}.
	\label{eqn: B2F new 1}
\end{equation}

On the other hand, combining Eqs. (\ref{eqn: B1jF}) and (\ref{eqn: B0jF}) gives:
\begin{equation}
	{^{\bf B_{0j}}\boldsymbol F} = {^{\bf B_{0j}}\boldsymbol F^*} +   {^{\bf B_{0j}}\mathbf{U_{B_{1j}}}}  {^{\bf B_{1j}}\boldsymbol F^*} + {^{\bf B_{0j}}\mathbf{U_{B_{1j}}}} {^{\bf B_{1j}}\mathbf{U_{E_{1j}}}}	\, {^{\bf E_{1j}}\boldsymbol F}	- {^{\bf B_{0j}}\mathbf{U_{B_{1j}}}} {^{\bf B_{1j}}\mathbf{U_{P_{1j}}}}	\, {^{\bf P_{1j}} \boldsymbol F}.
	\label{eqn: B1F new 1}
\end{equation}

Combining Eqs.  (\ref{eqn: B2F new 1}) and (\ref{eqn: B1F new 1}) per Eq. (\ref{eqn: BcjF}) gives a final expression for the force at the driven point, free of any internal forces:
\begin{equation}
	\begin{array}{ll}
		{^{\bf B_{cj}}\boldsymbol F} = & {^{\bf B_{0j}}\boldsymbol F^*} + {^{\bf B_{0j}}\mathbf{U_{B_{1j}}}}  {^{\bf B_{1j}}\boldsymbol F^*}  + {^{\bf B_{2j}}\mathbf{U_{B_{3j}}}}  {^{\bf B_{3j}}\boldsymbol F^*} + \\ & {^{\bf B_{2j}}\mathbf{U_{B_{3j}}}}   {^{\bf B_{3j}}\mathbf{U_{B_{4j}}}} {^{\bf B_{4j}}\boldsymbol F^*} + {^{\bf B_{0j}}\mathbf{U_{B_{1j}}}} {^{\bf B_{1j}}\mathbf{U_{E_{1j}}}}	\, {^{\bf E_{1j}}\boldsymbol F}.
	\end{array}
\end{equation}
\end{pf}

\section{Proof of virtual stability}

Stability analysis is carried out for the general case of a manipulator structure, including both revolute and prismatic segments. If the prismatic segment does not exist in the $j$-th manipulator structure, only specific terms drop from the analysis, and all the conclusions remain the same. 

\begin{pf}
Time derivatives of accompanying non-negative functions given by Eq. (\ref{eqn: accomp function}), where $\left\lbrace \bf{A} \right\rbrace \in \bf S_j$ and the set $\bf S_j$ is given with Eq. (\ref{eqn: set of frames S}), are the first to be found and combined. Using Eq. (\ref{eqn: tot force 2}) and Eq. (\ref{eqn: tot force 2 r}) along with kinematic relations Eqs. (\ref{eqn: BV})–(\ref{eqn: P3jV}) and dynamics relations given with Eqs. (\ref{eqn: O2jF})–(\ref{eqn: P2jF}) and Eqs. (\ref{eqn: B1jF})–(\ref{eqn: BcjF}), the following expressions are obtained for each subsystem-related non-negative accompanying function:

\begin{equation}
	\begin{array}{ll}
		\dot{\nu}_{B_{0j}} =  & \left( {^{\bf B_{0j}}{\boldsymbol V}_r} - {^{\bf B_{0j}}{\boldsymbol V}}  \right)^T \, \left( {^{\bf B_{0j}}{\boldsymbol F}_r^*} - {^{\bf B_{0j}}{\boldsymbol F}^*}  \right) = \\
		&  \left( {^{\bf B_{0j}}{\boldsymbol V}_r} - {^{\bf B_{0j}}{\boldsymbol V}}  \right)^T \, \left( {^{\bf B_{0j}}{\boldsymbol F}_r} - {^{\bf B_{0j}}{\boldsymbol F}}  \right) - \\ & \left( {^{\bf B_{0j}}{\boldsymbol V}_r} - {^{\bf B_{0j}}{\boldsymbol V}}  \right)^T \,  {^{\bf B_{0j}}\mathbf{U_{B_{1j}}}} \, \left( {^{\bf B_{1j}}{\boldsymbol F}_r} - {^{\bf B_{1j}}{\boldsymbol F}}  \right)  - \\
		&  \left( {^{\bf B_{0j}}{\boldsymbol V}_r} - {^{\bf B_{0j}}{\boldsymbol V}}  \right)^T \, \, \mathbf{K}_{\bf B_{0j}} \, \left( {^{\bf B_{0j}}{\boldsymbol V}_r} - {^{\bf B_{0j}}{\boldsymbol V}}  \right) = \\
		&  \left( {^{\bf B_{0j}}{\boldsymbol V}_r} - {^{\bf B_{0j}}{\boldsymbol V}}  \right)^T \, \left( {^{\bf B_{0j}}{\boldsymbol F}_r} - {^{\bf B_{0j}}{\boldsymbol F}}  \right) - \\&   \left( {^{\mathbf{B}_{\bf 1j}}\mathbf{U}_\mathbf{B_{0j}}^T}  \, \left( {^{\bf B_{1j}}{\boldsymbol V}_r} - {^{\bf B_{1j}}{\boldsymbol V}}  \right) \right)^T {^{\bf B_{0j}}\mathbf{U_{B_{1j}}}} \left( {^{\bf B_{1j}}{\boldsymbol F}_r} - {^{\bf B_{1j}}{\boldsymbol F}}  \right)  - \\
		&  \left( {^{\bf B_{0j}}{\boldsymbol V}_r} - {^{\bf B_{0j}}{\boldsymbol V}}  \right)^T \, \mathbf{K}_{\bf B_{0j}} \, \left( {^{\bf B_{0j}}{\boldsymbol V}_r} - {^{\bf B_{0j}}{\boldsymbol V}}  \right) =\\
		& p_{\bf B_{0j}} - p_{\bf B_{1j}} - \left( {^{\bf B_{0j}}{\boldsymbol V}_r} - {^{\bf B_{0j}}{\boldsymbol V}}  \right)^T \, \, \mathbf{K}_{\bf B_{0j}} \, \left( {^{\bf B_{0j}}{\boldsymbol V}_r} - {^{\bf B_{0j}}{\boldsymbol V}}  \right),
	\end{array} 
\end{equation}

\begin{equation}
	\begin{array}{ll}
		\dot{\nu}_{B_{1j}} =  & \left( {^{\bf B_{1j}}{\boldsymbol V}_r} - {^{\bf B_{1j}}{\boldsymbol V}}  \right)^T \, \left( {^{\bf B_{1j}}{\boldsymbol F}_r^*} - {^{\bf B_{1j}}{\boldsymbol F}^*}  \right) = \\
		&  \left( {^{\bf B_{1j}}{\boldsymbol V}_r} - {^{\bf B_{1j}}{\boldsymbol V}}  \right)^T \, \left( {^{\bf B_{1j}}{\boldsymbol F}_r} - {^{\bf B_{1j}}{\boldsymbol F}}  \right) - \\ 
		& \left( {^{\bf B_{1j}}{\boldsymbol V}_r} - {^{\bf B_{1j}}{\boldsymbol V}}  \right)^T \,  {^{\bf B_{1j}}\mathbf{U_{E_{1j}}}} \, \left( {^{\bf E_{1j}}{\boldsymbol F}_r} - {^{\bf E_{1j}}{\boldsymbol F}}  \right)+ \\
		&  \left( {^{\bf B_{1j}}{\boldsymbol V}_r} - {^{\bf B_{1j}}{\boldsymbol V}}  \right)^T \,  {^{\bf B_{1j}}\mathbf{U_{P_{1j}}}} \, \left( {^{\bf P_{1j}}{\boldsymbol F}_r} - {^{\bf P_{1j}}{\boldsymbol F}}  \right)  -  \\ & \left( {^{\bf B_{1j}}{\boldsymbol V}_r} - {^{\bf B_{1j}}{\boldsymbol V}}  \right)^T \, \, \mathbf{K}_{\bf B_{1j}} \, \left( {^{\bf B_{1j}}{\boldsymbol V}_r} - {^{\bf B_{1j}}{\boldsymbol V}}  \right) = \\
		&  \left( {^{\bf B_{1j}}{\boldsymbol V}_r} - {^{\bf B_{1j}}{\boldsymbol V}}  \right)^T \, \left( {^{\bf B_{1j}}{\boldsymbol F}_r} - {^{\bf B_{1j}}{\boldsymbol F}}  \right)  - \\
		& \left( {^{\mathbf{E}_{\bf 1j}}\mathbf{U}_\mathbf{B_{1j}}^T}  \, \left( {^{\bf E_{1j}}{\boldsymbol V}_r} - {^{\bf E_{1j}}{\boldsymbol V}}  \right) \right)^T \,  {^{\bf B_{1j}}\mathbf{U_{E_{1j}}}} \,\left( {^{\bf E_{\bf 1j}}{\boldsymbol F}_r} - {^{\bf E_{1j}}{\boldsymbol F}}  \right) +\\
		&   \left( {^{\mathbf{P}_{\bf 1j}}\mathbf{U}_\mathbf{B_{1j}}^T}  \left( {^{\bf P_{1j}}{\boldsymbol V}_r} - {^{\bf P_{1j}}{\boldsymbol V}}  \right) \right)^T {^{\bf B_{1j}}\mathbf{U_{P_{1j}}}}   \left( {^{\bf P_{1j}}{\boldsymbol F}_r} - {^{\bf P_{1j}}{\boldsymbol F}}  \right) - \\
		& \left( {^{\bf B_{1j}}{\boldsymbol V}_r} - {^{\bf B_{1j}}{\boldsymbol V}}  \right)^T   \mathbf{K}_{\bf B_{11}}  \left( {^{\bf B_{1j}}{\boldsymbol V}_r} - {^{\bf B_{1j}}{\boldsymbol V}}  \right)= \\
		
		&\left( {^{\bf B_{1j}}{\boldsymbol V}_r} - {^{\bf B_{1j}}{\boldsymbol V}}  \right)^T \, \left( {^{\bf B_{1j}}{\boldsymbol F}_r} - {^{\bf B_{1j}}{\boldsymbol F}}  \right) + \\ & \left( {^{\bf P_{1j}}{\boldsymbol V}_r} - {^{\bf P_{1j}}{\boldsymbol V}}  \right)^T \, \left( {^{\bf P_{1j}}{\boldsymbol F}_r} - {^{\bf P_{1j}}{\boldsymbol F}}  \right) -
		\\
		&\left( {^{\bf E_{1j}}{\boldsymbol V}_r} - {^{\bf E_{1j}}{\boldsymbol V}}  \right)^T \, \left( {^{\bf E_{1j}}{\boldsymbol F}_r} - {^{\bf E_{1j}}{\boldsymbol F}}  \right)  - \\ & \left( {^{\bf B_{1j}}{\boldsymbol V}_r} - {^{\bf B_{1j}}{\boldsymbol V}}  \right)^T \, \, \mathbf{K}_{\bf B_{1j}} \, \left( {^{\bf B_{1j}}{\boldsymbol V}_r} - {^{\bf B_{1j}}{\boldsymbol V}}  \right) =
		\\
		& p_{\bf B_{1j}} - p_{\bf E_{1j}} + p_{\bf P_{1j}} - \\ 
		&\left( {^{\bf B_{1j}}{\boldsymbol V}_r} - {^{\bf B_{1j}}{\boldsymbol V}}  \right)^T \, \, \mathbf{K}_{\bf B_{1j}} \, \left( {^{\bf B_{1j}}{\boldsymbol V}_r} - {^{\bf B_{1j}}{\boldsymbol V}}  \right),\\
	\end{array} 
	\label{eqn: dnuB1j}
\end{equation}

\begin{equation}
	\begin{array}{ll}
		\dot{\nu}_{B_{4j}} =  & \left( {^{\bf B_{4j}}{\boldsymbol V}_r} - {^{\bf B_{4j}}{\boldsymbol V}}  \right)^T \, \left( {^{\bf B_{4j}}{\boldsymbol F}_r^*} - {^{\bf B_{4j}}{\boldsymbol F}^*}  \right) = \\
		&  \left( {^{\bf B_{4j}}{\boldsymbol V}_r} - {^{\bf B_{4j}}{\boldsymbol V}}  \right)^T \, \left( {^{\bf B_{4j}}{\boldsymbol F}_r} - {^{\bf B_{4j}}{\boldsymbol F}}  \right) -  \\ & \left( {^{\bf B_{4j}}{\boldsymbol V}_r} - {^{\bf B_{4j}}{\boldsymbol V}}  \right)^T \,  {^{\bf B_{4j}}\mathbf{U_{P_{1j}}}} \, \left( {^{\bf P_{1j}}{\boldsymbol F}_r} - {^{\bf P_{1j}}{\boldsymbol F}}  \right) -\\
		&  \left( {^{\bf B_{4j}}{\boldsymbol V}_r} - {^{\bf B_{4j}}{\boldsymbol V}}  \right)^T \, \, \mathbf{K}_{\bf B_{4j}} \, \left( {^{\bf B_{4j}}{\boldsymbol V}_r} - {^{\bf B_{4j}}{\boldsymbol V}}  \right) = \\
		&  \left( {^{\bf B_{4j}}{\boldsymbol V}_r} - {^{\bf B_{4j}}{\boldsymbol V}}  \right)^T \, \left( {^{\bf B_{4j}}{\boldsymbol F}_r} - {^{\bf B_{4j}}{\boldsymbol F}}  \right) - \\&   \left( {^{\mathbf{P}_{\bf 1j}}\mathbf{U}_\mathbf{B_{4j}}^T}  \, \left( {^{\bf P_{1j}}{\boldsymbol V}_r} -  {^{\bf P_{1j}}{\boldsymbol V}}  \right) \right)^T  {^{\bf B_{4j}}\mathbf{U_{P_{1j}}}}  \left( {^{\bf P_{1j}}{\boldsymbol F}_r} - {^{\bf P_{1j}}{\boldsymbol F}}  \right) - \\
		&  \left( {^{\bf B_{4j}}{\boldsymbol V}_r} - {^{\bf B_{\bf 4j}}{\boldsymbol V}}  \right)^T \, \, \mathbf{K}_{\bf B_{4j}} \, \left( {^{\bf B_{4j}}{\boldsymbol V}_r} - {^{\bf B_{4j}}{\boldsymbol V}}  \right) = \\
		& p_{\bf B_{4j}} - p_{\bf P_{1j}} - \left( {^{\bf B_{4j}}{\boldsymbol V}_r} - {^{\bf B_{4j}}{\boldsymbol V}}  \right)^T \, \, \mathbf{K}_{\bf B_{4j}} \, \left( {^{\bf B_{4j}}{\boldsymbol V}_r} - {^{\bf B_{4j}}{\boldsymbol V}}  \right),
	\end{array} 
\end{equation}

\begin{equation}
	\begin{array}{ll}
		\dot{\nu}_{B_{3j}} =  & \left( {^{\bf B_{3j}}{\boldsymbol V}_r} - {^{\bf B_{3j}}{\boldsymbol V}}  \right)^T \, \left( {^{\bf B_{3j}}{\boldsymbol F}_r^*} - {^{\bf B_{3j}}{\boldsymbol F}^*}  \right) = \\
		&  \left( {^{\bf B_{3j}}{\boldsymbol V}_r} - {^{\bf B_{3j}}{\boldsymbol V}}  \right)^T \, \left( {^{\bf B_{3j}}{\boldsymbol F}_r} - {^{\bf B_{3j}}{\boldsymbol F}}  \right) - \\&  \left( {^{\bf B_{3j}}{\boldsymbol V}_r} - {^{\bf B_{3j}}{\boldsymbol V}}  \right)^T \,  {^{\bf B_{3j}}\mathbf{U_{B_{4j}}}} \, \left( {^{\bf B_{4j}}{\boldsymbol F}_r} - {^{\bf B_{4j}}{\boldsymbol F}}  \right) - \\
		& \left( {^{\bf B_{3j}}{\boldsymbol V}_r} - {^{\bf B_{3j}}{\boldsymbol V}}  \right)^T \, \mathbf{K}_{\bf B_{3j}} \, \left( {^{\bf B_{3j}}{\boldsymbol V}_r} - {^{\bf B_{3j}}{\boldsymbol V}}  \right) = \\
		&  (\dot{q}_{j1r} - \dot{q}_{j1}) \, \mathbf{z}_{\tau}^T \, \left( {^{\bf B_{3j}}{\boldsymbol F}_r} - {^{\bf B_{3j}}{\boldsymbol F}}  \right) + \\&  \left( {^{\mathbf{B}_{\bf 2j}}\mathbf{U}_\mathbf{B_{3j}}^T}  \,  \left({^{\bf B_{2j}}{\boldsymbol V}_r} - {^{\bf B_{2j}}{\boldsymbol V}}\right)  \right)^T \, \left( {^{\bf B_{3j}}{\boldsymbol F}_r} - {^{\bf B_{3j}}{\boldsymbol F}}  \right) - \\
		&  \left({^{\mathbf{B}_{\bf 3j}}\mathbf{U}_\mathbf{B_{4j}}^T}  \,  \left( {^{\bf B_{3j}}{\boldsymbol V}_r} - {^{\bf B_{3j}}{\boldsymbol V}}\right)  \right)^T \, \left( {^{\bf B_{4j}}{\boldsymbol F}_r} - {^{\bf B_{4j}}{\boldsymbol F}}  \right) - \\& \left( {^{\bf B_{3j}}{\boldsymbol V}_r} - {^{\bf B_{3j}}{\boldsymbol V}}  \right)^T \, \mathbf{K}_{\bf B_{3j}} \, \left( {^{\bf B_{3j}}{\boldsymbol V}_r} - {^{\bf B_{3j}}{\boldsymbol V}}  \right) = \\
		&  (\dot{q}_{j1r} - \dot{q}_{j1}) \, \mathbf{z}_{\tau}^T \, \left( {^{\bf B_{3j}}{\boldsymbol F}_r} - {^{\bf B_{3j}}{\boldsymbol F}}  \right) + \\&  \left( {^{\bf B_{2j}}{\boldsymbol V}_r} - {^{\bf B_{2j}}{\boldsymbol V}}  \right)^T   \, \left( {^{\bf B_{2j}}{\boldsymbol F}_r} - {^{\bf B_{2j}}{\boldsymbol F}}  \right) \, - \\	
		&   \left( {^{\bf B_{4j}}{\boldsymbol V}_r} - {^{\bf B_{4j}}{\boldsymbol V}}  \right)^T \, \left( {^{\bf B_{4j}}{\boldsymbol F}_r} - {^{\bf B_{4j}}{\boldsymbol F}}  \right)  + \\& (\dot{x}_{jr} - \dot{x}_{j}) \, \mathbf{x}_{f}^T \, \left( {^{\bf B_{4j}}{\boldsymbol F}_r} - {^{\bf B_{4j}}{\boldsymbol F}}  \right)  - \\
		&  \left( {^{\bf B_{3j}}{\boldsymbol V}_r} - {^{\bf B_{3j}}{\boldsymbol V}}  \right)^T \, \mathbf{K}_{\bf B_{3j}} \, \left( {^{\bf B_{3j}}{\boldsymbol V}_r} - {^{\bf B_{3j}}{\boldsymbol V}}  \right) =  \\& p_{\bf B_{2j}} - p_{\bf B_{4j}} + (\dot{x}_{jr} - \dot{x}_{j}) \, (f_{cjr} - f_{cj}) - \\ &
		\left( {^{\bf B_{3j}}{\boldsymbol V}_r} - {^{\bf B_{3j}}{\boldsymbol V}}  \right)^T \, \mathbf{K}_{\bf B_{3j}} \, \left( {^{\bf B_{3j}}{\boldsymbol V}_r} - {^{\bf B_{3j}}{\boldsymbol V}}  \right),
	\end{array} 
\end{equation}

\begin{equation}
	\begin{array}{ll}
		\dot{\nu}_{P_{2j}} =  & \left( {^{\bf P_{2j}}{\boldsymbol V}_r} - {^{\bf P_{2j}}{\boldsymbol V}}  \right)^T \, \left( {^{\bf P_{2j}}{\boldsymbol F}_r^*} - {^{\bf P_{2j}}{\boldsymbol F}^*}  \right) = \\
		&  \left( {^{\bf P_{2j}}{\boldsymbol V}_r} - {^{\bf P_{2j}}{\boldsymbol V}}  \right)^T \, \left( {^{\bf P_{2j}}{\boldsymbol F}_r} - {^{\bf P_{2j}}{\boldsymbol F}}  \right) -  \\ &   \left( {^{\bf P_{2j}}{\boldsymbol V}_r} - {^{\bf P_{2j}}{\boldsymbol V}}  \right)^T \,  {^{\bf P_{2j}}\mathbf{U_{B_{5j}}}} \, \left( {^{\bf B_{5j}}{\boldsymbol F}_r} - {^{\bf B_{5j}}{\boldsymbol F}}  \right) -\\
		&  \left( {^{\bf P_{2j}}{\boldsymbol V}_r} - {^{\bf P_{2j}}{\boldsymbol V}}  \right)^T \, \, \mathbf{K}_{\bf P_{2j}} \, \left( {^{\bf P_{2j}}{\boldsymbol V}_r} - {^{\bf P_{2j}}{\boldsymbol V}}  \right) = \\
		&  \left( {^{\bf P_{2j}}{\boldsymbol V}_r} - {^{\bf P_{2j}}{\boldsymbol V}}  \right)^T \, \left( {^{\bf P_{2j}}{\boldsymbol F}_r} - {^{\bf P_{2j}}{\boldsymbol F}}  \right) - \\ &    \left( {^{\mathbf{P}_{\bf 2j}}\mathbf{U}_\mathbf{B_{5j}}^T}  \, \left( {^{\bf P_{2j}}{\boldsymbol V}_r} - {^{\bf P_{2j}}{\boldsymbol V}}  \right) \right)^T  \, \left( {^{\bf B_{5j}}{\boldsymbol F}_r} - {^{\bf B_{5j}}{\boldsymbol F}}  \right) - \\
		&  \left( {^{\bf P_{2j}}{\boldsymbol V}_r} - {^{\bf P_{\bf 2j}}{\boldsymbol V}}  \right)^T \, \, \mathbf{K}_{\bf P_{2j}} \, \left( {^{\bf P_{2j}}{\boldsymbol V}_r} - {^{\bf P_{2j}}{\boldsymbol V}}  \right) = \\
		& p_{\bf P_{2j}} - p_{\bf B_{5j}} - \left( {^{\bf P_{2j}}{\boldsymbol V}_r} - {^{\bf P_{2j}}{\boldsymbol V}}  \right)^T \, \, \mathbf{K}_{\bf P_{2j}} \, \left( {^{\bf P_{2j}}{\boldsymbol V}_r} - {^{\bf P_{2j}}{\boldsymbol V}}  \right),
	\end{array} 
\end{equation}

\begin{equation}
	\begin{array}{ll}
		\dot{\nu}_{B_{5j}} =  & \left( {^{\bf B_{5j}}{\boldsymbol V}_r} - {^{\bf B_{5j}}{\boldsymbol V}}  \right)^T \, \left( {^{\bf B_{5j}}{\boldsymbol F}_r^*} - {^{\bf B_{5j}}{\boldsymbol F}^*}  \right) = \\
		&  \left( {^{\bf B_{5j}}{\boldsymbol V}_r} - {^{\bf B_{5j}}{\boldsymbol V}}  \right)^T \, \left( {^{\bf B_{5j}}{\boldsymbol F}_r} - {^{\bf B_{5j}}{\boldsymbol F}}  \right) -  \\ &  \left( {^{\bf B_{5j}}{\boldsymbol V}_r} - {^{\bf B_{5j}}{\boldsymbol V}}  \right)^T \,  {^{\bf B_{5j}}\mathbf{U_{P_{3j}}}} \, \left( {^{\bf P_{3j}}{\boldsymbol F}_r} - {^{\bf P_{3j}}{\boldsymbol F}}  \right) -\\
		&  \left( {^{\bf B_{5j}}{\boldsymbol V}_r} - {^{\bf B_{5j}}{\boldsymbol V}}  \right)^T \, \, \mathbf{K}_{\bf B_{5j}} \, \left( {^{\bf B_{5j}}{\boldsymbol V}_r} - {^{\bf B_{5j}}{\boldsymbol V}}  \right) = \\
		
		&  \left( \dot{x}_{tjr} - \dot{x}_{tj} \right) \, (f_{ctjr} - f_{ctj} )  +  \\ & \left( {^{\mathbf{P}_{\bf 2j}}\mathbf{U}_\mathbf{B_{5j}}^T}  \, \left( {^{\bf P_{2j}}{\boldsymbol V}_r} - {^{\bf P_{2j}}{\boldsymbol V}}  \right) \right)^T \, \left( {^{\bf B_{5j}}{\boldsymbol F}_r} - {^{\bf B_{5j}}{\boldsymbol F}}  \right) -    \\
		
		& \left( {^{\mathbf{P}_{\bf 3j}}\mathbf{U}_\mathbf{B_{5j}}^T}  \, \left( {^{\bf P_{3j}}{\boldsymbol V}_r} - {^{\bf P_{3j}}{\boldsymbol V}}  \right) \right)^T  {^{\bf B_{5j}}\mathbf{U_{P_{3j}}}}  \left( {^{\bf P_{3j}}{\boldsymbol F}_r} - {^{\bf P_{3j}}{\boldsymbol F}}  \right) - \\
		&  \left( {^{\bf B_{5j}}{\boldsymbol V}_r} - {^{\bf B_{\bf 5j}}{\boldsymbol V}}  \right)^T \, \, \mathbf{K}_{\bf B_{5j}} \, \left( {^{\bf B_{5j}}{\boldsymbol V}_r} - {^{\bf B_{5j}}{\boldsymbol V}}  \right) = \\
		& \left( \dot{x}_{tjr} - \dot{x}_{tj} \right) \, (f_{ctjr} - f_{ctj} )  + p_{\bf B_{5j}} - p_{\bf P_{3j}} - \\ &  \left( {^{\bf B_{5j}}{\boldsymbol V}_r} - {^{\bf B_{5j}}{\boldsymbol V}}  \right)^T \, \, \mathbf{K}_{\bf B_{5j}} \, \left( {^{\bf B_{5j}}{\boldsymbol V}_r} - {^{\bf B_{5j}}{\boldsymbol V}}  \right),
	\end{array} 
\end{equation}

\begin{equation}
	\begin{array}{ll}
		\dot{\nu}_{P_{3j}} = & \left( {^{\bf P_{3j}}{\boldsymbol V}_r} - {^{\bf P_{3j}}{\boldsymbol V}}  \right)^T \, \left( {^{\bf P_{3j}}{\boldsymbol F}_r^*} - {^{\bf P_{3j}}{\boldsymbol F}^*}  \right) = \\
		& \left( {^{\bf P_{3j}}{\boldsymbol V}_r} - {^{\bf P_{3j}}{\boldsymbol V}}  \right)^T \, \left( {^{\bf P_{3j}}{\boldsymbol F}_r} - {^{\bf P_{3j}}{\boldsymbol F}}  \right)	- \\& \left( {^{\bf P_{3j}}{\boldsymbol V}_r} - {^{\bf P_{3j}}{\boldsymbol V}}  \right)^T \, {^{\bf P_{3j}}\mathbf{U_{E_{2j}}}}  \, \left( {^{\bf E_{2j}}{\boldsymbol F}_r} - {^{\bf E_{2j}}{\boldsymbol F}}  \right)	- \\
		& \left( {^{\bf P_{3j}}{\boldsymbol V}_r} - {^{\bf P_{3j}}{\boldsymbol V}}  \right)^T \, \, \mathbf{K}_{\bf P_{3j}} \, \left( {^{\bf P_{3j}}{\boldsymbol V}_r} - {^{\bf P_{3j}}{\boldsymbol V}}  \right)= \\
		&  \left( {^{\bf P_{3j}}{\boldsymbol V}_r} - {^{\bf P_{3j}}{\boldsymbol V}}  \right)^T \, \left( {^{\bf P_{3j}}{\boldsymbol F}_r} - {^{\bf P_{3j}}{\boldsymbol F}}  \right)	- \\
		&  \left({^{\mathbf{E}_{\bf 2j}}\mathbf{U}_\mathbf{P_{3j}}^T} \left({^{\bf E_{2j}}{\boldsymbol V}_r} - {^{\bf E_{2j}}{\boldsymbol V}}  \right) \right)^T \, {^{\bf P_{3j}}\mathbf{U_{E_{2j}}}}  \, \left( {^{\bf E_{2j}}{\boldsymbol F}_r} - {^{\bf E_{2j}}{\boldsymbol F}}  \right)	- \\	
		& \left( {^{\bf P_{3j}}{\boldsymbol V}_r} - {^{\bf P_{3j}}{\boldsymbol V}}  \right)^T \, \, \mathbf{K}_{\bf P_{3j}} \, \left( {^{\bf P_{3j}}{\boldsymbol V}_r} - {^{\bf P_{3j}}{\boldsymbol V}}  \right) = \\	
		& p_{\bf P_{3j}} - p_{\bf E_{2j}} - \left( {^{\bf P_{3j}}{\boldsymbol V}_r} - {^{\bf P_{3j}}{\boldsymbol V}}  \right)^T \, \, \mathbf{K}_{\bf P_{3j}} \, \left( {^{\bf P_{3j}}{\boldsymbol V}_r} - {^{\bf P_{3j}}{\boldsymbol V}}  \right).
	\end{array} 
	\label{eqn: dnuO2j}
\end{equation}

In the case when the prismatic segment does not exist, the accompanying function for the $j$-th manipulator structure is:
\begin{equation}
	\begin{array}{lc}
		\dot{\nu}_{cl}  = & \dot{\nu}_{B_{1j}}  +  \dot{\nu}_{B_{0j}} + \dot{\nu}_{B_{4j}} + \dot{\nu}_{B_{3j}} =  \\
		&  (\dot{x}_{jr} - \dot{x}_{j}) \, (f_{cjr} - f_{cj}) - p_{\bf E_{1j}} + p_{\bf B_{cj}} -  \\ & \left( {^{\bf B_{1j}}{\boldsymbol V}_r} - {^{\bf B_{1j}}{\boldsymbol V}}  \right)^T \, \, \mathbf{K}_{\bf B_{1j}} \, \left( {^{\bf B_{1j}}{\boldsymbol V}_r} - {^{\bf B_{1j}}{\boldsymbol V}}  \right) - \\ & \left( {^{\bf B_{0j}}{\boldsymbol V}_r} - {^{\bf B_{0j}}{\boldsymbol V}}  \right)^T \, \, \mathbf{K}_{\bf B_{0j}} \, \left( {^{\bf B_{0j}}{\boldsymbol V}_r} - {^{\bf B_{0j}}{\boldsymbol V}}  \right) - \\
		&   \left( {^{\bf B_{4j}}{\boldsymbol V}_r} - {^{\bf B_{4j}}{\boldsymbol V}}  \right)^T \, \, \mathbf{K}_{\bf B_{4j}} \, \left( {^{\bf B_{4j}}{\boldsymbol V}_r} - {^{\bf B_{4j}}{\boldsymbol V}}  \right) -  \\ & \left( {^{\bf B_{3j}}{\boldsymbol V}_r} - {^{\bf B_{3j}}{\boldsymbol V}}  \right)^T \, \mathbf{K}_{\bf B_{3j}} \,  \left( {^{\bf B_{3j}}{\boldsymbol V}_r} - {^{\bf B_{3j}}{\boldsymbol V}}  \right).
	\end{array}
\end{equation}

In the case when a prismatic segment does exist, the accompanying function for the $j$-th manipulator structure becomes:

\begin{equation}
	\begin{array}{lc}
		\dot{\nu}_{jc} = & \dot{\nu}_{B_{5j}} + \dot{\nu}_{P_{2j}}  + \dot{\nu}_{P_{3j}}  + \dot{\nu}_{B_{1j}}  +  \dot{\nu}_{B_{0j}} + \dot{\nu}_{B_{4j}} + \dot{\nu}_{B_{3j}}= \\
		& (\dot{x}_{jr} - \dot{x}_{j}) \, (f_{cjr} - f_{cj}) + \left( \dot{x}_{tjr} - \dot{x}_{tj} \right) \, (f_{ctjr} - f_{ctj} )  +  p_{\bf B_{cj}} - p_{\bf E_{2j}} - \\
		& \left( {^{\bf B_{1j}}{\boldsymbol V}_r} - {^{\bf B_{1j}}{\boldsymbol V}}  \right)^T \, \, \mathbf{K}_{\bf B_{1j}} \, \left( {^{\bf B_{1j}}{\boldsymbol V}_r} - {^{\bf B_{1j}}{\boldsymbol V}}  \right) - \\& \left( {^{\bf B_{0j}}{\boldsymbol V}_r} - {^{\bf B_{0j}}{\boldsymbol V}}  \right)^T \, \, \mathbf{K}_{\bf B_{0j}} \, \left( {^{\bf B_{0j}}{\boldsymbol V}_r} - {^{\bf B_{0j}}{\boldsymbol V}}  \right) - \\
		&   \left( {^{\bf B_{4j}}{\boldsymbol V}_r} - {^{\bf B_{4j}}{\boldsymbol V}}  \right)^T \, \, \mathbf{K}_{\bf B_{4j}} \, \left( {^{\bf B_{4j}}{\boldsymbol V}_r} -  {^{\bf B_{4j}}{\boldsymbol V}}  \right) - \\&  \left( {^{\bf B_{3j}}{\boldsymbol V}_r} - {^{\bf B_{3j}}{\boldsymbol V}}  \right)^T \, \mathbf{K}_{\bf B_{3j}} \,  \left( {^{\bf B_{3j}}{\boldsymbol V}_r} - {^{\bf B_{3j}}{\boldsymbol V}}  \right)-\\
		&  \left( {^{\bf B_{5j}}{\boldsymbol V}_r} - {^{\bf B_{5j}}{\boldsymbol V}}  \right)^T \, \, \mathbf{K}_{\bf B_{5j}} \, \left( {^{\bf B_{5j}}{\boldsymbol V}_r} - {^{\bf B_{5j}}{\boldsymbol V}}  \right) -  \\& \left( {^{\bf P_{2j}}{\boldsymbol V}_r} - {^{\bf P_{2j}}{\boldsymbol V}}  \right)^T \, \, \mathbf{K}_{\bf P_{2j}} \, \left( {^{\bf P_{2j}}{\boldsymbol V}_r} - {^{\bf P_{2j}}{\boldsymbol V}}  \right) - \\
		&  \left( {^{\bf P_{3j}}{\boldsymbol V}_r} - {^{\bf P_{3j}}{\boldsymbol V}}  \right)^T \, \, \mathbf{K}_{\bf P_{3j}} \, \left( {^{\bf P_{3j}}{\boldsymbol V}_r} - {^{\bf P_{3j}}{\boldsymbol V}}  \right).
	\end{array}
	\label{eqn: nujc}
\end{equation}

Time derivatives of the proposed non-negative accompanying functions Eqs. (\ref{eqn: nupj}) and (\ref{eqn: nuptj}) are:
\begin{eqnarray}
	\begin{array}{ll}
		\dot{\nu}_{pj} =  & \dfrac{1}{k_{xj}} \left( f_{pjr} - f_{pj} \right) \, \dfrac{ \dot{f}_{pjr} - \dot{f}_{pj} }{\beta}  = \\
		&  \dfrac{1}{k_{xj}} \left( f_{pjr} - f_{pj} \right) \, \left( u_{fjr} - u_{fj}\right) - \left( f_{cjr} - f_{cj} \right) \, \left( \dot{x}_{jr} - \dot{x}_{j}\right) -
		\\& \left( f_{fjr} - f_{fj} \right) \, \left( \dot{x}_{jr} - \dot{x}_{j}\right) - \dfrac{k_{fj}}{k_{xj}} \, \left( f_{pjr} - f_{pj} \right)^2,
	\end{array}
	\label{eqn: dnupj}
\end{eqnarray}
and
\begin{eqnarray}
	\begin{array}{ll}
		\dot{\nu}_{ptj} = &  \dfrac{1}{k_{xtj}} \left( f_{ptjr} - f_{ptj} \right) \, \dfrac{ \dot{f}_{ptjr} - \dot{f}_{ptj} }{\beta}  = \\
		&  \dfrac{1}{k_{xtj}} \left( f_{ptjr} - f_{ptj} \right) \, \left( u_{ftjr} - u_{ftj}\right) - \left( f_{ctjr} - f_{ctj} \right) \, \left( \dot{x}_{tjr} - \dot{x}_{tj}\right)  - \\ & \left( f_{ftjr} - f_{ftj} \right) \, \left( \dot{x}_{tjr} - \dot{x}_{tj}\right) - \dfrac{k_{ftj}}{k_{xtj}} \, \left( f_{ptjr} - f_{ptj} \right)^2.
	\end{array}
	\label{eqn: dnuptj}
\end{eqnarray}

The time derivative of the proposed accompanying function Eq. (\ref{eqn: nuj}) for the manipulator structure  is obtained by summing  Eqs. (\ref{eqn: dnuB1j})–(\ref{eqn: dnuO2j}) and Eqs. (\ref{eqn: dnupj}) and (\ref{eqn: dnuptj}) as:
\begin{equation}
	\begin{array}{ll}
		\dot{\nu}_{j} = 	& \dot{\nu}_{B_{5j}} + \dot{\nu}_{P_{2j}}  + \dot{\nu}_{P_{3j}}  + \dot{\nu}_{B_{1j}}  +  \dot{\nu}_{B_{0j}} + \dot{\nu}_{B_{4j}} + \dot{\nu}_{B_{3j}} + \dot{\nu}_{pj} + \dot{\nu}_{ptj} = \\
		& - \displaystyle\sum_{\bf A \in S_j} \left( {^{\bf A}{\boldsymbol V}_r} - {^{\bf A}{\boldsymbol V}}  \right)^T \, \, \mathbf{K}_{\bf A} \, \left( {^{\bf A}{\boldsymbol V}_r} - {^{\bf A}{\boldsymbol V}}  \right) +  p_{\bf B_{cj}} - p_{\bf E_{2j}} + \\
		& \dfrac{ \left( f_{pjr} - f_{pj} \right) \, \left( u_{fjr} - u_{fj}\right)}{k_{xj}} + \dfrac{\left( f_{ptjr} - f_{ptj} \right) \, \left( u_{ftjr} - u_{ftj}\right)}{k_{xtj}}  - \\
		&  - \left( f_{fjr} - f_{fj} \right) \, \left( \dot{x}_{jr} - \dot{x}_{j}\right)  - \left( f_{ftjr} - f_{ftj} \right) \, \left( \dot{x}_{tjr} - \dot{x}_{tj}\right) -\\
		& \dfrac{k_{fj}}{k_{xj}} \, \left( f_{pjr} - f_{pj} \right)^2 - \dfrac{k_{ftj}}{k_{xtj}} \, \left( f_{ptjr} - f_{ptj} \right)^2.
	\end{array}
	\label{eqn: dnuj}
\end{equation}

\newpage

The stability-preventing terms $(\dot{x}_{jr} - \dot{x}_{j}) \, (f_{cjr} - f_{cj})$ and  $\left( \dot{x}_{tjr} - \dot{x}_{tj} \right) \, (f_{ctjr} - f_{ctj} )$ from Eq. (\ref{eqn: nujc})  are replaced with terms  $\left( f_{pjr} - f_{pj} \right) \, \left( u_{fjr} - u_{fj}\right) $, \newline $ \left( f_{ptjr} - f_{ptj} \right) \, \left( u_{ftjr} - u_{ftj}\right) $, $\left( f_{fjr} - f_{fj} \right) \, \left( \dot{x}_{jr} - \dot{x}_{j}\right)$ and $\left( f_{ftjr} - f_{ftj} \right) \, \left( \dot{x}_{tjr} - \dot{x}_{tj}\right)$ in the expression Eq. (\ref{eqn: dnuj}) for the whole $j$-th manipulator structure.

Simply choosing $u_{fjr} = u_{fj}$ and $u_{ftjr} = u_{ftj}$, two out of four stability-preventing terms in Eq. (\ref{eqn: dnuj}) disappear.

The friction model is already assumed to be increasing, continuous and antisymmetric, so using this property, it follows that:
\begin{equation}
	- \left( f_{fjr} - f_{fj} \right) \, \left( \dot{x}_{jr} - \dot{x}_{j}\right)  \leqslant 0,
\end{equation}
and
\begin{equation}
	- \left( f_{ftjr} - f_{ftj} \right) \, \left( \dot{x}_{tjr} - \dot{x}_{tj}\right)  \leqslant 0.
\end{equation}
This makes it possible to write the time derivative Eq. (\ref{eqn: dnuj}) in its final form:
\begin{equation}
	\begin{array}{ll}
		\dot{\nu}_{j} = 	& \dot{\nu}_{B_{5j}} + \dot{\nu}_{P_{2j}}  + \dot{\nu}_{P_{3j}}  + \dot{\nu}_{B_{1j}}  +  \dot{\nu}_{B_{0j}} + \dot{\nu}_{B_{4j}} + \dot{\nu}_{B_{3j}} + \dot{\nu}_{pj} + \dot{\nu}_{ptj} \leqslant \\
		& - \displaystyle\sum_{\bf A \in S_j} \left( {^{\bf A}{\boldsymbol V}_r} - {^{\bf A}{\boldsymbol V}}  \right)^T \, \, \mathbf{K}_{\bf A} \, \left( {^{\bf A}{\boldsymbol V}_r} - {^{\bf A}{\boldsymbol V}}  \right) +  p_{\bf B_{cj}} - p_{\bf E_{2j}} - \\
		& \dfrac{k_{fj}}{k_{xj}} \, \left( f_{pjr} - f_{pj} \right)^2 - \dfrac{k_{ftj}}{k_{xtj}} \, \left( f_{ptjr} - f_{ptj} \right)^2.
	\end{array}
	\label{eqn: dnuj2}
\end{equation}
\noindent This qualifies the $j$-th hydraulic manipulator structure from Fig. \ref{fig: closed chain} as  virtually stable in the sense of Definition \ref{def: virt stab}.

Consequently, from  Theorem \ref{thm: virt stab}, it follows that:
\begin{equation}
	\begin{array}{cl}
		f_{pjr} - f_{pj}  & \in L_2 \cap L_\infty,
	\end{array}
	\label{eqn: dnuj3}
\end{equation}
\begin{equation}
	\begin{array}{cl}
		f_{ptjr} - f_{ptj}  & \in L_2 \cap L_\infty, 
	\end{array}
	\label{eqn: dnuj4}
\end{equation}
and
\begin{equation}
	\begin{array}{cl}
		{^{\bf A}{\boldsymbol V}_r} - {^{\bf A}{\boldsymbol V}}  & \in L_2 \cap L_\infty .
	\end{array}
	\label{eqn: dnuj5}
\end{equation}

In addition, Eqs. (\ref{eqn: dnuj5}) imply that:
\begin{equation}
	\begin{array}{cl}
		\dot{q}_{jr} - \dot{q}_{j}  & \in L_2 \cap L_\infty,
	\end{array}
	\label{eqn: dnuj6}
\end{equation}
and
\begin{equation}
	\begin{array}{cl}
		\dot{x}_{tjr} - \dot{x}_{tj}  & \in L_2 \cap L_\infty.
	\end{array}
	\label{eqn: dnuj7}
\end{equation}

Given a bounded $\dot{q}_{jr}$ and $\dot{x}_{tjr}$,  the boundedness of $\dot{q}_{j}$ and $\dot{x}_{tj}$ is ensured from Eqs. (\ref{eqn: dnuj6}) and  (\ref{eqn: dnuj7}), respectively. This guarantees the boundedness of friction forces $f_{fj}$ and $f_{ftj}$. Having bounded required accelerations $\ddot{q}_{jr} \in L_{\infty}$ and $\ddot{x}_{tjr} \in L_{\infty}$ implies bounded  $f_{cjr}$ and $f_{ctjr}$ and consequently $f_{pjr} \in L_{\infty}$ and $f_{ptjr} \in L_{\infty}$. In turn, Eqs. (\ref{eqn: dnuj3}) and  (\ref{eqn: dnuj4}) imply $f_{pj} \in L_{\infty}$ and $f_{ptj} \in L_{\infty}$. The boundedness of $f_{pj}$, $f_{ptj}$, $\dot{q}_j$ and $\dot{x}_{tj}$ implies the boundedness of $f_{cj}$ and $f_{ctj}$. Boundedness of all actuation forces implies bounded accelerations $\ddot{q}_j$ and $\ddot{x}_{tj}$. Moreover, asymptotic convergence of all $L_2$ signals with bounded time derivatives are ensured from Lemma \ref{lemma: Tao}.

That is, from $\ddot{q}_{jr} - \ddot{q}_j \in L_{\infty}$ and Eq. (\ref{eqn: dnuj6}). it follows that $\dot{q}_{jr} - \dot{q}_j \rightarrow 0$, and this in turn guarantees $\dot{q}_{jd} - \dot{q}_j \rightarrow 0$ and ${q}_{jd} - {q}_j \rightarrow 0$ from Eq. (\ref{eqn: qjr}). As a consequence, $\dot{x}_{jd} - \dot{x}_{j} \rightarrow 0$ and ${x}_{jd} - {x}_{j} \rightarrow 0$ are guaranteed, per Eq. (\ref{eqn: xj}) and  Eq. (\ref{eqn: dxj}). On the other hand, $\ddot{x}_{tjd} - \ddot{x}_{tj} \in L_{\infty}$ and Eq. (\ref{eqn: dnuj7}) imply $\dot{x}_{tjr} - \dot{x}_{tj} \rightarrow 0$, so $\dot{x}_{tjd} - \dot{x}_{tj} \rightarrow 0$ and ${x}_{tjd} - {x}_{tj} \rightarrow 0$.

\end{pf}

\bibliographystyle{unsrtnat}

\begin{thebibliography}{10}
	\expandafter\ifx\csname url\endcsname\relax
	\def\url#1{\texttt{#1}}\fi
	\expandafter\ifx\csname urlprefix\endcsname\relax\def\urlprefix{URL }\fi
	\expandafter\ifx\csname href\endcsname\relax
	\def\href#1#2{#2} \def\path#1{#1}\fi
	
	\bibitem{slotine1987adaptive}
	J.-J.~E. Slotine, W.~Li, On the adaptive control of robot manipulators, The
	International Journal Of Robotics Research 6~(3) (1987) 49--59.
	
	\bibitem{khosla1988experimental}
	P.~K. Khosla, T.~Kanade, Experimental evaluation of nonlinear feedback and
	feedforward control schemes for manipulators, The International Journal of
	Robotics Research 7~(1) (1988) 18--28.
	
	\bibitem{lewis2003robot}
	F.~L. Lewis, D.~M. Dawson, C.~T. Abdallah, Robot manipulator control: theory
	and practice, CRC Press, 2003.
	
	\bibitem{zhu2013precision}
	W.-H. Zhu, T.~Lamarche, E.~Dupuis, D.~Jameux, P.~Barnard, G.~Liu, Precision
	control of modular robot manipulators: {T}he {VDC} approach with embedded
	{FPGA}, IEEE Transactions on Robotics 29~(5) (2013) 1162--1179.
	
	\bibitem{semini2010hyq}
	C.~Semini, Hy{Q}-design and development of a hydraulically actuated quadruped
	robot, Doctor of {P}hilosophy ({Ph. D.}) {T}hesis, University of Genoa, Italy
	(2010).
	
	\bibitem{son2020expert}
	B.~Son, C.~Kim, C.~Kim, D.~Lee, Expert-emulating excavation trajectory planning
	for autonomous robotic industrial excavator, in: 2020 IEEE/RSJ International
	Conference on Intelligent Robots and Systems (IROS), IEEE, 2020, pp.
	2656--2662.
	
	\bibitem{mattila2017survey}
	J.~Mattila, J.~Koivum{\"a}ki, D.~G. Caldwell, C.~Semini, A survey on control of
	hydraulic robotic manipulators with projection to future trends, IEEE/ASME
	Transactions on Mechatronics 22~(2) (2017) 669--680.
	
	\bibitem{bech2013experimental}
	M.~M. Bech, T.~O. Andersen, H.~C. Pedersen, L.~Schmidt, Experimental evaluation
	of control strategies for hydraulic servo robot, in: 2013 IEEE International
	Conference on Mechatronics and Automation, IEEE, 2013, pp. 342--347.
	
	\bibitem{siciliano2016springer}
	B.~Siciliano, O.~Khatib, Springer handbook of robotics, Springer, 2016.
	
	\bibitem{jazar2010theory}
	R.~N. Jazar, Theory of applied robotics: kinematics, dynamics, and control,
	Springer Science \& Business Media, 2010.
	
	\bibitem{kane1983use}
	T.~R. Kane, D.~A. Levinson, The use of {K}ane's dynamical equations in
	robotics, The International Journal of Robotics Research 2~(3) (1983) 3--21.
	
	\bibitem{marques2021examination}
	F.~Marques, I.~Roupa, M.~T. Silva, P.~Flores, H.~M. Lankarani, Examination and
	comparison of different methods to model closed loop kinematic chains using
	lagrangian formulation with cut joint, clearance joint constraint and elastic
	joint approaches, Mechanism and Machine Theory 160 (2021) 104294.
	
	\bibitem{habibi1991computed}
	S.~Habibi, R.~Richards, Computed-torque and variable-structure multi-variable
	control of a hydraulic industrial robot, Proceedings of the Institution of
	Mechanical Engineers, Part I: Journal of Systems and Control Engineering
	205~(2) (1991) 123--140.
	
	\bibitem{bu2000observer}
	F.~Bu, B.~Yao, Observer based coordinated adaptive robust control of robot
	manipulators driven by single-rod hydraulic actuators, in: Proceedings 2000
	ICRA. Millennium Conference. IEEE International Conference on Robotics and
	Automation. Symposia Proceedings (Cat. No. 00CH37065), Vol.~3, IEEE, 2000,
	pp. 3034--3039.
	
	\bibitem{mattila2000energy}
	J.~Mattila, T.~Virvalo, Energy-efficient motion control of a hydraulic
	manipulator, in: Proceedings 2000 ICRA. Millennium Conference. IEEE
	International Conference on Robotics and Automation. Symposia Proceedings
	(Cat. No. 00CH37065), Vol.~3, IEEE, 2000, pp. 3000--3006.
	
	\bibitem{vsalinic2014dynamic}
	S.~{\v{S}}alini{\'c}, G.~Bo{\v{s}}kovi{\'c}, M.~Nikoli{\'c}, Dynamic modelling
	of hydraulic excavator motion using {K}ane's equations, Automation in
	Construction 44 (2014) 56--62.
	
	\bibitem{cibicik2019dynamic}
	A.~Cibicik, O.~Egeland, Dynamic modelling and force analysis of a knuckle boom
	crane using screw theory, Mechanism and Machine Theory 133 (2019) 179--194.
	
	\bibitem{featherstone1987robot}
	R.~Featherstone, Robot dynamics algorithms, Kluwer Academic Publishers,
	Boston/Dordrecht/Lancaster, 1987.
	
	\bibitem{zhu2010virtual}
	W.-H. Zhu, Virtual decomposition control: toward hyper degrees of freedom
	robots, Vol.~60, Springer Science \& Business Media, 2010.
	
	\bibitem{mastellone2021impact}
	S.~Mastellone, A.~van Delft, The impact of control research on industrial
	innovation: What would it take to make it happen?, Control Engineering
	Practice 111 (2021) 104737.
	
	\bibitem{zhu1998adaptive}
	W.-H. Zhu, Z.~Bien, J.~De~Schutter, Adaptive motion/force control of multiple
	manipulators with joint flexibility based on virtual decomposition, IEEE
	Transactions on Automatic Control 43~(1) (1998) 46--60.
	
	\bibitem{zhu2002experimental}
	W.-H. Zhu, J.~De~Schutter, Experimental verifications of
	virtual-decomposition-based motion/force control, IEEE Transactions on
	Robotics and Automation 18~(3) (2002) 379--386.
	
	\bibitem{zhu2005adaptive}
	W.-H. Zhu, J.-C. Piedboeuf, Adaptive output force tracking control of hydraulic
	cylinders with applications to robot manipulators, Journal of Dynamic
	Systems, Measurement and Control 127~(2) (2005) 206--217.
	
	\bibitem{koivumaki2015stability}
	J.~Koivum{\"a}ki, J.~Mattila, Stability-guaranteed force-sensorless contact
	force/motion control of heavy-duty hydraulic manipulators, IEEE Transactions
	on Robotics 31~(4) (2015) 918--935.
	
	\bibitem{koivumaki2016stability}
	J.~Koivum{\"a}ki, J.~Mattila, Stability-guaranteed impedance control of
	hydraulic robotic manipulators, IEEE/ASME Transactions On Mechatronics 22~(2)
	(2016) 601--612.
	
	\bibitem{koivumaki2019energy}
	J.~Koivum{\"a}ki, W.-H. Zhu, J.~Mattila, Energy-efficient and high-precision
	control of hydraulic robots, Control Engineering Practice 85 (2019) 176--193.
	
	\bibitem{koivumaki2018addressing}
	J.~Koivum{\"a}ki, W.-H. Zhu, J.~Mattila, Addressing closed-chain dynamics for
	high-precision control of hydraulic cylinder actuated manipulators, in:
	BATH/ASME 2018 Symposium on Fluid Power and Motion Control, American Society
	of Mechanical Engineers Digital Collection, 2018.
	
	\bibitem{tao1997simple}
	G.~Tao, A simple alternative to the {B}arbalat lemma, IEEE Transactions on
	Automatic Control 42~(5) (1997) 698.
	
	\bibitem{murray1989dynamic}
	J.~J. Murray, G.~H. Lovell, Dynamic modeling of closed-chain robotic
	manipulators and implications for trajectory control, IEEE Transactions on
	Robotics and Automation 5~(4) (1989) 522--528.
	
	\bibitem{luh1985computation}
	J.~Luh, Y.-F. Zheng, Computation of input generalized forces for robots with
	closed kinematic chain mechanisms, IEEE Journal on Robotics and Automation
	1~(2) (1985) 95--103.
	
	\bibitem{lin1990dynamics}
	S.-K. Lin, Dynamics of the manipulator with closed chains, IEEE Transactions on
	Robotics and Automation 6~(4) (1990) 496--501.
	
	\bibitem{lampinen2019improved}
	S.~Lampinen, J.~Koivum{\"a}ki, J.~Mattila, Improved hydraulic cylinder model
	for the virtual decomposition control approach, in: 2019 IEEE International
	Conference on Cybernetics and Intelligent Systems (CIS) and IEEE Conference
	on Robotics, Automation and Mechatronics (RAM), IEEE, 2019, pp. 113--118.
	
	\bibitem{mustalahti2017nonlinear}
	P.~Mustalahti, J.~Mattila, Nonlinear full-model-based controller for unactuated
	joints in vertical plane, in: 2017 IEEE International Conference on
	Cybernetics and Intelligent Systems (CIS) and IEEE Conference on Robotics,
	Automation and Mechatronics (RAM), IEEE, 2017, pp. 201--206.
	
	\bibitem{mustalahti2019nonlinear}
	P.~Mustalahti, J.~Mattila, Nonlinear model-based control design for a
	hydraulically actuated spherical wrist, in: Fluid Power Systems Technology,
	Vol. 59339, American Society of Mechanical Engineers, 2019, p. V001T01A027.
	
\end{thebibliography}

\end{document}